\documentclass[10pt]{article} 
\usepackage[accepted]{rlc}

\usepackage{amsmath}
\usepackage{amssymb}
\usepackage{mathtools}
\usepackage{amsthm}
\usepackage[pdftex]{graphicx}           
\usepackage[space]{grffile}     
\usepackage{url}                

\usepackage{algorithm}
\usepackage{algpseudocode}
\algrenewcommand\algorithmicrequire{\textbf{Input:}}
\algrenewcommand\algorithmicensure{\textbf{Output:}}
\usepackage{hyperref}
\usepackage[capitalise]{cleveref}
\usepackage{subcaption}
\usepackage{xcolor,colortbl}

\DeclareMathOperator*{\argmax}{arg\,max}

\title{Planning to Go Out-of-Distribution \\ in Offline-to-Online Reinforcement Learning}

\author{Trevor McInroe \\ University of Edinburgh \\ \texttt{t.mcinroe@ed.ac.uk}
    \And
    Adam Jelley \\ University of Edinburgh \\ \texttt{adam.jelley@ed.ac.uk}
    \And
    Stefano V. Albrecht \\ University of Edinburgh \\ \texttt{s.albrecht@ed.ac.uk}
    \And
    Amos Storkey \\ University of Edinburgh \\ \texttt{a.storkey@ed.ac.uk}
}

\begin{document}

\maketitle

\begin{abstract}
Offline pretraining with a static dataset followed by online fine-tuning (offline-to-online, or OtO) is a paradigm well matched to a real-world RL deployment process. In this scenario, we aim to find the best-performing policy within a limited budget of online interactions. Previous work in the OtO setting has focused on correcting for bias introduced by the policy-constraint mechanisms of offline RL algorithms. Such constraints keep the learned policy close to the behavior policy that collected the dataset, but we show this can unnecessarily limit policy performance if the behavior policy is far from optimal. Instead, we forgo constraints and frame OtO RL as an exploration problem that aims to maximize the benefit of online data-collection. We first study the major online RL exploration methods based on intrinsic rewards and UCB in the OtO setting, showing that intrinsic rewards add training instability through reward-function modification, and UCB methods are myopic and it is unclear which learned-component's ensemble to use for action selection. We then introduce an algorithm for \textbf{p}lanning \textbf{t}o \textbf{g}o \textbf{o}ut-\textbf{o}f-\textbf{d}istribution (PTGOOD) that avoids these issues. PTGOOD uses a non-myopic planning procedure that targets exploration in relatively high-reward regions of the state-action space unlikely to be visited by the behavior policy. By leveraging concepts from the Conditional Entropy Bottleneck, PTGOOD encourages data collected online to provide new information relevant to improving the final deployment policy without altering rewards. We show empirically in several continuous control tasks that PTGOOD significantly improves agent returns during online fine-tuning and avoids the suboptimal policy convergence that many of our baselines exhibit in several environments.
\end{abstract}

\section{Introduction}

In real-world reinforcement learning (RL), there is great value in being able to train an agent offline with a static dataset~\citep{offline-rl-levine}. While offline RL (also called batch RL~\citep{batch-rl1,batch-rl2}) removes traditional RL's potentially costly data-collection step, the resulting policy may be suboptimal. This could occur if the offline dataset does not cover all areas of the state-action space relevant to our task or if the policy that collected the dataset was itself suboptimal. Given this risk, fine-tuning an RL agent over a small budget of online interactions would be useful in real-world deployments.

In this study, we view this offline-to-online (OtO) scenario as an exploration problem. Because the agent has a limit on its environment interactions, it must choose carefully which state-action pairs to collect during online fine-tuning. This contrasts starkly with prior work in OtO RL, which has focused on correcting for bias introduced by the constraint mechanisms used in existing offline RL algorithms~\citep{td3bc-improvement,cal-ql,td3-c}. Such policy-constraint mechanisms are used during offline training to keep the learned policy close to the behavior policy that collected the offline dataset (e.g., the inclusion of a behavior-cloning term~\citep{minimalist}). While these methods can work well offline, they can cause detrimental learning instabilities during online fine-tuning, due to overly-conservative value functions~\citep{cal-ql}. Instead, \textbf{we do not use these constraint mechanisms at any point}. In doing so, we shift the problem set away from bias correction to data-collection strategy during the online fine-tuning phase.

While exploration is widely studied in the online RL literature, the OtO problem differs from the standard online learning setup in two unique ways. First, the OtO setting greatly constrains the number of online data-collection steps. Second, the online phase in OtO RL can benefit from information available from offline pretraining. Because exploration methods have generally not featured in the OtO RL literature, we evaluate the compatibility of major online RL exploration paradigms with the OtO setting. In particular, we analyze intrinsic motivation~\citep{intrinsic-orig} and upper confidence bound (UCB) exploration~\citep{ucb-original}. We find that intrinsic-motivation methods can forget initializations from offline pretraining due to reward-function alteration and that the implementation details of UCB-style methods can affect exploration behavior. Further, UCB methods only consider exploration consequences in the immediate next-state (i.e., are myopic). Ultimately, we propose modifications to intrinsic-motivation methods to address their issues and highlight UCB methods' shortcomings, leading to several effective OtO baselines.

The aforementioned issues with online exploration methods in OtO RL lead us to develop an algorithm for \textbf{p}lanning \textbf{t}o \textbf{g}o \textbf{o}ut \textbf{o}f \textbf{d}istribution (PTGOOD) that can be exploited by existing model-based RL algorithms. PTGOOD first learns a density of state-action pairs in the offline dataset via the Conditional Entropy Bottleneck~\citep{ceb}. This density is used to identify transitions during online fine-tuning that are out-of-distribution relative to the data in the offline dataset without altering rewards. By targeting such state-action pairs, PTGOOD continually increases the diversity of the information available in the total (offline plus online) data. 
PTGOOD also targets high-reward state-action pairs by ensuring that exploration occurs near the current-best policy to ensure \emph{relevance} of the collected data. PTGOOD uses the learned density in a non-myopic planning procedure, thereby considering exploration fruitfulness in future steps. 

Our experiments in continuous control tasks demonstrate that PTGOOD consistently and significantly outperforms our OtO baselines in terms of evaluation returns and avoids suboptimal policy convergence, a problem we find with many OtO methods in several environments. In addition, we find that PTGOOD often finds the optimal policy in simpler environments such as Walker in as few as 10k online steps and in as few as 50k in more complex control tasks like Humanoid, even when the behavior policy was highly suboptimal (e.g., random). 


\section{Background}\label{sec:bg}
The RL problem usually studies an agent acting within a Markov decision process (MDP) parameterized by the tuple $(\mathcal{S}, \mathcal{A}, \mathcal{T}, R, \gamma)$. $\mathcal{S}, \mathcal{A}$ are the state- and action-spaces, respectively, $\mathcal{T}(s^{\prime} | s, a)$ is the transition function that describes the distribution over next-states conditioned on the current state and action, $R(s,a)$ is the reward function, and $\gamma \in (0,1)$ is the discount factor. The agent acts within the MDP according to its policy $\pi(a \vert s)$, which maps states to a distribution over actions. An agent's policy $\pi$ induces a (discounted) occupancy measure $\rho_{\pi}(s,a)$, which is the stationary distribution over the $\mathcal{S} \times \mathcal{A}$ space unique to policy $\pi$~\citep{occupancy-measure-unique1,occupancy-measure-unique2}. After executing an action $a_t$ in state $s_t$ at timestep $t$, the next state is sampled $s_{t+1} \sim \mathcal{T}(\cdot | s_t, a_t)$, the agent receives a reward $r_t = R(s_t, a_t)$, and the interaction loop continues. The agent's learning objective is to find the optimal policy that maximizes cumulative discounted returns $\pi^{\ast} = \arg\max_{\pi}  \mathbb{E}_{\pi} [\sum_{t=1}^\infty \gamma^{t-1}R(s_t,a_t)]$. Model-based RL approaches learn a model of the MDP's transition function $\hat{\mathcal{T}}$ and reward function $\hat{R}$, which can then be used to generate rollouts of ``imagined'' trajectories from a given state $s_t$: $\tau = (s_t, a_t, \hat{r}_t, \hat{s}_{t+1}, \dots)$. 

OtO RL assumes access to a dataset of transition tuples $D_{\pi_b} = \{(s, a, r, s^{\prime})_i\}_{i=1}^{\lvert D_{\pi_b} \rvert}$ collected by some (potentially) unknown behavior policy $\pi_b$. This behavior policy's performance can range from that of a random agent to an expert agent, which means that $D_{\pi_b}$ may contain trajectories of highly-suboptimal behavior. The goal in OtO RL is to leverage offline data $D_{\pi_b}$ to determine a policy $\pi_o$ to collect another dataset $D_{\pi_o}$ over a fixed-budget of agent-environment interactions, which are used together $D_{\pi_b} \cup D_{\pi_o}$ to train a final policy $\pi_f$ that is as close as possible in performance to the optimal policy $\pi^{\ast}$. The problem is to optimize over both the choice of final policy $\pi_f$ and the data collection process that leads to that final policy.

\section{Related Work}

\textbf{Exploration in RL.} Exploration is a key problem in RL and has been studied extensively in the online setting. Exploration algorithms cover many strategies such as dithering methods like $\epsilon$-greedy or randomized value functions~\citep{deep-explore}. Intrinsic reward methods leverage prediction error~\citep{icm,rnd} and count-based rewards~\citep{density-count} to guide agents towards unseen regions of the state-action space. Upper confidence bound (UCB) methods use uncertainty to guide agent exploration. For example, some algorithms measure uncertainty as disagreement within ensembles of Q-functions~\citep{orig-ucb-ens,sunrise,emax} or transition functions~\citep{max-model-explore,t-disagreement,plan2explore}. In contrast to these methods, PTGOOD uses prior information explicitly by estimating a density of already-collected data and uses this density to plan exploration.

\textbf{Offline RL.} Many offline RL methods are designed to constrain the learned policy to be similar to the behavior policy. For example, conservative methods incorporate a policy constraint either via behavior cloning terms~\citep{brac,awr,minimalist}, restricting the policy-search space~\citep{bear}, restricting the policy's action space~\citep{overest-fujimoto-2}, or policy-divergence regularization in the critic~\citep{nachum2019algaedice,fisher-brc}. Pessimistic methods suppress the value of out-of-distribution state-action pairs, disincentivizing the agent from traversing those regions. For example,~\citet{morel} and~\citet{mopo} penalize value based on ensemble disagreement,~\citet{rambo-rl} use an adversarial world model to generate pessimistic transitions, and~\citet{cql} penalize the value of actions too different from ones the behavior policy would choose. Tangentially related to offline RL is off-policy evaluation, which studies how to evaluate (but not improve) policies using an offline dataset (e.g.,~\citep{zhong2022datacollection}).

\textbf{OtO RL.} Some research in the OtO RL setting involves empirical studies of algorithm implementation choices. For example,~\citet{oto2021} and~\citet{moore} develop a replay sampling mechanism to mitigate distribution-shift issues, and~\citet{rlpd} study choices like using LayerNorm and sampling proportions between offline and online data. Most previous work in the OtO setting targets over-conservatism induced by a given offline RL algorithm~\citep{td3bc-improvement,cal-ql,td3-c}. In contrast, PTGOOD approaches the OtO RL setting as an exploration problem.~\citet{tabular-oto} show theoretically that the exploration perspective is useful for OtO in tabular MDPs when combined with pessimism. In contrast, we focus on MDPs with continuous state- and action-spaces, and PTGOOD does not use conservatism or pessimism.

\textbf{Control with Expert Demonstrations.}
Closely related to OtO RL is learning from demonstration (LFD)~\citep{lfd}. Many LFD methods use a form of behavior cloning on expert or hand-crafted trajectories for policy initialization followed by online fine-tuning with RL operators~\citep{lfd2,lfd3,expert-demonstrations1}. In contrast, we study a setting where the learned policy has \textbf{no prior access to demonstrations from expert or hand-crafted policies.}

\section{Online Exploration Methods in the OtO Setting}\label{sec:explore-in-oto}

Motivated by the lack of current OtO exploration algorithms, we now study two common online exploration methods based on intrinsic rewards (\S\ref{sec:intrinsic-rewards}) and UCB exploration (\S\ref{sec:ucb-methods}) in the OtO setting. In summary, we find that offline initializations can be unlearned when the intrinsic rewards introduced during online fine-tuning are too large relative to the true rewards used during offline pretraining. With UCB methods, we find that the choice of ensemble over which uncertainty is computed changes exploration behavior, which is critical in OtO RL. Despite the popularity of Q-function ensembles, it is not clear whether collecting data to reduce value uncertainty is better than reducing uncertainty in other learned components, such as learned transition functions. 

\subsection{Intrinsic Rewards}\label{sec:intrinsic-rewards}
Intrinsic-reward methods modify the reward $r_t = r_t^e + \lambda r_t^i$ at timestep $t$ as the sum of the MDP's true (extrinsic) reward $r^e_t$ and an intrinsic reward $r^i_t$ weighted by $\lambda$. Intrinsic rewards guide exploration by giving the agent a bonus reward in relatively unexplored areas of the MDP. For example, Random Network Distillation (RND)~\citep{rnd} trains a network to predict the output of a fixed randomly-initialized network that transforms a given state. Here, the prediction error is used as the bonus reward $r^i_t$, thereby leading the agent to explore unseen areas of the state space.

Because exploration is impossible during offline pretraining, intrinsic-reward methods must use a two-stage reward function in the OtO setting: one for offline exploitation (only $r^e$) and one for online exploration ($r^e$ and $r^i$ together). We hypothesize that this two-stage reward function is problematic in the OtO setting. If $r^i$ is too large relative to $r^e$, we risk destroying the initialization of the pretrained critic, which destroys the initialization of the pretrained actor. 

To test our hypothesis, we evaluate RND agents with $\lambda \in \{0, 0.1, 1, 10, 50\}$ in two environment-dataset combinations. We use the Halfcheetah (Random) dataset from D4RL~\citep{d4rl} and collect our own dataset in the DeepMind Control Suite~\citep{dmcontrol-paper,dmcontrol-software} in the Walker environment, which we call DMC Walker (Random). Both datasets were collected with behavior policies that select actions uniformly at random.\footnote{For more details on environments and datasets, see Appendix~\ref{app:envs}.} All agents are pretrained offline with the true rewards ($r^e$), fine-tuned online over 50k agent-environment interactions with the RND-altered rewards ($r^e$ and $r^i$ together), and use Model-Based Policy Optimization (MBPO)~\citep{mbpo} combined with Soft Actor-Critc (SAC)~\citep{sac} as the base agent.\footnote{For more details on agents, see Appendix~\ref{app:arch-and-hypers}.} Every 1k environment steps, we collect the agents' average undiscounted returns over ten evaluation episodes.

\begin{figure*}[!t]
    \centering
    \begin{subfigure}{0.45\textwidth}
        \includegraphics[width=\textwidth]{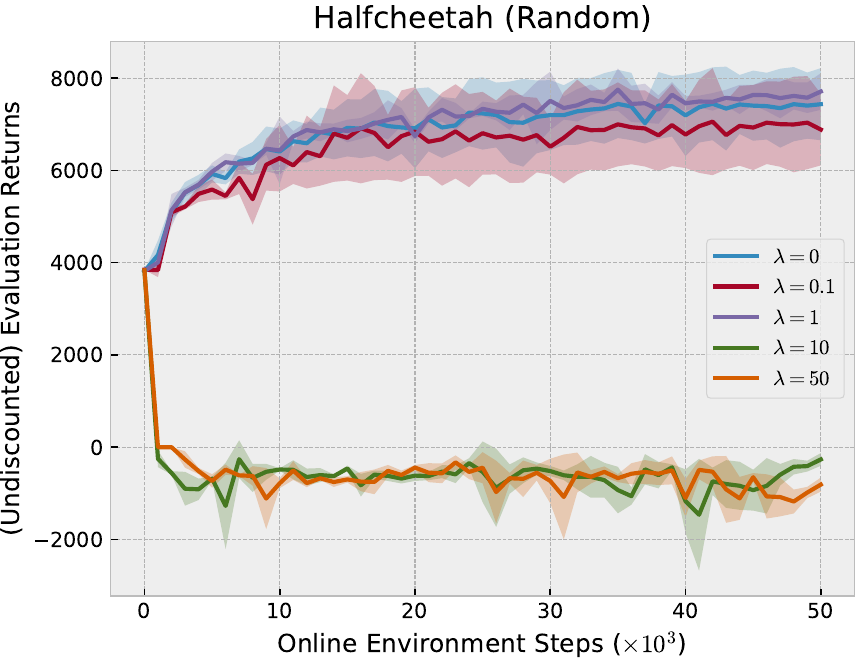}
    \end{subfigure}
    \begin{subfigure}{0.45\textwidth}
        \includegraphics[width=\textwidth]{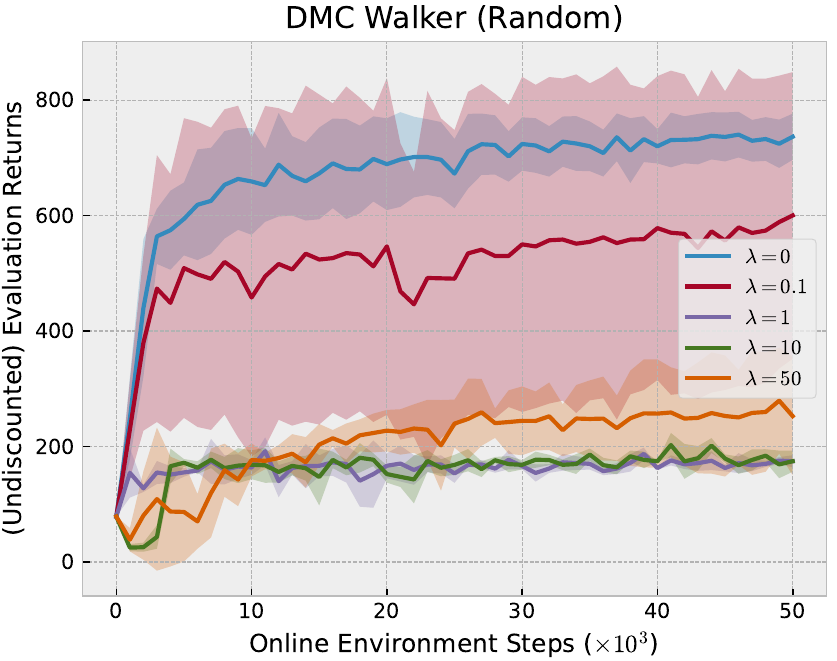}
    \end{subfigure}
    \caption{Undiscounted evaluation returns in Halfcheetah (Random) (left) and DMC Walker (Random) (right) for $\lambda \in \{0, 0.1, 1, 10, 50\}$ intrinsic-reward weights throughout online fine-tuning.}
    \label{fig:rnd-alone}
\end{figure*}

Figure~\ref{fig:rnd-alone} reports the average (bold) $\pm$ one standard deviation (shaded area) across five seeds. We note that when $\lambda$ is relatively small in Halfcheetah (Random), the agents perform roughly the same as when no exploration guidance is used (i.e., $\lambda=0$). In contrast, a relatively large $\lambda$ causes the agents to lose their pretrained initialization, as shown by the dramatic drop in evaluation returns at the beginning of online fine-tuning. Our hypothesis is also confirmed in DMC Walker (Random).


In order to overcome the issue of unlearned offline initializations, we propose using two agents: one for exploitation and one for exploration. Such a framework has been shown to improve learning stability in Decoupled RL (DeRL)~\citep{derl}. Both agents can be initialized with offline pretraining, but the exploitation agent only receives $r^e$, while the exploration agent receives $r^e + \lambda r^i$ during online fine-tuning. We only care about the exploitation agent for evaluation purposes and rely on the exploration agent for data collection. This strategy allows the exploitation agent to avoid the performance collapse shown in Figure~\ref{fig:rnd-alone} while also potentially benefiting from guided exploration. We refer to this agent as RND/DeRL in our main experiments.

\subsection{Upper Confidence Bound Exploration}\label{sec:ucb-methods}

Many recent implementations of UCB-style algorithms use ensembles of Q-functions to select actions $a_t$ at timestep $t$ according to a mixture of value and uncertainty: $a_t = \argmax_a Q_{\textrm{mean}}(s_t,a) + \lambda Q_{\textrm{std}}(s_t,a)$ (e.g.,~\citet{meanq}  and~\citet{emax}). Here, uncertainty $Q_{\textrm{std}}$ is measured as the standard deviation of Q-values over ensemble members for each action in the discrete-action case, or for a set of sampled actions in the continuous-action case (e.g.,~\citet{sunrise}). 

However, in the OtO setting, it is not clear whether it is better to guide exploration with value uncertainty or the uncertainty in another learned component. For example, when using MBPO+SAC, we could use an ensemble of transition functions, reward functions, value functions, or policies for the uncertainty computation. Given that these components are trained with different targets and update types (e.g., Bellman backups versus value and entropy maximization), can we reasonably expect the uncertainty of each component to drive exploration into the same regions of the state-action space during online fine-tuning?




To answer this question, we first train an MBPO+SAC agent with ensembles of all four previously-mentioned components on the Halfcheetah (Random) dataset and evaluate their uncertainties on 2,500 transition tuples from the Halfcheetah (Expert) dataset. We evaluate the ensembles' uncertainty on a dataset collected by an expert behavior policy, as it is likely to contain out-of-distribution tuples relative to the (Random) dataset, which is where we ultimately care about evaluating uncertainty in the OtO setting. We repeat this exercise with datasets from the Hopper environment from D4RL.\footnote{See Appendix~\ref{app:arch-and-hypers} for more details.} If uncertainty is the same across all learned components, then the order in which they rank the expert tuples in terms of uncertainty should be similar.  Table~\ref{tab:hc-uncertainty-comparison} shows Spearman's rho between the learned components uncertainty rankings of the tuples from the (Expert) dataset. We color cells in green when $\rho \geq 0.4$ and in red when $\rho \leq -0.4$ for ease of reading. 

\begin{table}[!t]
    \resizebox{\textwidth}{!}{
        \begin{tabular}{c|c|c|c|c}
        \label{tab:uncertainty}
         \cellcolor{gray!50} & Reward & Value & Transition & Policy  \\ \hline
         Reward & \cellcolor{gray!50} & -0.26 & 0.20 & 0.15  \\
         Value & \cellcolor{gray!50} & \cellcolor{gray!50} & \cellcolor{green!50} 0.55 & \cellcolor{red!50} -0.41 \\
         Transition & \cellcolor{gray!50} & \cellcolor{gray!50} & \cellcolor{gray!50} & 0.08 \\
         Policy & \cellcolor{gray!50} & \cellcolor{gray!50} & \cellcolor{gray!50} & \cellcolor{gray!50} 
        \end{tabular}
        \quad
        \begin{tabular}{c|c|c|c|c}
         \cellcolor{gray!50} & Reward & Value & Transition & Policy  \\ \hline
         Reward & \cellcolor{gray!50} & -0.13 & \cellcolor{green!50} 0.54 & 0.33  \\
         Value & \cellcolor{gray!50} & \cellcolor{gray!50} & \cellcolor{red!50} -0.57 & \cellcolor{red!50} -0.67 \\
         Transition & \cellcolor{gray!50} & \cellcolor{gray!50} & \cellcolor{gray!50} & \cellcolor{green!50} 0.53 \\
         Policy & \cellcolor{gray!50} & \cellcolor{gray!50} & \cellcolor{gray!50} & \cellcolor{gray!50} 
        \end{tabular}
    }
   \caption{Pair-wise rank correlation (Spearman's rho) between different ensembles' uncertainty in Halfcheetah (left) and Hopper (right). We color cells in green when $\rho \geq 0.4$ and in red when $\rho \leq -0.4$ for ease of reading.}
   \label{tab:hc-uncertainty-comparison}
\end{table}

We highlight that the rank correlation varies greatly. In some cases, two ensembles agree strongly (e.g., Value and Transition in Halfcheetah); in others, they disagree strongly (e.g., Value and Policy in Hopper) or show no relation (e.g., Transition and Policy in Halfcheetah). There is not necessarily a pattern that holds between the two environments. Hence, swapping learned components into the UCB action-selection equation would likely not result in similar data-collection behavior. This inconsistency is a potential issue because the limited budget of interactions in OtO RL makes data-collection strategy paramount. While methods such as intrinsic motivation have the clear strategy of guiding the policy towards previously-unseen areas of the MDP, there is no clear reason why we should prefer to reduce the uncertainty in one learned component versus any other using a UCB method in OtO RL. Instead of devising a complex and adaptive UCB method that balances the uncertainty of all learned components in this work, we evaluate one baseline that uses value-driven UCB (UCB(Q)) and one that uses dynamics-driven UCB (UCB(T)) in our main experiments.


\section{Planning to Go Out-of-Distribution}
The exploration methods we examined in \S\ref{sec:explore-in-oto} are lacking in two respects when considering the OtO setting. First, intrinsic reward methods use a moving-target reward function which can cause value functions to unlearn their offline pretraining, leading to instabilities in policy training. Second, UCB methods are myopic and there is no clear data-collection strategy in terms of which ensemble to use for exploration. This leads us to propose PTGOOD, a planning paradigm that overcomes and avoids these issues.

We posit that data collected during online fine-tuning in the OtO setting should meet two criteria: (1) be non-redundant to data in the offline dataset and (2) be of relatively high reward.  Violating criterion (1) would result in wasted interactions, as no new information would be gained. The importance of criterion (2) is highlighted by OtO RL's agent-environment interaction budget. As an exhaustive exploration of the MDP is likely impossible under  this budget, we should prioritize data-collection in portions of the state-action space that a well-performing policy would traverse. These regions are likely to satisfy criterion (2).

PTGOOD satisfies criterion (1) via a multi-step (i.e., non-myopic) planning procedure that maximizes the likelihood of collecting transition tuples that are out-of-distribution relative to the offline dataset. PTGOOD first estimates $\rho_{\pi_b}$, the occupancy measure (defined in \S\ref{sec:bg}) for policy $\pi_b$  via the Conditional Entropy Bottleneck (CEB)~\citep{ceb}. This estimate allows PTGOOD to infer the likelihood of $\pi_b$ executing a given action in a given state. PTGOOD satisfies criterion (2) by ensuring that the exploration guidance does not stray too far from the policy being fine-tuned. This is accomplished by sampling the policy and adding a small amount of noise during planning. As RL policy updates target high-reward regions in the vicinity of the current policy, exploring ``close'' to the improving policy should naturally target increasingly higher-reward regions. The notion and importance of closeness is explored in \S\ref{sec:planning-noise}. In the following subsections, we describe how PTGOOD uses the CEB to learn representations, the metric that PTGOOD targets during planning, and the planning algorithm itself.



\subsection{Conditional Entropy Bottleneck}
PTGOOD uses the CEB to estimate $\rho_{\pi_b}$ using samples from the offline dataset. The CEB is an information-theoretic method for learning a representation $Z$ of input data $X$ useful for predicting target data $Y$. CEB's simplest formulation is to learn a $Z$ that minimizes $\beta I(X;Z \vert Y) - I(Z;Y)$, where $\beta$ is a weighting hyperparameter and $I(\cdot)$ denotes mutual information. Intuitively, CEB learns a representation that minimizes the extra information $Z$ captures about $X$ when $Y$ is known and maximizes the information $Z$ captures about $Y$. While the CEB has many different forms, we use the contrastive ``CatGen'' formulation as described by~\citet{ceb} with the following upper bound:

\begin{equation}\label{eqn:catgen-ceb-mainbody}
    \begin{aligned}
        \mathrm{CEB_{CatGen}} \leq \mathrm{min}_{e(\cdot), b(\cdot)} \: &\mathbb{E} \left[  \mathbb{E}_{z_X \sim e(z_X \vert x)} [ \beta \: \mathrm{log} \: \frac{e(z_X \vert x)}{b(z_X \vert x^{\prime})} - \mathrm{log} \: \frac{b(z_X \vert x^{\prime})}{\frac{1}{K} \sum_{i=1}^K b(z_X \vert x^{\prime}_i)} ] \right. \\
        &\left. + \mathbb{E}_{z_{X^{\prime}} \sim b(z_{X^{\prime}} \vert x^{\prime})} [ \beta \: \mathrm{log} \: \frac{b(z_{X^{\prime}} \vert x^{\prime})}{e(z_{X^{\prime}} \vert x)} - \mathrm{log} \: \frac{e(z_{X^{\prime}} \vert x)}{\frac{1}{K} \sum_{i=1}^K e(z_{X^{\prime}} \vert x_i)} ] \right],
    \end{aligned}
\end{equation}
where the outer expectation is over the joint distribution $x,x^{\prime} \sim p(x, x^{\prime}, u, z_X, z_{X^{\prime}})$, $x$ is a state-action pair, $x^{\prime}$ is a state-action pair with a small amount of multiplicative noise drawn from a uniform distribution $u \sim U(0.99, 1.01)$: $x^{\prime} = u \odot x$, $e(\cdot)$ is the encoder, and $b(\cdot)$ is the backwards encoder. For more details, we refer the reader to Appendix~\ref{app:ceb} and the original paper.

\subsection{The Rate $\mathcal{R}$ and Modeling $\rho_{\pi_b}$}
PTGOOD uses the \emph{rate}~\citep{broken-elbo,uncert-vib} to measure how out-of-distribution a sample is relative to $\rho_{\pi_b}$. Rate has been used successfully in computer vision as a thresholding tool for out-of-distribution detection and has been shown to work well with the CEB representations that we use here~\citep{ceb}.


We first fit an encoder $e(z_X \vert x)$ and backward encoder $b(z_{X^\prime} \vert x^\prime)$ to a latent space $Z$ with Equation~\ref{eqn:catgen-ceb-mainbody} and state-action pairs sampled uniformly at random from the offline dataset. Next, we learn a marginal $m(z_X)$ of our training data in the representation space of the encoder $e(\cdot)$ as a mixture of Gaussians. See Appendix~\ref{app:arch-and-hypers} for more details. Given this encoder conditional density $e$, and marginal $m$, the \textit{rate} of a given state-action pair $x$ is computed as:
\begin{equation}\label{eqn:rate}
\mathcal{R}(x) \triangleq \mathrm{log}\;e(z_X \vert x) - \mathrm{log}\;m(z_X).
\end{equation}
In short, the representation produced by the encoder $z_X \sim e(\cdot \vert x)$ for an out-of-distribution $x$ should be highly unlikely according to $m(\cdot)$, thereby producing a rate value much larger than for an in-distribution $x$. Ultimately, this allows PTGOOD to estimate the likelihood of a given state-action pair being collected by $\pi_b$.

\subsection{PTGOOD}
PTGOOD is a planning paradigm designed to leverage offline pretraining to maximize the benefit of online data-collection. PTGOOD can be applied in combination with any OtO RL method that uses a dynamics model. Given a learnt offline policy and dynamics model, PTGOOD plans the data collection process one step at a time to collect the next transition tuple, which then augments the offline data and all data collected so far. The policy can now be updated with the new data. The data-collection planning process can then be repeated as many times as our budget of online interactions allows. 


The planning part of this process is given in Algorithm~\ref{alg:ptgood-planning} in Appendix~\ref{app:ptgood-pseudocode}. PTGOOD's planning procedure has a width $w$ and a depth $d$. Starting from a given state $s$, we sample the policy $w$ times and add a small amount of randomly-sampled Gaussian noise $\mathcal{N}(0,\epsilon)$ with variance hyperparameter $\epsilon$ to the actions. Then, the learned dynamics model $\hat{\mathcal{T}}$ predicts one step forward from state $s$ for each $w$ actions, and action sampling is repeated with each new state. The sampling and forward-step process is repeated $d$ times, forming a tree of state-nodes connected by action-branches of possible paths from the original state $s$. For each state-node and action-branch associated with that state-node in the tree beyond the original state $s$, PTGOOD computes the rate per Equation~\ref{eqn:rate}. 

\begin{figure}[!t]
    \begin{center}
        \includegraphics[width=0.995\textwidth]{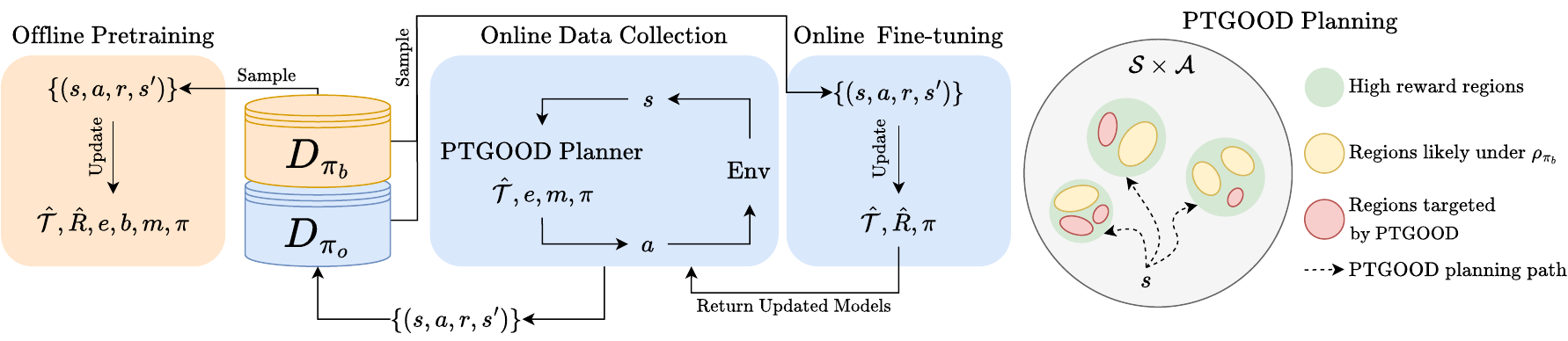}
    \end{center}
    \caption{Offline (orange) and online (blue) components in OtO RL, with PTGOOD planning shown on the far right. During offline pre-training, dynamics $\hat{\mathcal{T}}$, reward $\hat{R}$, encoder $e$, backward encoder $b$, marginal $m$, and policy $\pi$ (and other agent-related networks, depending on algorithm) are trained with data from $D_{\pi_b}$. During the online data-collection phase, PTGOOD's planner interacts with the environment using $\hat{\mathcal{T}}, e, m, \pi$, and stores data in $D_{\pi_o}$. Interleaved with data collection is fine-tuning, which occurs with data sampled from both $D_{\pi_b}$ and $D_{\pi_o}$. As shown on the right,  PTGOOD's planning procedure follows the improving policy $\pi$ from a given $s$ towards increasingly higher reward regions of the $\mathcal{S} \times \mathcal{A}$ space, and targets data in those spaces that are unlikely under $\rho_{\pi_b}$.}
    \label{fig:oto-plus-ptgood}
\end{figure}

After the tree is fully formed, PTGOOD traverses the tree in reverse, summing the rates associated to each state-node back to the original $w$ actions in the original state $s$. Finally, PTGOOD returns the action from the set of original $w$ actions associated with the highest rate sum. This action is then executed and the MDP steps forward to a new state. See Figure~\ref{fig:oto-plus-ptgood} for a depiction of the two phases of OtO RL and PTGOOD's planning procedure.

 \section{Experiments}
In our experiments, we aim to answer the following questions: (1) Can PTGOOD improve agent evaluation returns within the given agent-environment interaction budget in the online fine-tuning phase? (2) How important is guided exploration to agent evaluation returns during online fine-tuning? (3) Are the policy-constraint mechanisms that are important in the purely-offline setting important in the OtO setting? 

\subsection{Baselines}
We carefully design baselines that reflect prominent categories of exploration strategies in RL (\S\ref{sec:explore-in-oto}). We tune each of our baselines on a per-environment per-dataset basis and report results for the best-performing hyperparameters for each method. Below we briefly list and describe the baselines we benchmark against PTGOOD. Unless otherwise noted, all algorithms use MBPO+SAC as the core model-based RL algorithm. See Appendix~\ref{app:baseline-tuning} for more details and results.

The \textbf{No Pretrain} baseline does not perform offline pretraining, but does use both the offline dataset and data collected online for online training. The \textbf{Naive} baseline performs offline pretraining and online fine-tuning, but only samples the policy to choose actions during online fine-tuning instead of using exploration methods. The Naive agent contextualizes the added benefit of guided exploration. We use the \textbf{RND/DeRL} baseline as described in \S\ref{sec:intrinsic-rewards}. We train the RND predictor using the offline dataset before online fine-tuning begins and periodically update the predictor's weights throughout the fine-tuning process. We also use the \textbf{UCB(Q)} and \textbf{UCB(T)} baselines described in \S\ref{sec:ucb-methods}. \textbf{Cal-QL}~\citep{cal-ql} is a model-free OtO algorithm built on top of CQL~\citep{cql}, a pessimistic offline RL algorithm. Cal-QL corrects for instabilities during online fine-tuning induced by CQL's value constraint. Finally, we benchmark \textbf{PROTO}~\citep{proto} and \textbf{PEX}~\citep{pex}, model-free methods designed for the OtO setting. PROTO uses a trust-region update on top of EQL~\citep{eq3} and TD3, and PEX learns a set of policies for action selection on top of IQL~\citep{iql}. None of the agents except for Cal-QL, PROTO, and PEX use conservatism or pessimism of any form during any stage of training. See Appendix~\ref{app:arch-and-hypers} for architecture and hyperparameter details along with full implementation details for PTGOOD.

\subsection{Environments and Datasets}
We evaluate PTGOOD and our baselines on a set of environment-dataset combinations that satisfy two criteria: (a) it must not be possible for current algorithms to learn an optimal policy during the offline pretraining phase, and (b) we must be able to surpass a random agent during offline pretraining. If criterion (a) is violated, there is no need for online fine-tuning. If criterion (b) is violated, then the offline pretraining phase is not useful, and training from scratch online (i.e., No Pretrain) would be unlikely to be beaten.\footnote{We show empirically in Appendix~\ref{app:cal-ql-study} that this is indeed the case.} We use datasets in the Halfcheetah and Hopper environments from the D4RL study. Additionally, we collect our own datasets from environments not represented in D4RL, including Ant, Humanoid, and the Walker task from the DeepMind Control Suite (DMC). The datasets that we collect follow the same dataset design principles of D4RL. See Appendix~\ref{app:envs} for more details on our environments and datasets.

\subsection{OtO Results}\label{sec:oto-results}
For each environment-dataset combination, we first pretrain agents offline to convergence and then fine-tune online for 50k environment steps across five seeds. Every 1k environment steps, we collect undiscounted returns across 10 evaluation episodes. Reporting comparative results between RL algorithms is a complex problem~\citep{empirical-rl}; therefore, we present results across various views and mediums. Table~\ref{tab:oto-results} shows the average $\pm$ one standard deviation of evaluation returns at the 50k online-steps mark with the highest returns bolded. We highlight in blue when the highest returns are statistically significantly different via a two-sided Welch's t-test. Figure~\ref{fig:all-algos-full-results} displays undiscounted evaluation return curves for all algorithms in all environment-dataset combinations across the 50k online fine-tuning steps. Figure~\ref{fig:first-second-compare} displays undiscounted evaluation return curves in all five training runs for the best and second-best performing algorithms in each environment-dataset combination.

\begin{table}[!t]
    \centering
    \resizebox{\textwidth}{!}{
    \begin{tabular}{l||ccccccc}
         Algorithm & Halfcheetah (R) & DMC Walker (R) & Hopper (R) & Ant (R) & DMC Walker (MR) & Ant (MR) & Humanoid (MR) \\ \hline
         PTGOOD & \cellcolor{cyan!25}\textbf{8867 $\pm$ 88} & \cellcolor{cyan!25}\textbf{959 $\pm$ 8} & \cellcolor{cyan!25}\textbf{3246 $\pm$ 123} & \textbf{5624 $\pm$ 235} & \cellcolor{cyan!25}\textbf{953 $\pm$ 6} & \cellcolor{cyan!25}\textbf{5866 $\pm$ 114} & \textbf{15050 $\pm$ 878} \\
         No Pretrain & 7249 $\pm$ 814 & 668 $\pm$ 88 & 1231 $\pm$ 648 & 3703 $\pm$ 901 & 778 $\pm$ 93 & 4777 $\pm$ 1085 & 10723 $\pm$ 3903 \\
         Naive & 7434 $\pm$ 782 & 736 $\pm$ 40 & 1576 $\pm$ 880 & 4663 $\pm$ 626 & 732 $\pm$ 21 & 4973 $\pm$ 337 & 11706 $\pm$ 3403 \\
         RND/DeRL & 6782 $\pm$ 2013 & 677 $\pm$ 63 & 1818 $\pm$ 786 & 5258 $\pm$ 191 & 700 $\pm$ 164 & 4836 $\pm$ 695 & 1954 $\pm$ 1199 \\
         UCB(Q) & 7300 $\pm$ 861 & 740 $\pm$ 50 & 2037 $\pm$ 382 & 5290 $\pm$ 272 & 783 $\pm$ 75 & 5328 $\pm$ 224 & 13183 $\pm$ 885 \\
         UCB(T) & 8170 $\pm$ 513 & 811 $\pm$ 68 & 2251 $\pm$ 830 & 5022 $\pm$ 299 & 772 $\pm$ 93 & 4509 $\pm$ 1364 & 12079 $\pm$ 2461 \\
         Cal-QL & -315 $\pm$ 122 & 45 $\pm$ 4 & 57 $\pm$ 39 & -309 $\pm$ 575 & 106 $\pm$ 57 & 990 $\pm$ 864 & 381 $\pm$ 174 \\
         PROTO & 7877 $\pm$ 703 & 583 $\pm$ 282 & 511 $\pm$ 298 & 1174 $\pm$ 291 & 874 $\pm$ 66 & 1696 $\pm$ 595 & 696 $\pm$ 120 \\
         PEX & 4953 $\pm$ 454 & 83 $\pm$ 21 & 1889 $\pm$ 951 & 1436 $\pm$ 482 & 541 $\pm$ 65 & 2960 $\pm$ 119 & 8320 $\pm$ 4187 \\
    \end{tabular}
    }
    \caption{Average $\pm$ one standard deviation of undiscounted evaluation returns after 50k environment steps of online fine-tuning. Highest returns per algorithm-dataset combination bolded. Statistical significance is shown with blue highlight. (R)=Random and (MR)=Medium Replay.}
    \label{tab:oto-results}
\end{table}

First, we answer question (1) in the affirmative by highlighting that PTGOOD consistently provides the strongest performance across all environment-dataset combinations. Table~\ref{tab:oto-results} shows that PTGOOD provides the highest returns in $7/7$ environment-dataset combinations, which are statistically significant in $5/7$. Figure~\ref{fig:all-algos-full-results} shows that PTGOOD is generally stable relative to other baselines (e.g., RND/DeRL in Halfcheetah (Random)). We also note that PTGOOD tends to avoid the premature policy convergence that other methods sometimes exhibit (e.g., DMC Walker (Random), DMC Walker (Medium Replay), and Hopper (Random) in Figure~\ref{fig:first-second-compare}). See Appendix~\ref{app:subopt-convergence} for more analysis. Aside from higher returns after training has finished, PTGOOD often outperforms other baselines during the middle portions of fine-tuning (e.g., Halfcheetah (Random) and Ant (Medium Replay) in Figure~\ref{fig:first-second-compare}).

Second, we address question (2). We note that the Naive method is a strong baseline across all environment-dataset combinations that we tested. Additionally, we highlight that the Naive baseline outperforms some guided-exploration baselines on occasion (e.g., RND/DeRL in Halfcheetah (Random) and UCB(T) in Ant (Medium Replay)). These results suggest that certain types of exploration are not universally helpful in OtO RL.

Third, we answer question (3) by observing Cal-QL results in Table~\ref{tab:oto-results} and training curves in Figure~\ref{fig:all-algos-full-results}. We note that Cal-QL performs poorly consistently. This is unsurprising because Cal-QL's base algorithm encourages the learned policy to remain close to the behavior policy. Due to our environment-dataset selection criteria, the behavior policies are highly suboptimal, which makes conservatism and pessimism an unideal choice. We investigate Cal-QL's poor performance further in Appendix~\ref{app:cal-ql-study} by training it for two million online steps in all environment-dataset combinations. In short, we find that Cal-QL does not learn anything useful in any Random dataset nor in Humanoid (Medium Replay), but it does learn a good policy in the remaining Medium Replay datasets at the end of the two million online steps. In contrast, PTGOOD is able to find the optimal policy in less than 50k online steps in all environment-dataset combinations.


Finally, we note that neither UCB type is consistently better than the other. Additionally, in some environment-dataset combinations, either method is outperformed by the Naive baseline (e.g., in Halfcheetah (Random) for UCB(Q) and Ant (Medium Replay) for UCB(T)). This evidence, when combined with our experiment in \S\ref{sec:ucb-methods}, suggests that further research in multi-ensemble UCB exploration could prove fruitful.

\subsection{Planning Noise}\label{sec:planning-noise}
\begin{figure*}[!t]
    \centering
    \begin{subfigure}{0.45\textwidth}
        \includegraphics[width=\textwidth]{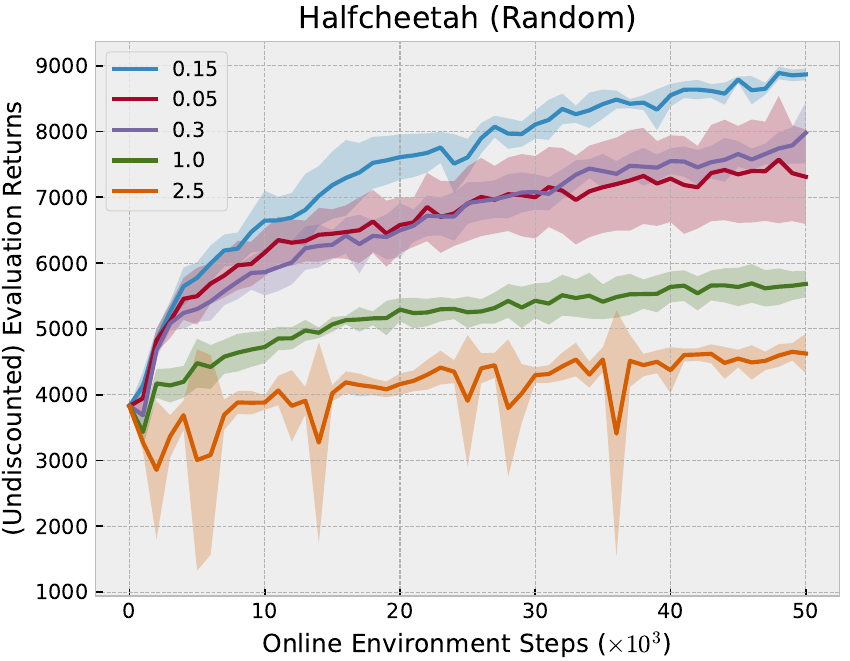}
    \end{subfigure}
    \begin{subfigure}{0.45\textwidth}
        \includegraphics[width=\textwidth]{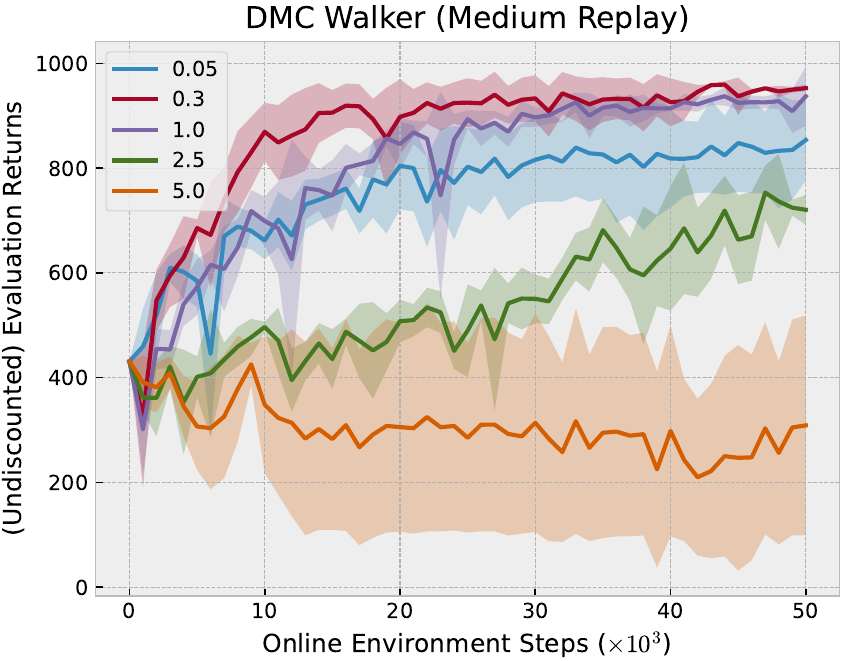}
    \end{subfigure}
    \caption{Average (bold line) $\pm$ one standard deviation (shaded area) of evaluation returns for different $\epsilon$ values in PTGOOD's planner in Halfcheetah (Random) (left) and DMC Walker (Medium Replay) (right).}
    \label{fig:rnd/noise-exp}
\end{figure*}

Key to PTGOOD is exploring both unknown and high-reward regions of the state-action space. Instead of targeting high-reward state-action pairs with a Q-function value estimate, PTGOOD remains ``close'' to the improving policy by adding a small amount of noise to actions during the planning process. Using noise instead of explicit value estimation has computational benefits (see Appendix~\ref{app:compute-cost-compare}) and does not rely on values that may be overestimated due to distributional shift~\citep{overest-fujimoto-1,overest-fujimoto-2}.

The meanings of ``far'' and ``close'' in the context of action selection are likely to be environment-dependent. We perform a sweep over $\epsilon$ values in all environment-dataset combinations. Figure~\ref{fig:rnd/noise-exp} shows the average $\pm$ one standard deviation of undiscounted evaluation returns for Halfcheetah (Random) and DMC Walker (Medium Replay) for various noise levels. We note that there is an optimal noise hyperparameter in either environment. If $\epsilon$ is too small, evaluation returns degrade slightly due to the reduced exploration. If $\epsilon$ grows too large, PTGOOD's exploration strays too far from the improving policy and may become close to random exploration, which produces significantly reduced evaluation returns. We perform this exercise for all other environment-dataset combinations in Appendix~\ref{app:planning-noise-extended}, and find the same pattern.

\section{Conclusion}
In this work, we studied the OtO setting from the exploration perspective. First, we examined intrinsic motivation and UCB exploration from the lens of OtO RL, identifying compatibility issues and other shortcomings. Then, we introduced PTGOOD, a planning paradigm for model-based RL algorithms for exploration in the OtO setting. PTGOOD uses an estimate of the behavior policy's occupancy measure within a non-myopic planner to target high-reward state-action pairs unrepresented in the offline dataset. We demonstrated in diverse continuous-control tasks that PTGOOD consistently provides the highest returns and avoids suboptimal policy convergence. PTGOOD could be improved further with adaptive noise in the planning process, which could account for state-dependent exploration noise or action-space characteristics (e.g., different joint types in musculoskeletal control).

\bibliography{main}

\begin{thebibliography}{56}
\providecommand{\natexlab}[1]{#1}
\providecommand{\url}[1]{\texttt{#1}}
\expandafter\ifx\csname urlstyle\endcsname\relax
  \providecommand{\doi}[1]{doi: #1}\else
  \providecommand{\doi}{doi: \begingroup \urlstyle{rm}\Url}\fi

\bibitem[Alemi et~al.(2018{\natexlab{a}})Alemi, Poole, Fischer, Dillon, Saurous, and Murphy]{broken-elbo}
Alexander Alemi, Ben Poole, Ian Fischer, Joshua Dillon, Rif~A. Saurous, and Kevin Murphy.
\newblock Fixing a broken {ELBO}.
\newblock In \emph{Proceedings of the 35th International Conference on Machine Learning}, 2018{\natexlab{a}}.

\bibitem[Alemi et~al.(2018{\natexlab{b}})Alemi, Fischer, and Dillon]{uncert-vib}
Alexander~A. Alemi, Ian Fischer, and Joshua~V. Dillon.
\newblock Uncertainty in the variational information bottleneck.
\newblock \emph{arXiv preprint: arXiv:1807.00906}, 2018{\natexlab{b}}.

\bibitem[Auer(2002)]{ucb-original}
Peter Auer.
\newblock Using confidence bounds for exploitation-exploration trade-offs.
\newblock \emph{Journal of Machine Learning Research}, 3:\penalty0 397--422, 2002.

\bibitem[Ball et~al.(2023)Ball, Smith, Kostrikov, and Levine]{rlpd}
Philip~J. Ball, Laura Smith, Ilya Kostrikov, and Sergey Levine.
\newblock Efficient online reinforcement learning with offline data.
\newblock In \emph{Proceedings of the 40th International Conference on Machine Learning (ICML)}, 2023.

\bibitem[Beeson \& Montana(2022)Beeson and Montana]{td3bc-improvement}
Alex Beeson and Giovanni Montana.
\newblock Improving td3-bc: Relaxed policy constraint for offline learning and stable online fine-tuning.
\newblock In \emph{3rd Offline Reinforcement Learning Workshop at Neural Information Processing Systems}, 2022.

\bibitem[Burda et~al.(2019)Burda, Edwards, Storkey, and Klimov]{rnd}
Yuri Burda, Harrison Edwards, Amos Storkey, and Oleg Klimov.
\newblock Exploration by random network distillation.
\newblock In \emph{International Conference on Learning Representations (ICLR)}, 2019.

\bibitem[Chen et~al.(2017)Chen, Sidor, Abbeel, and Schulman]{orig-ucb-ens}
Richard~Y. Chen, Szymon Sidor, Pieter Abbeel, and John Schulman.
\newblock Ucb exploration via q-ensembles.
\newblock \emph{arXiv preprint: arXiv:1706.01502}, 2017.

\bibitem[Chentanez et~al.(2004)Chentanez, Barto, and Singh]{intrinsic-orig}
Nuttapong Chentanez, Andrew Barto, and Satinder Singh.
\newblock Intrinsically motivated reinforcement learning.
\newblock In \emph{Neural Information Processing Systems}, 2004.

\bibitem[Ernst et~al.(2005)Ernst, Geurts, and Wehenke]{batch-rl1}
Damien Ernst, Pierre Geurts, and Louis Wehenke.
\newblock Tree-based batch mode reinforcement learning.
\newblock \emph{Journal of Machine Learning Research}, 6:\penalty0 503--556, 2005.

\bibitem[Fischer(2020)]{ceb}
Ian Fischer.
\newblock The conditional entropy bottleneck.
\newblock \emph{Entropy}, 2020.

\bibitem[Fu et~al.(2020)Fu, Kumar, Nachum, Tucker, and Levine]{d4rl}
Justin Fu, Aviral Kumar, Ofir Nachum, George Tucker, and Sergey Levine.
\newblock D4rl: Datasets for deep data-driven reinforcement learning, 2020.

\bibitem[Fujimoto \& Gu(2021)Fujimoto and Gu]{minimalist}
Scott Fujimoto and Shixiang~Shane Gu.
\newblock A minimalist approach to offline reinforcement learning.
\newblock In \emph{35th Conference on Neural Information Processing Systems (NeurIPS)}, 2021.

\bibitem[Fujimoto et~al.(2018)Fujimoto, van Hoof, and Meger]{overest-fujimoto-1}
Scott Fujimoto, Herke van Hoof, and David Meger.
\newblock Addressing function approximation error in actor-critic methods.
\newblock In \emph{Proceedings of the 35th International Conference on Machine Learning}, 2018.

\bibitem[Fujimoto et~al.(2019)Fujimoto, Meger, and Precup]{overest-fujimoto-2}
Scott Fujimoto, David Meger, and Doina Precup.
\newblock Off-policy deep reinforcement learning without exploration.
\newblock In \emph{Proceedings of the 36th International Conference on Machine Learning}, 2019.

\bibitem[Haarnoja et~al.(2017)Haarnoja, Zhou, Abbeel, and Levine]{sac}
Tuomas Haarnoja, Aurick Zhou, Pieter Abbeel, and Sergey Levine.
\newblock Soft actor-critic: Off-policy maximum entropy deep reinforcement learning with a stochastic actor.
\newblock 2017.

\bibitem[Henaff(2019)]{t-disagreement}
Mikael Henaff.
\newblock Explicit explore-exploit algorithms in continuous state spaces.
\newblock In \emph{Neural Information Processing Systems (NeurIPS),}, 2019.

\bibitem[Hester et~al.(2018)Hester, Vecerik, Pietquin, Lanctot, Schaul, Piot, Horgan, Quan, Sendonaris, Dulac-Arnold, Osband, Agapiou, Leibo, and Gruslys]{lfd2}
Todd Hester, Matej Vecerik, Olivier Pietquin, Marc Lanctot, Tom Schaul, Bilal Piot, Dan Horgan, John Quan, Andrew Sendonaris, Gabriel Dulac-Arnold, Ian Osband, John Agapiou, Joel~Z. Leibo, and Audrunas Gruslys.
\newblock Deep q-learning from demonstrations.
\newblock In \emph{AAAI}, 2018.

\bibitem[Janner et~al.(2019)Janner, Fu, Zhang, and Levine]{mbpo}
Michael Janner, Justin Fu, Marvin Zhang, and Sergey Levine.
\newblock When to trust your model: Model-based policy optimization.
\newblock In \emph{Advances in Neural Information Processing Systems}, 2019.

\bibitem[Kang et~al.(2018)Kang, Jie, and Feng]{occupancy-measure-unique2}
Bingyi Kang, Zequn Jie, and Jiashi Feng.
\newblock Policy optimization with demonstrations.
\newblock In \emph{Proceedings of the 35th International Conference on Machine Learning (ICML)}, 2018.

\bibitem[Kidambi et~al.(2020)Kidambi, Aravind~Rajeswaran, and Joachims]{morel}
Rahul Kidambi, Praneeth~Netrapalli Aravind~Rajeswaran, and Thorsten Joachims.
\newblock Morel : Model-based offline reinforcement learning.
\newblock In \emph{34th Conference on Neural Information Processing Systems (NeurIPS)}, 2020.

\bibitem[Kostrikov et~al.(2021)Kostrikov, Tompson, Fergus, and Nachum]{fisher-brc}
Ilya Kostrikov, Jonathan Tompson, Rob Fergus, and Ofir Nachum.
\newblock Offline reinforcement learning with fisher divergence critic regularization.
\newblock In \emph{Proceedings of the 38th International Conference on Machine Learning}, 2021.

\bibitem[Kostrikov et~al.(2022)Kostrikov, Nair, and Levine]{iql}
Ilya Kostrikov, Ashvin Nair, and Sergey Levine.
\newblock Offline reinforcement learning with implicit q-learning.
\newblock In \emph{International Conference on Learning Representations (ICLR)}, 2022.

\bibitem[Kumar et~al.(2020)Kumar, Zhou, Tucker, and Levine]{cql}
Aviral Kumar, Aurick Zhou, George Tucker, and Sergey Levine.
\newblock Conservative q-learning for offline reinforcement learning.
\newblock In \emph{34th Conference on Neural Information Processing Systems (NeurIPS)}, 2020.

\bibitem[Kumar et~al.(2021)Kumar, Fu, Tucker, and Levine]{bear}
Aviral Kumar, Justin Fu, George Tucker, and Sergey Levine.
\newblock Stabilizing off-policy q-learning via bootstrapping error reduction.
\newblock In \emph{35th Conference on Neural Information Processing Systems (NeurIPS)}, 2021.

\bibitem[Lee et~al.(2021{\natexlab{a}})Lee, Laskin, Srinivas, and Abbeel]{sunrise}
Kimin Lee, Michael Laskin, Aravind Srinivas, and Pieter Abbeel.
\newblock Sunrise: A simple unified framework for ensemble learning in deep reinforcement learning.
\newblock In \emph{Proceedings of the 38th International Conference on Machine Learning}, 2021{\natexlab{a}}.

\bibitem[Lee et~al.(2021{\natexlab{b}})Lee, Seo, Lee, Abbeel, and Shin]{oto2021}
Seunghyun Lee, Younggyo Seo, Kimin Lee, Pieter Abbeel, and Jinwoo Shin.
\newblock Offline-to-online reinforcement learning via balanced replay and pessimistic q-ensemble.
\newblock In \emph{Conference on Robot Learning (CoRL)}, 2021{\natexlab{b}}.

\bibitem[Levine et~al.(2020)Levine, Kumar, Tucker, and Fu]{offline-rl-levine}
Sergey Levine, Aviral Kumar, George Tucker, and Justin Fu.
\newblock Offline reinforcement learning: Tutorial, review, and perspectives on open problems.
\newblock \emph{arXiv preprint: arXiv:2005.01643v3}, 2020.

\bibitem[Li et~al.(2023{\natexlab{a}})Li, Zhan, Lee, Chi, and Chen]{tabular-oto}
Gen Li, Wenhao Zhan, Jason~D. Lee, Yuejie Chi, and Yuxin Chen.
\newblock Reward-agnostic fine-tuning: Provable statistical benefits of hybrid reinforcement learning.
\newblock \emph{arXiv preprint: arXiv:2305.10282}, 2023{\natexlab{a}}.

\bibitem[Li et~al.(2023{\natexlab{b}})Li, Hu, Xu, Liu, Zhan, and Zhang]{proto}
Jianxiong Li, Xiao Hu, Haoran Xu, Jingjing Liu, Xianyuan Zhan, and Ya-Qin Zhang.
\newblock {PROTO}: Iterative policy regularized offline-to-online reinforcement learning.
\newblock \emph{arXiv preprint: arXiv:2305.15669}, 2023{\natexlab{b}}.

\bibitem[Liang et~al.(2022)Liang, Xu, McAleer, Hu, Ihler, Abbeel, and Fox]{meanq}
Litian Liang, Yaosheng Xu, Stephen McAleer, Dailin Hu, Alexander Ihler, Pieter Abbeel, and Roy Fox.
\newblock Reducing variance in temporal-difference value estimation via ensemble of deep networks.
\newblock In \emph{International Conference on Machine Learning (ICML)}, 2022.

\bibitem[Luo et~al.(2023)Luo, Kay, Grefenstette, and Deisenroth]{td3-c}
Yicheng Luo, Jackie Kay, Edward Grefenstette, and Marc~Peter Deisenroth.
\newblock Finetuning from offline reinforcement learning: Challenges, trade-offs and practical solutions.
\newblock \emph{arXiv preprint: arXiv:2303.17396}, 2023.

\bibitem[Mao et~al.(2022)Mao, Wang, Wang, and Zhang]{moore}
Yihuan Mao, Chao Wang, Bin Wang, and Chongjie Zhang.
\newblock {MOORe}: Model-based offline-to-online reinforcement learning.
\newblock \emph{arXiv preprint: arXiv:2201.10070}, 2022.

\bibitem[Nachum et~al.(2019)Nachum, Dai, Kostrikov, Chow, Li, and Schuurmans]{nachum2019algaedice}
Ofir Nachum, Bo~Dai, Ilya Kostrikov, Yinlam Chow, Lihong Li, and Dale Schuurmans.
\newblock Algaedice: Policy gradient from arbitrary experience.
\newblock \emph{arXiv preprint arXiv:1912.02074}, 2019.

\bibitem[Nakamoto et~al.(2023)Nakamoto, Zhai, Singh, Mark, Ma, Finn, Kumar, and Levine]{cal-ql}
Mitsuhiko Nakamoto, Yuexiang Zhai, Anikait Singh, Max~Sobol Mark, Yi~Ma, Chelsea Finn, Aviral Kumar, and Sergey Levine.
\newblock Cal-ql: Calibrated offline rl pre-training for efficient online fine-tuning.
\newblock \emph{arXiv preprint arXiv:2303.05479}, 2023.

\bibitem[Osband et~al.(2016)Osband, Blundell, Pritzel, and Roy]{deep-explore}
Ian Osband, Charles Blundell, Alexander Pritzel, and Benjamin~Van Roy.
\newblock Deep exploration via bootstrapped dqn.
\newblock In \emph{Advances in Neural Information Processing (NeurIPS)}, 2016.

\bibitem[Ostrovski et~al.(2017)Ostrovski, Bellemare, van~den Oord, and Munos.]{density-count}
Georg Ostrovski, Marc~G Bellemare, Aäron van~den Oord, and Rémi Munos.
\newblock Count-based exploration with neural density models.
\newblock In \emph{International Conference on Machine Learning (ICML)}, 2017.

\bibitem[Pathak et~al.(2017)Pathak, Agrawal, Efros, and Darrell.]{icm}
Deepak Pathak, Pulkit Agrawal, Alexei~A Efros, and Trevor Darrell.
\newblock Curiosity-driven exploration by self-supervised prediction.
\newblock In \emph{International Conference on Machine Learning (ICML)}, 2017.

\bibitem[Patterson et~al.(2023)Patterson, Neumann, White, and White]{empirical-rl}
Andrew Patterson, Samuel Neumann, Martha White, and Adam White.
\newblock Empirical design in reinforcement learning.
\newblock \emph{arXiv preprint: arXiv:2304.01315}, 2023.

\bibitem[Peng et~al.(2019)Peng, Kumar, Zhang, and Levine]{awr}
Xue~Bin Peng, Aviral Kumar, Grace Zhang, and Sergey Levine.
\newblock Advantage-weighted regression: Simple and scalable off-policy reinforcement learning.
\newblock 2019.

\bibitem[Rajeswaran et~al.(2018)Rajeswaran, Kumar, Gupta, Vezzani, Schulman, Todorov, and Levine]{expert-demonstrations1}
Aravind Rajeswaran, Vikash Kumar, Abhishek Gupta, Giulia Vezzani, John Schulman, Emanuel Todorov, and Sergey Levine.
\newblock Learning complex dexterous manipulation with deep reinforcement learning and demonstrations.
\newblock In \emph{RSS}, 2018.

\bibitem[Reidmiller(2005)]{batch-rl2}
Martin Reidmiller.
\newblock Neural fitted q iteration–first experiences with a data efficient neural reinforcement learning method.
\newblock In \emph{European Conference on Machine Learning}, 2005.

\bibitem[Rigter et~al.(2022)Rigter, Lacerda, and Hawes]{rambo-rl}
Marc Rigter, Bruno Lacerda, and Nick Hawes.
\newblock Rambo-rl: Robust adversarial model-based offline reinforcement learning.
\newblock In \emph{Advances in Neural Information Processing Systems (NeurIPS)}, 2022.

\bibitem[Schaal(1996)]{lfd}
Stefan Schaal.
\newblock Learning from demonstration.
\newblock In \emph{Neural Information Processing Systems (NeurIPS)}, 1996.

\bibitem[Schäfer et~al.(2022)Schäfer, Christianos, Hanna, and Albrecht]{derl}
Lukas Schäfer, Filippos Christianos, Josiah~P. Hanna, and Stefano~V. Albrecht.
\newblock Decoupled reinforcement learning to stabilise intrinsically-motivated exploration.
\newblock In \emph{International Conference on Autonomous Agents and Multiagent Systems}, 2022.

\bibitem[Schäfer et~al.(2023)Schäfer, Slumbers, McAleer, Du, Albrecht, and Mguni]{emax}
Lukas Schäfer, Oliver Slumbers, Stephen McAleer, Yali Du, Stefano~V. Albrecht, and David Mguni.
\newblock Ensemble value functions for efficient exploration in multi-agent reinforcement learning.
\newblock \emph{arXiv preprint: arXiv:2302.03439}, 2023.

\bibitem[Sekar et~al.(2020)Sekar, Rybkin, Daniilidis, Abbeel, Hafner, and Pathak]{plan2explore}
Ramanan Sekar, Oleh Rybkin, Kostas Daniilidis, Pieter Abbeel, Danijar Hafner, and Deepak Pathak.
\newblock Planning to explore via self-supervised world models.
\newblock In \emph{International Conference on Machine Learning (ICML)}, 2020.

\bibitem[Shyam et~al.(2019)Shyam, Jaśkowski, and Gomez]{max-model-explore}
Pranav Shyam, Wojciech Jaśkowski, and Faustino Gomez.
\newblock Model-based active exploration.
\newblock In \emph{International Conference on Machine Learning (ICML)}, 2019.

\bibitem[Syed et~al.(2008)Syed, Bowling, and Schapire]{occupancy-measure-unique1}
Umar Syed, Michael Bowling, and Robert~E. Schapire.
\newblock Apprenticeship learning using linear programming.
\newblock In \emph{Proceedings of the 25th International Conference on Machine Learning (ICML)}, 2008.

\bibitem[Tassa et~al.(2018)Tassa, Doron, Muldal, Erez, Li, de~Las~Casas, Budden, Abdolmaleki, Merel, Lefrancq, Lillicrap, and Riedmiller]{dmcontrol-paper}
Yuval Tassa, Yotam Doron, Alistair Muldal, Tom Erez, Yazhe Li, Diego de~Las~Casas, David Budden, Abbas Abdolmaleki, Josh Merel, Andrew Lefrancq, Timothy Lillicrap, and Martin Riedmiller.
\newblock {DeepMind} control suite.
\newblock \emph{arXiv preprint arXiv:1801.00690}, 2018.

\bibitem[Tassa et~al.(2020)Tassa, Tunyasuvunakool, Muldal, Doron, Liu, Bohez, Merel, Erez, Lillicrap, and Heess]{dmcontrol-software}
Yuval Tassa, Saran Tunyasuvunakool, Alistair Muldal, Yotam Doron, Siqi Liu, Steven Bohez, Josh Merel, Tom Erez, Timothy Lillicrap, and Nicolas Heess.
\newblock dm\_control: Software and tasks for continuous control.
\newblock \emph{arXiv preprint arXiv:2006.12983}, 2020.

\bibitem[Vecerik et~al.(2017)Vecerik, Hester, Scholz, Wang, Pietquin, Piot, Heess, Rothörl, Lampe, and Riedmiller]{lfd3}
Mel Vecerik, Todd Hester, Jonathan Scholz, Fumin Wang, Olivier Pietquin, Bilal Piot, Nicolas Heess, Thomas Rothörl, Thomas Lampe, and Martin Riedmiller.
\newblock Leveraging demonstrations for deep reinforcement learning on robotics problems with sparse rewards.
\newblock \emph{arXiv preprint: arXiv:1707.08817}, 2017.

\bibitem[Wu et~al.(2019)Wu, Tucker, and Nachum]{brac}
Yifan Wu, George Tucker, and Ofir Nachum.
\newblock Behavior regularized offline reinforcement learning.
\newblock \emph{arXiv preprint: arXiv:1911.11361}, 2019.

\bibitem[Xu et~al.(2023)Xu, Jiang, Li, Yang, Wang, Chan, and Zhan]{eq3}
Haoran Xu, Li~Jiang, Jianxiong Li, Zhuoran Yang, Zhaoran Wang, Victor Wai~Kin Chan, and Xianyuan Zhan.
\newblock Offline rl with no ood actions: In-sample learning via implicit value regularization.
\newblock In \emph{International Conference on Learning Representations (ICLR)}, 2023.

\bibitem[Yu et~al.(2020)Yu, Thomas, Yu, Ermon, Zou, Levine, Finn, and Ma]{mopo}
Tianhe Yu, Garrett Thomas, Lantao Yu, Stefano Ermon, James Zou, Sergey Levine, Chelsea Finn, and Tengyu Ma.
\newblock Mopo: Model-based offline policy optimization.
\newblock In \emph{34th Conference on Neural Information Processing Systems (NeurIPS)}, 2020.

\bibitem[Zhang et~al.(2023)Zhang, Xu, and Yu]{pex}
Haichao Zhang, We~Xu, and Haonan Yu.
\newblock Policy expansion for bridging offline-to-online reinforcement learning.
\newblock In \emph{International Conference on Learning Representations (ICLR)}, 2023.

\bibitem[Zhong et~al.(2022)Zhong, Zhang, Sch\"afer, Albrecht, and Hanna]{zhong2022datacollection}
Rujie Zhong, Duohan Zhang, Lukas Sch\"afer, Stefano~V. Albrecht, and Josiah~P. Hanna.
\newblock Robust on-policy data collection for data efficient policy evaluation.
\newblock In \emph{Conference on Neural Information Processing Systems}, 2022.

\end{thebibliography}
\bibliographystyle{rlc}

\appendix

\section{Baselines}\label{app:baseline-tuning}

\begin{figure*}[!t]
    \centering
    \begin{subfigure}{0.32\textwidth}
        \includegraphics[width=\textwidth]{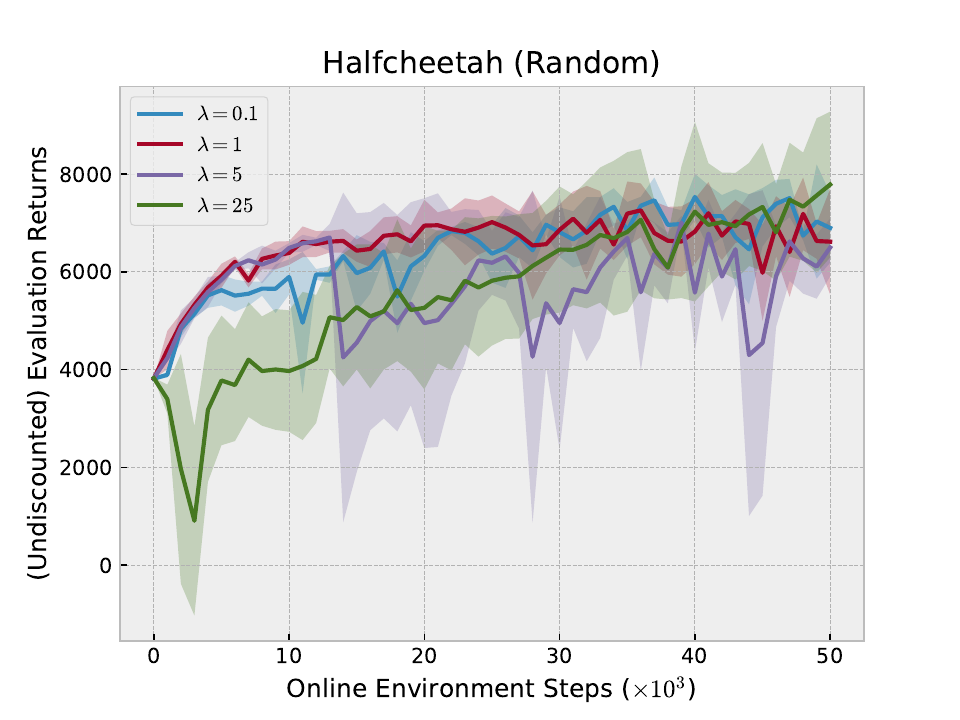}
    \end{subfigure}
    \begin{subfigure}{0.32\textwidth}
        \includegraphics[width=\textwidth]{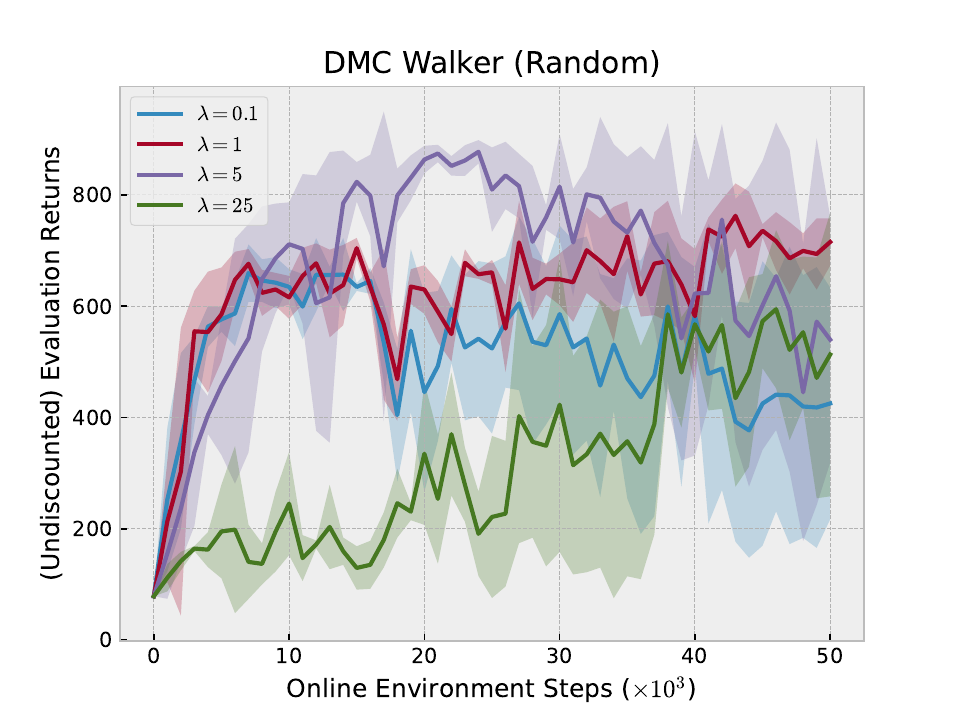}
    \end{subfigure}
    \begin{subfigure}{0.32\textwidth}
        \includegraphics[width=\textwidth]{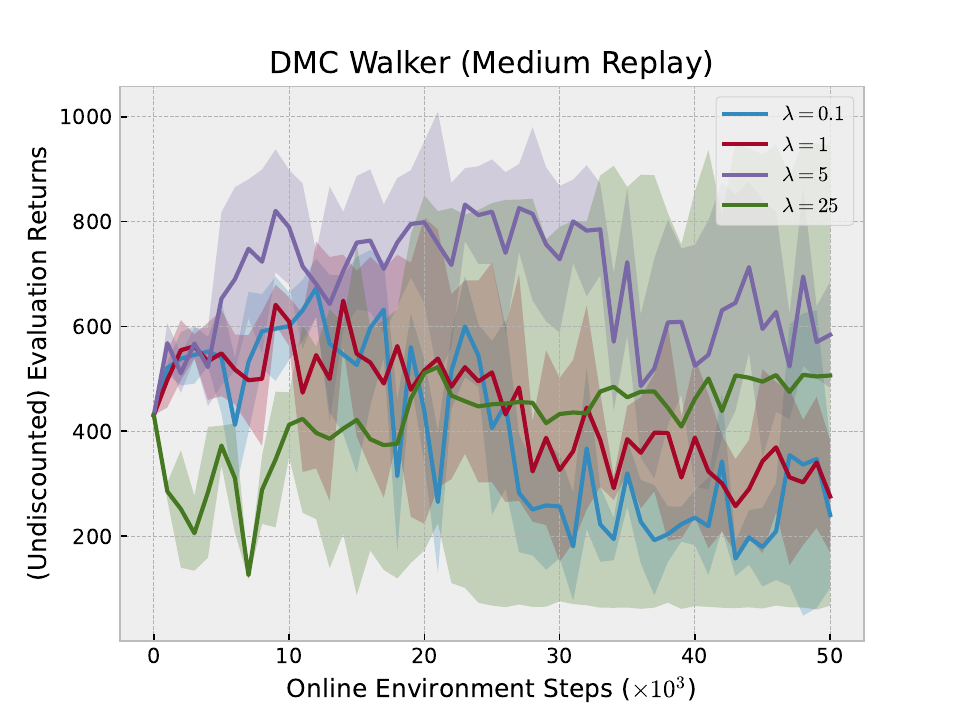}
    \end{subfigure} \hfill
    \begin{subfigure}{0.32\textwidth}
        \includegraphics[width=\textwidth]{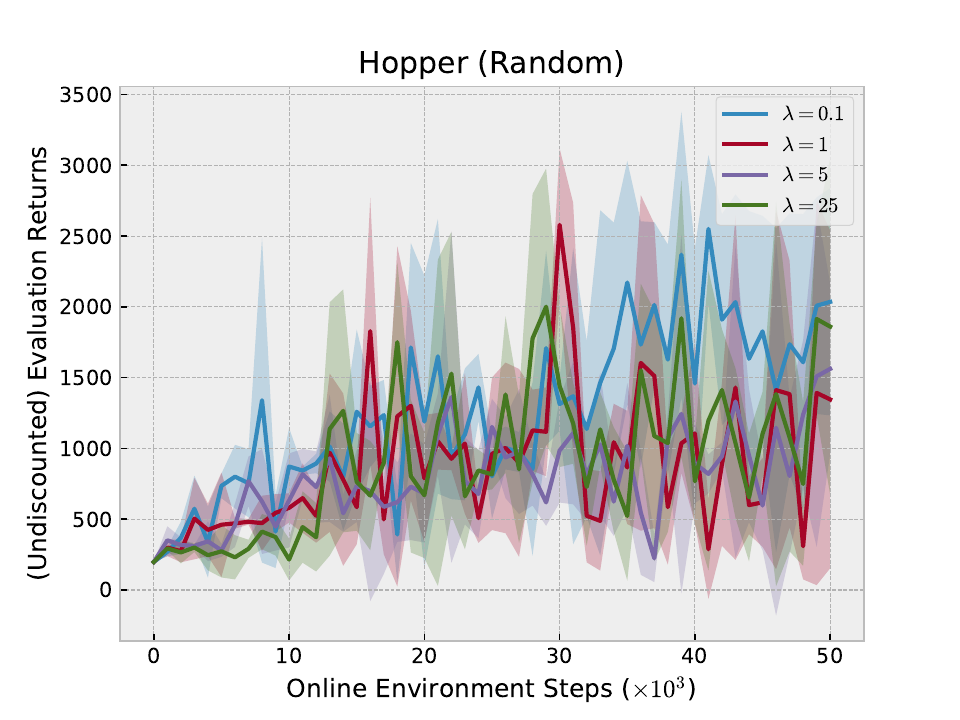}
    \end{subfigure}
    \begin{subfigure}{0.32\textwidth}
        \includegraphics[width=\textwidth]{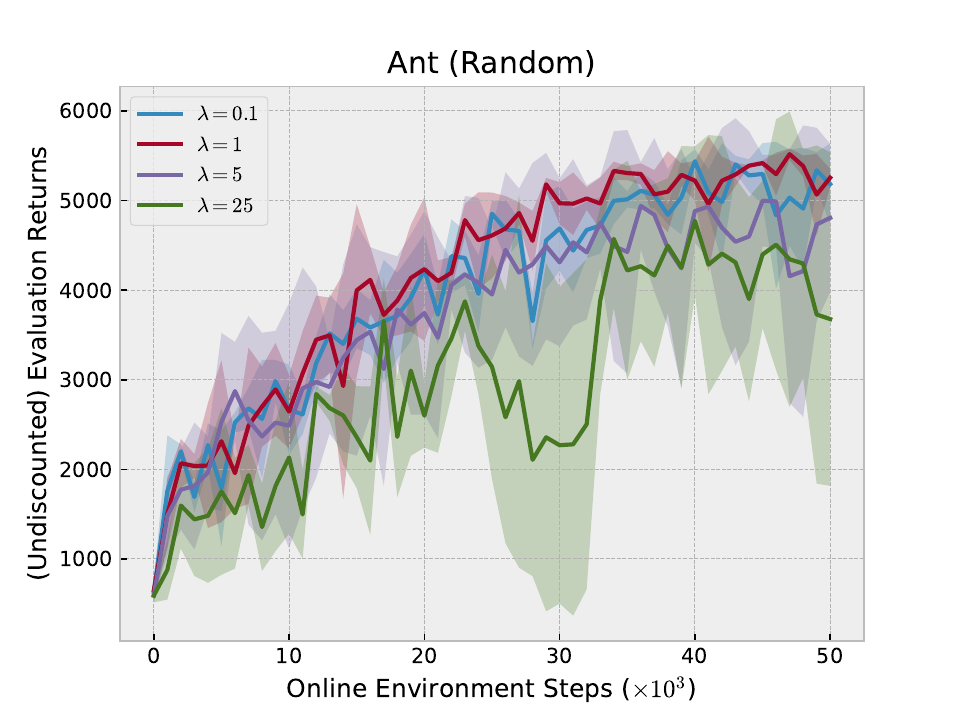}
    \end{subfigure}
    \begin{subfigure}{0.32\textwidth}
        \includegraphics[width=\textwidth]{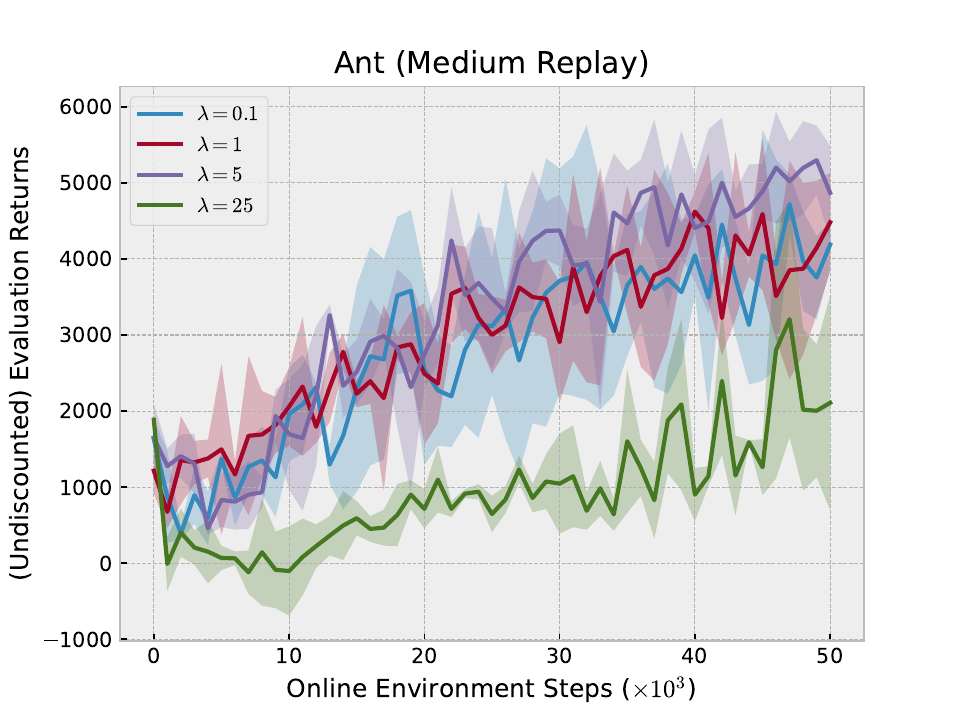}
    \end{subfigure}
    \begin{subfigure}{0.32\textwidth}
        \includegraphics[width=\textwidth]{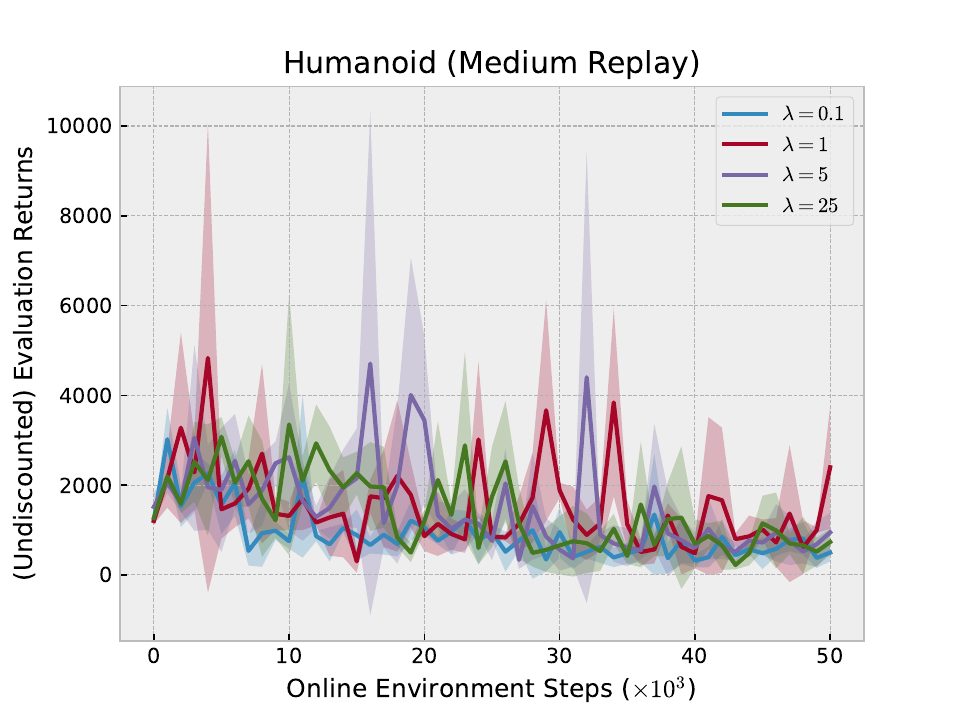}
    \end{subfigure}
    \caption{Undiscounted evaluation returns for RND/DeRL hyperparameter tuning.}
    \label{fig:rnd/derl-tuning}
\end{figure*}

\begin{figure*}[!ht]
    \centering
    \begin{subfigure}{0.32\textwidth}
        \includegraphics[width=\textwidth]{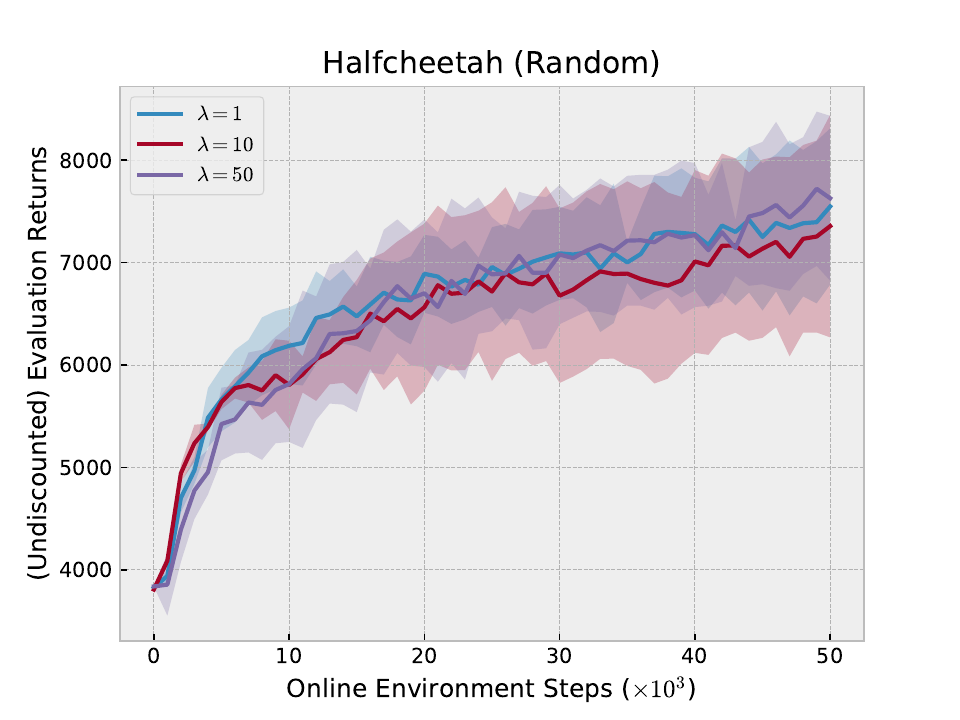}
    \end{subfigure}
    \begin{subfigure}{0.32\textwidth}
        \includegraphics[width=\textwidth]{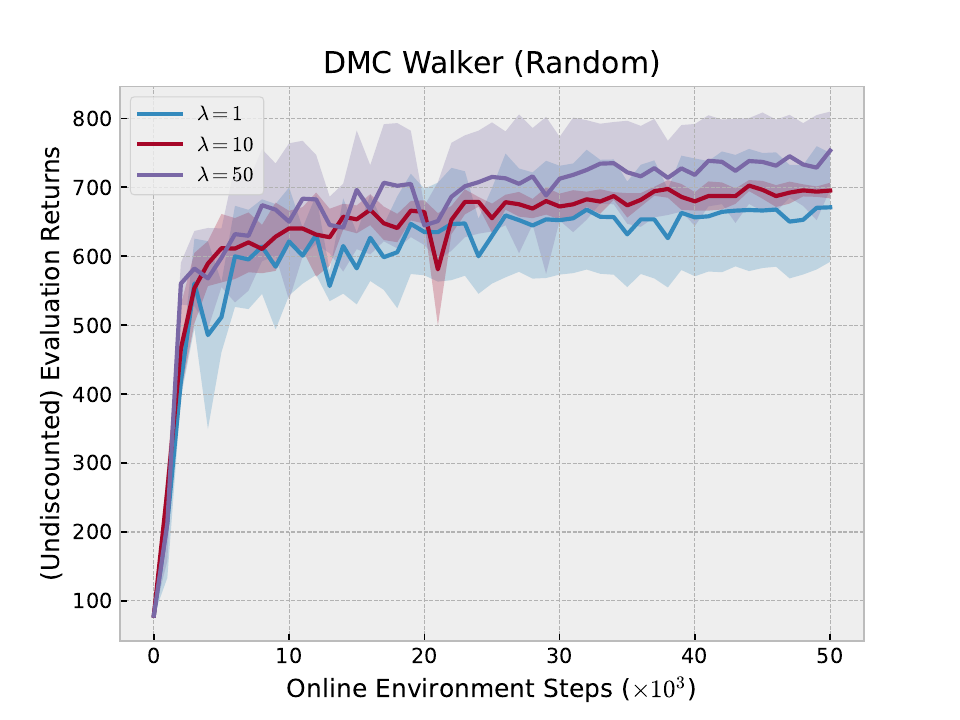}
    \end{subfigure}
    \begin{subfigure}{0.32\textwidth}
        \includegraphics[width=\textwidth]{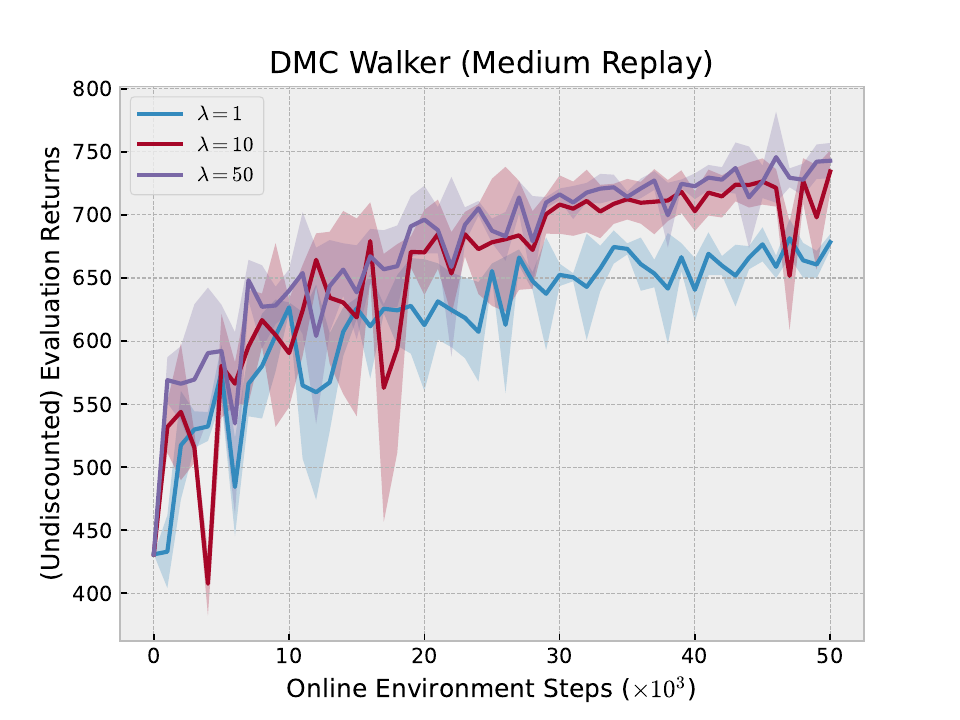}
    \end{subfigure} \hfill
    \begin{subfigure}{0.32\textwidth}
        \includegraphics[width=\textwidth]{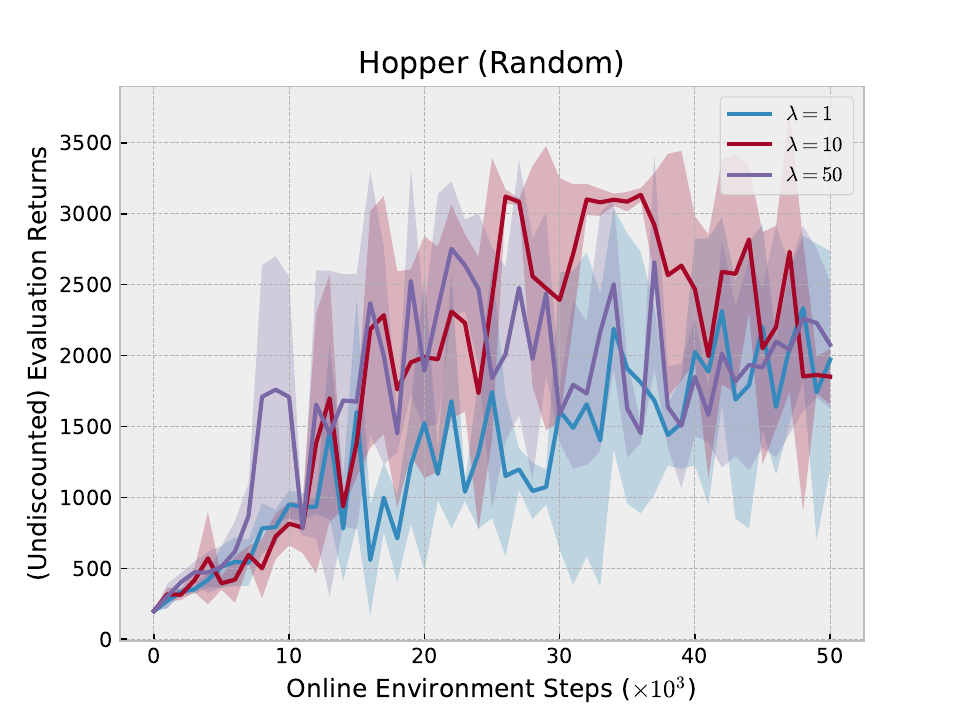}
    \end{subfigure}
    \begin{subfigure}{0.32\textwidth}
        \includegraphics[width=\textwidth]{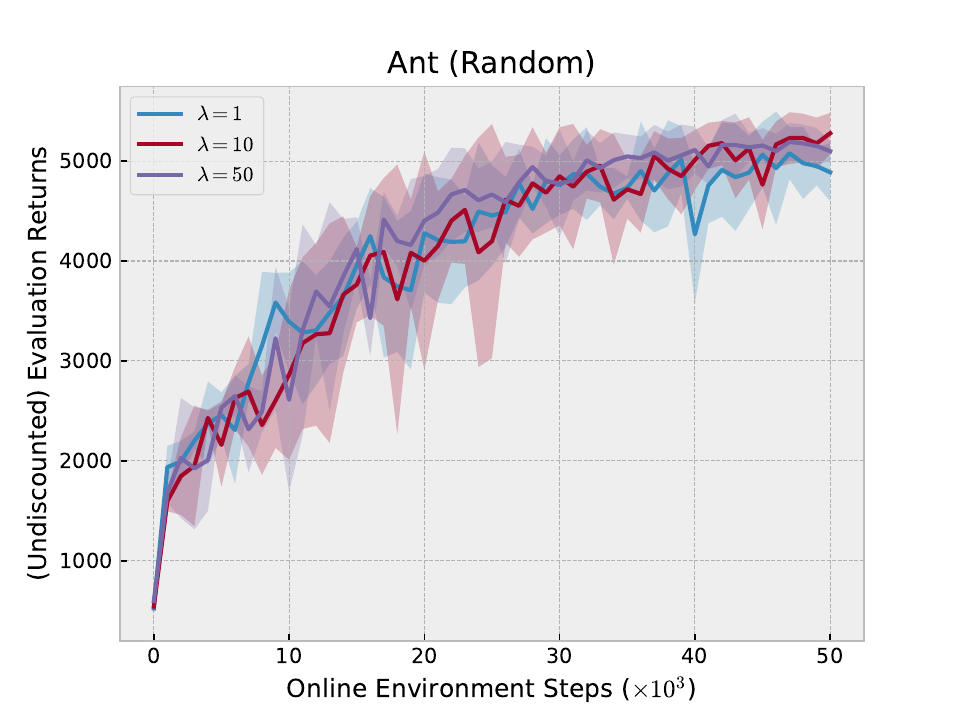}
    \end{subfigure}
    \begin{subfigure}{0.32\textwidth}
        \includegraphics[width=\textwidth]{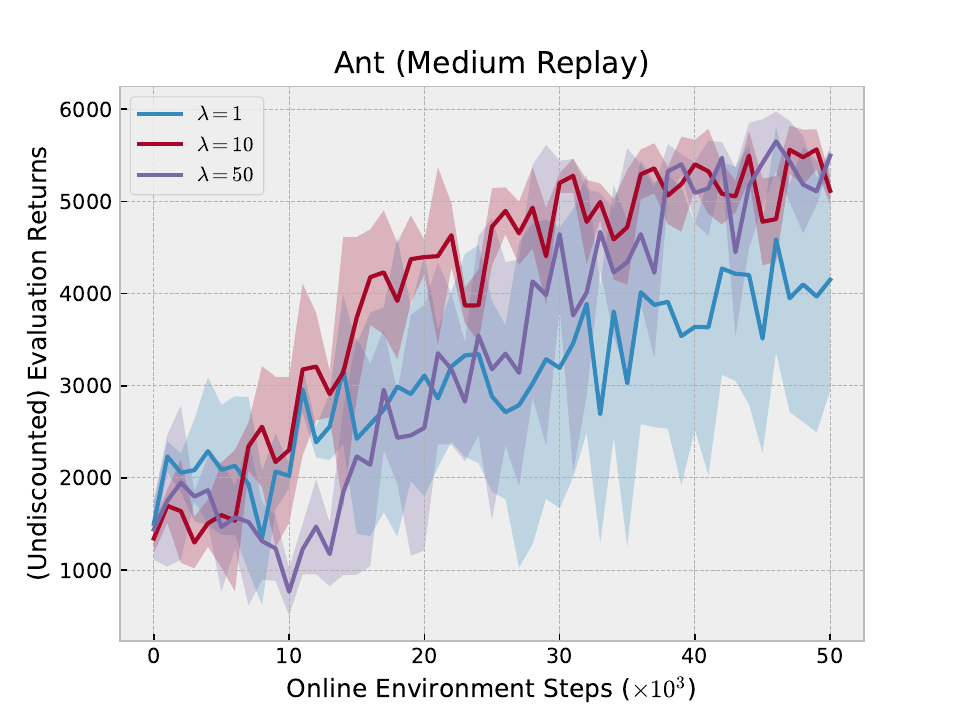}
    \end{subfigure}
    \begin{subfigure}{0.32\textwidth}
        \includegraphics[width=\textwidth]{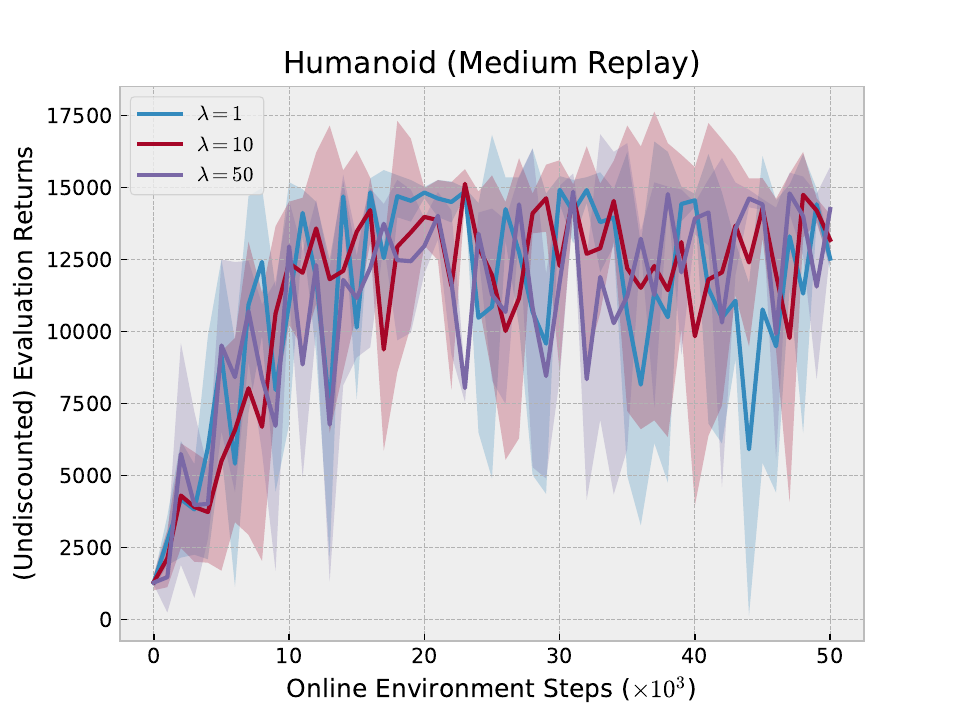}
    \end{subfigure}
    \caption{Undiscounted evaluation returns for UCB(Q) hyperparameter tuning.}
    \label{fig:ucb(q)-tuning}
\end{figure*}

\begin{figure*}[!ht]
    \centering
    \begin{subfigure}{0.32\textwidth}
        \includegraphics[width=\textwidth]{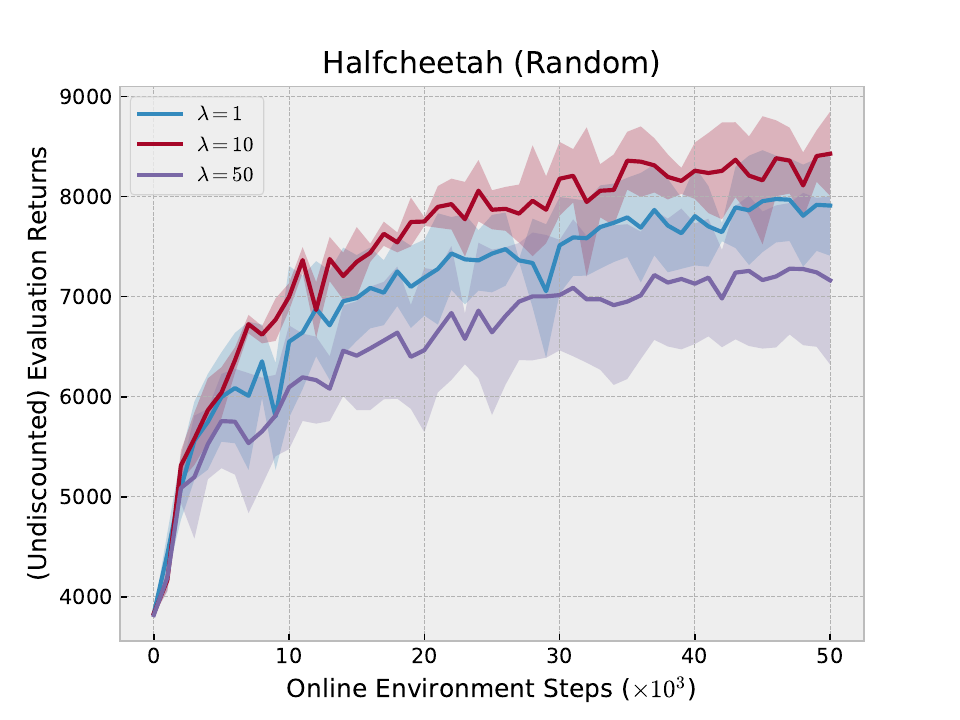}
    \end{subfigure}
    \begin{subfigure}{0.32\textwidth}
        \includegraphics[width=\textwidth]{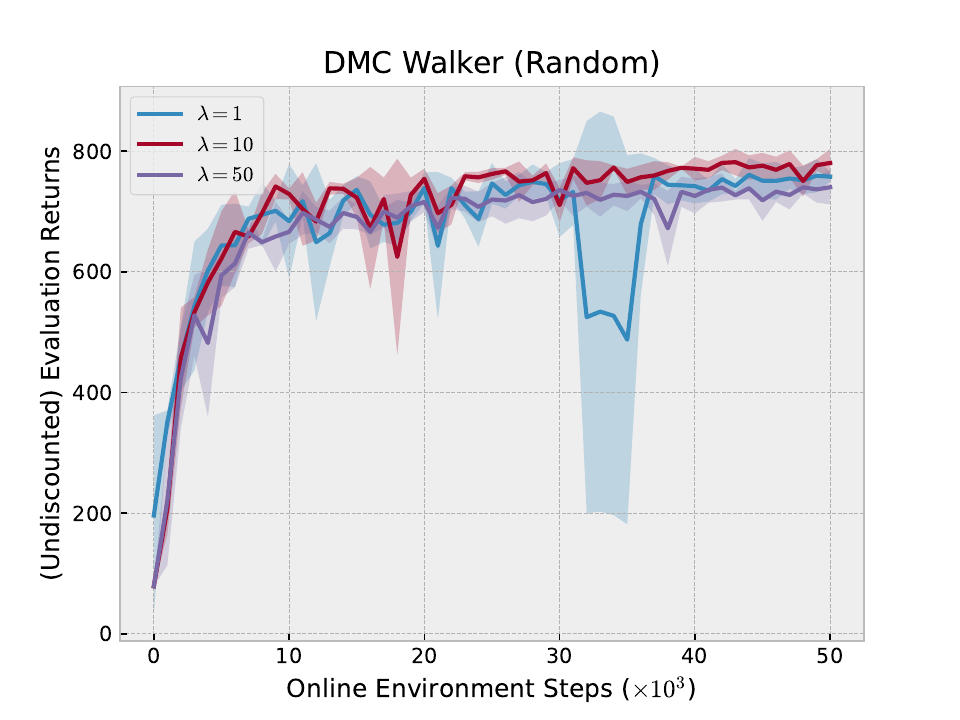}
    \end{subfigure}
    \begin{subfigure}{0.32\textwidth}
        \includegraphics[width=\textwidth]{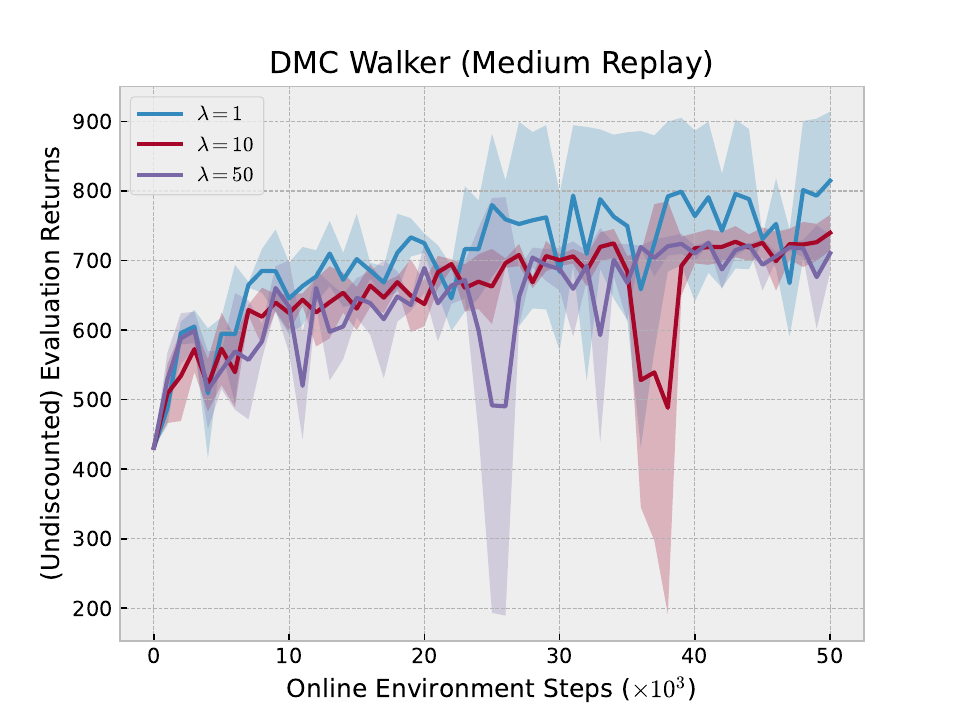}
    \end{subfigure} \hfill
    \begin{subfigure}{0.32\textwidth}
        \includegraphics[width=\textwidth]{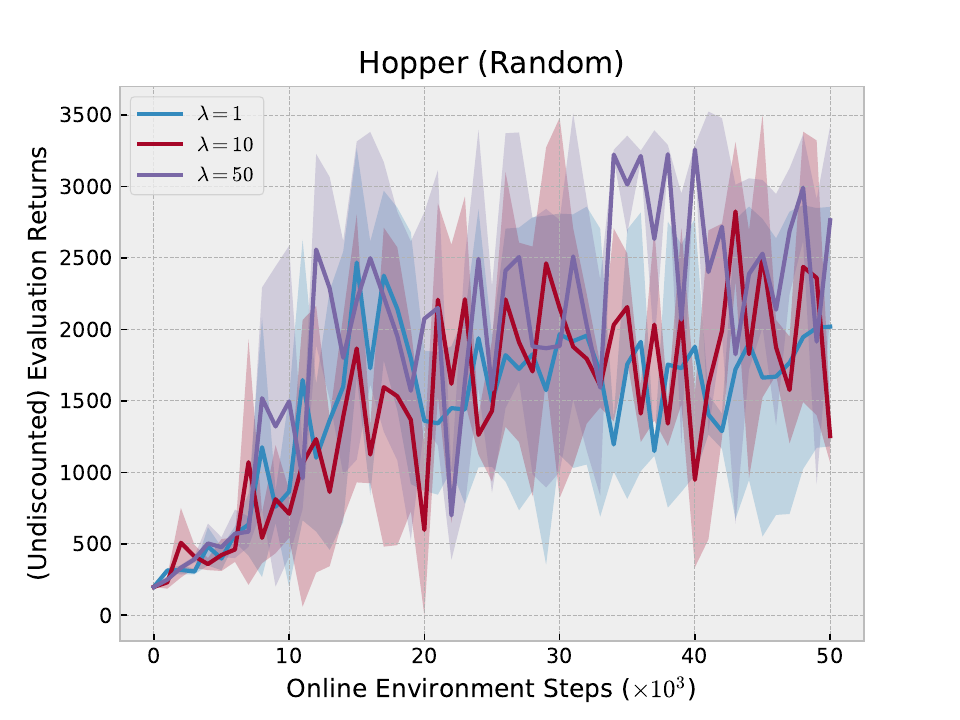}
    \end{subfigure}
    \begin{subfigure}{0.32\textwidth}
        \includegraphics[width=\textwidth]{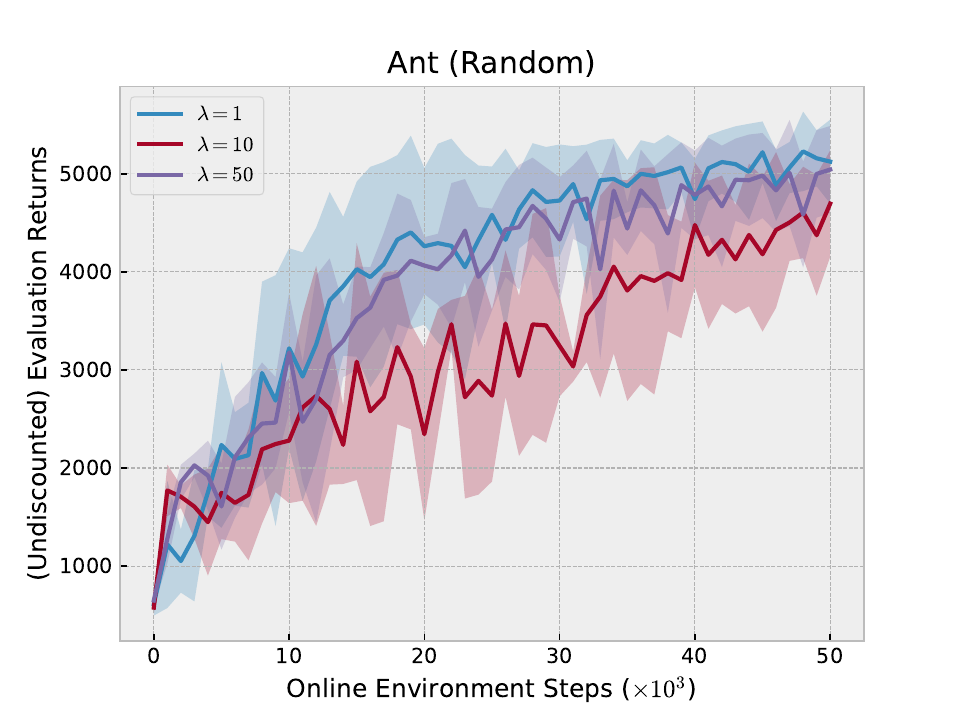}
    \end{subfigure}
    \begin{subfigure}{0.32\textwidth}
        \includegraphics[width=\textwidth]{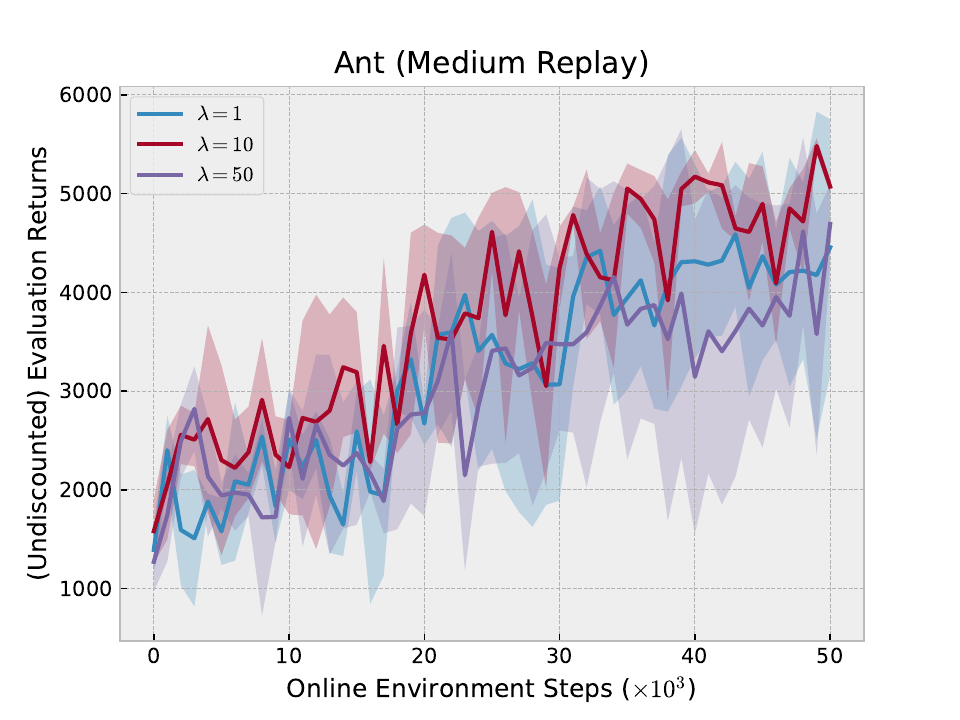}
    \end{subfigure}
    \begin{subfigure}{0.32\textwidth}
        \includegraphics[width=\textwidth]{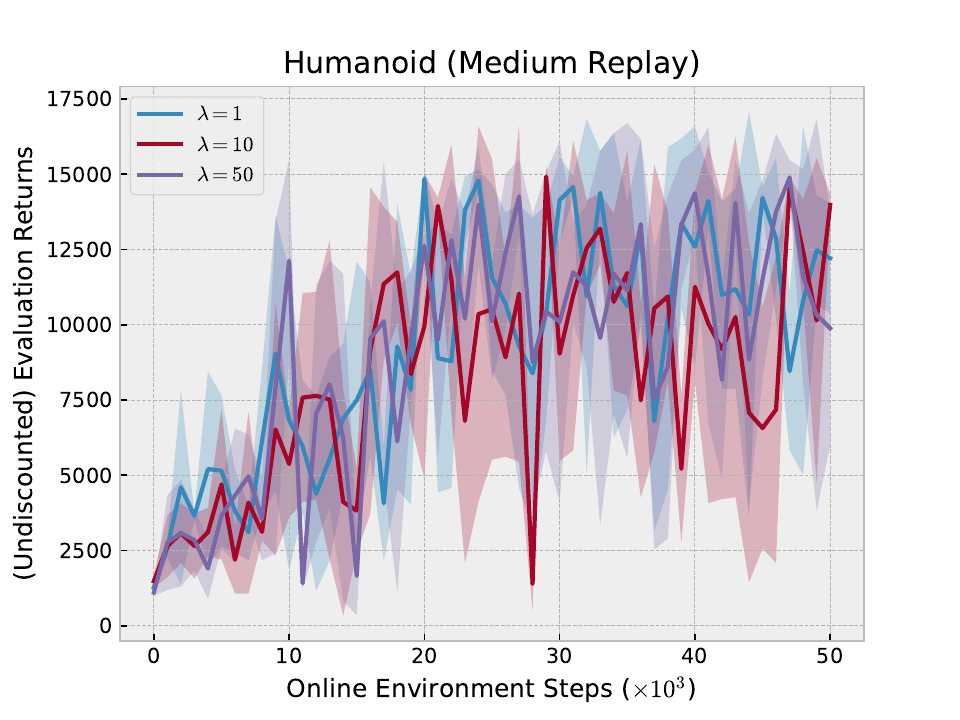}
    \end{subfigure}
    \caption{Undiscounted evaluation returns for UCB(T) hyperparameter tuning.}
    \label{fig:ucb(t)-tuning}
\end{figure*}

\begin{figure*}[!ht]
    \centering
    \begin{subfigure}{0.32\textwidth}
        \includegraphics[width=\textwidth]{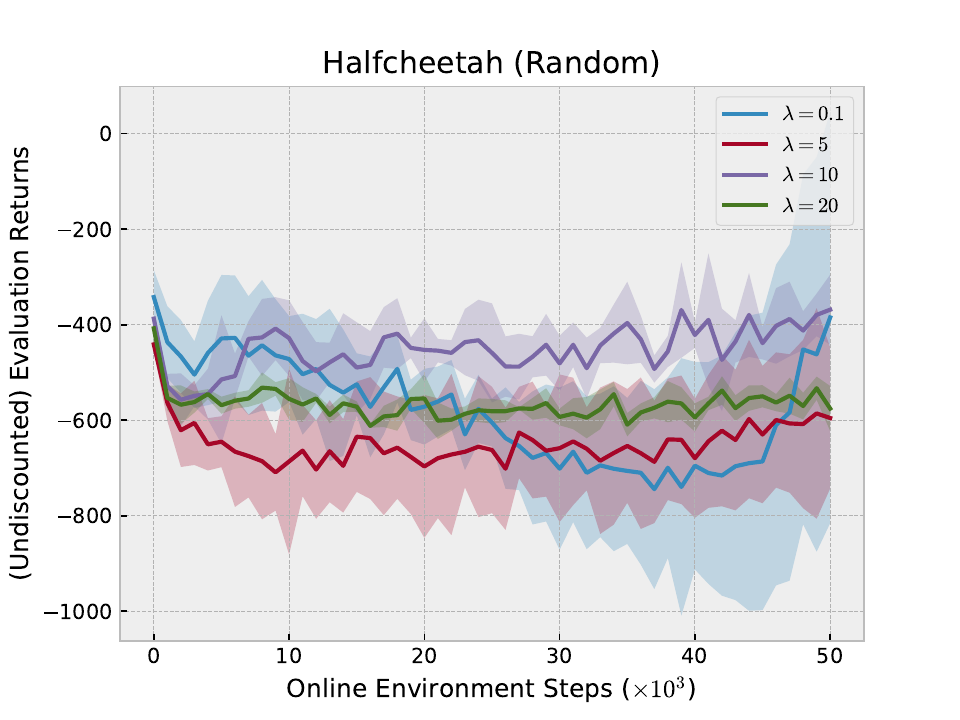}
    \end{subfigure}
    \begin{subfigure}{0.32\textwidth}
        \includegraphics[width=\textwidth]{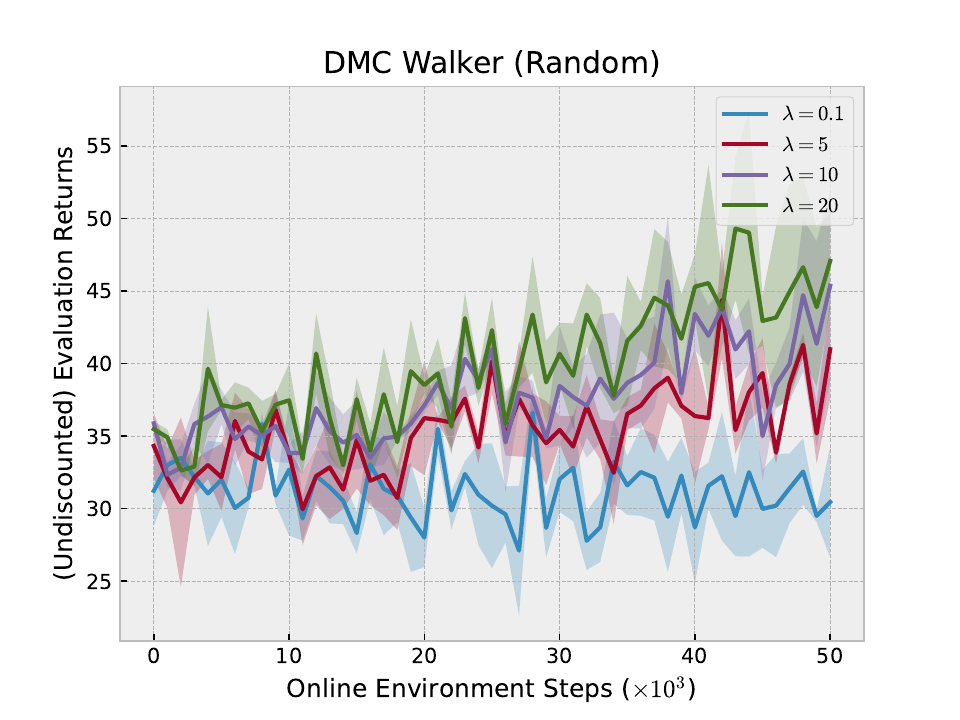}
    \end{subfigure}
    \begin{subfigure}{0.32\textwidth}
        \includegraphics[width=\textwidth]{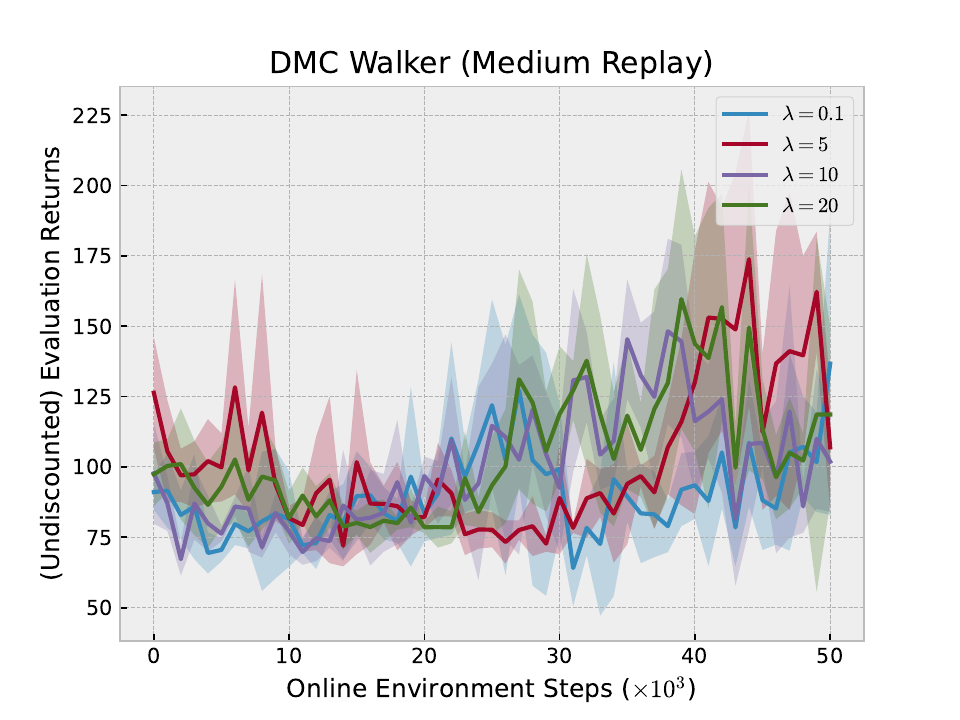}
    \end{subfigure} \hfill
    \begin{subfigure}{0.32\textwidth}
        \includegraphics[width=\textwidth]{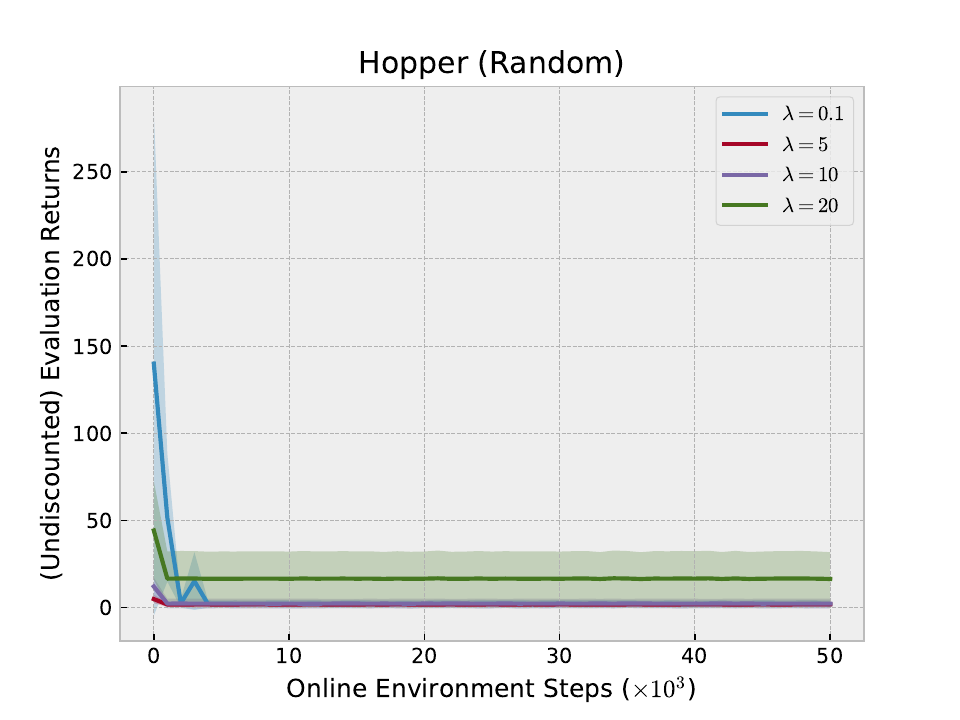}
    \end{subfigure}
    \begin{subfigure}{0.32\textwidth}
        \includegraphics[width=\textwidth]{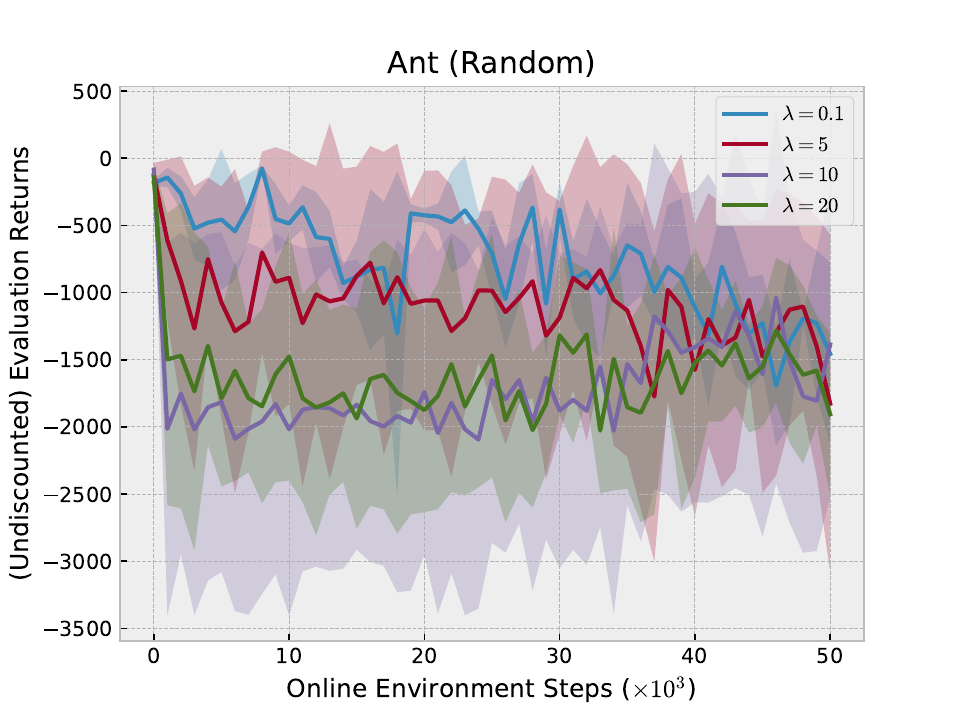}
    \end{subfigure}
    \begin{subfigure}{0.32\textwidth}
        \includegraphics[width=\textwidth]{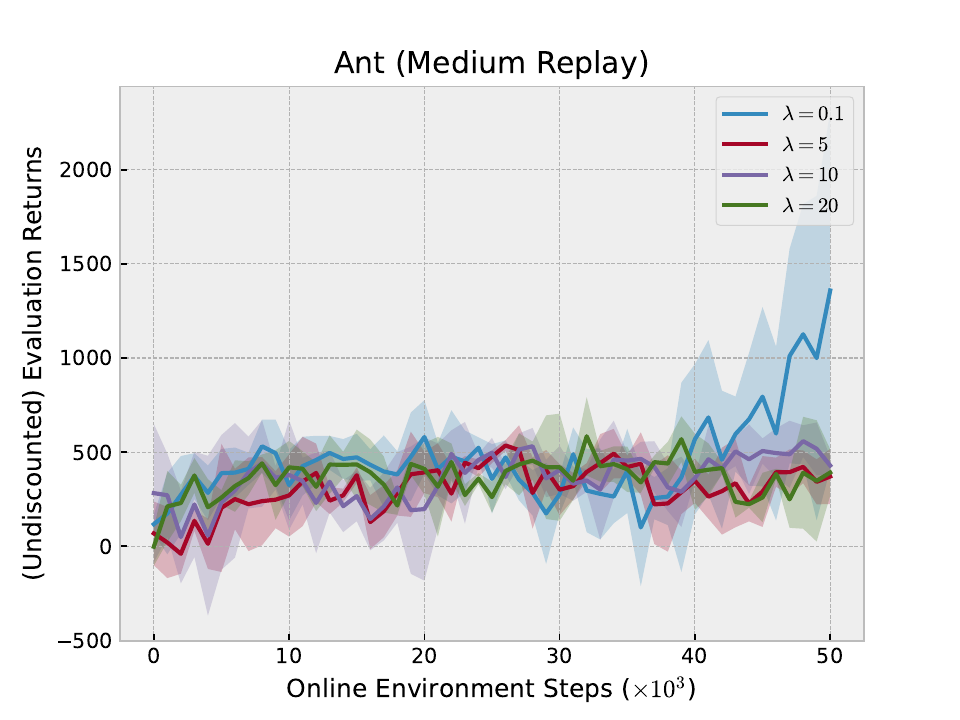}
    \end{subfigure}
    \begin{subfigure}{0.32\textwidth}
        \includegraphics[width=\textwidth]{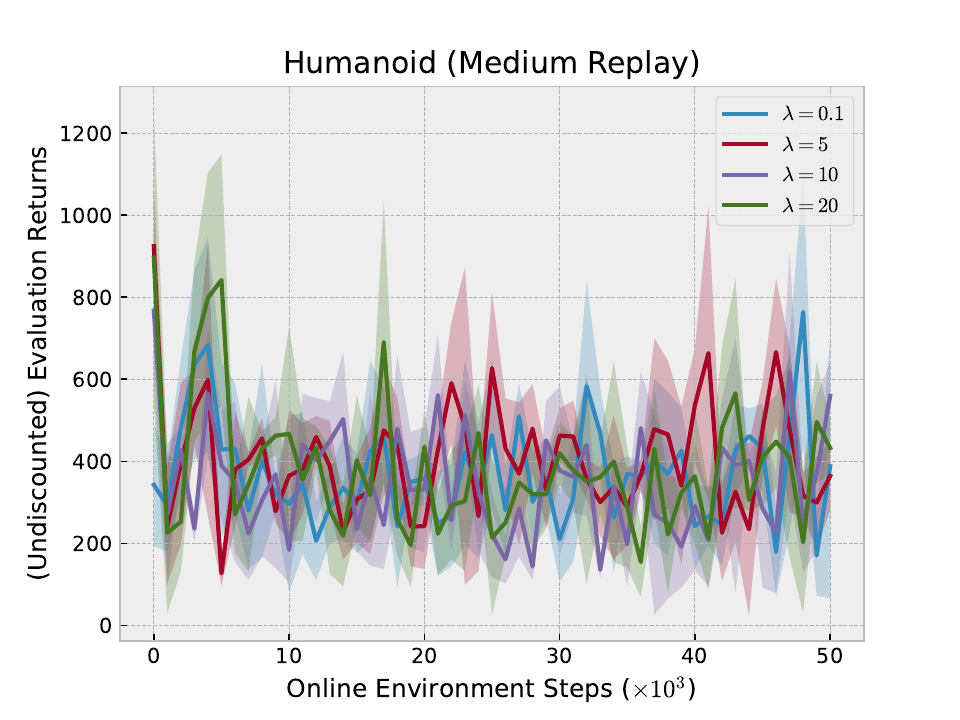}
    \end{subfigure}
    \caption{Undiscounted evaluation returns for Cal-QL (Min Q-Weight) hyperparameter tuning.}
    \label{fig:calql-tuning-minq}
\end{figure*}

\begin{figure*}[!ht]
    \centering
    \begin{subfigure}{0.32\textwidth}
        \includegraphics[width=\textwidth]{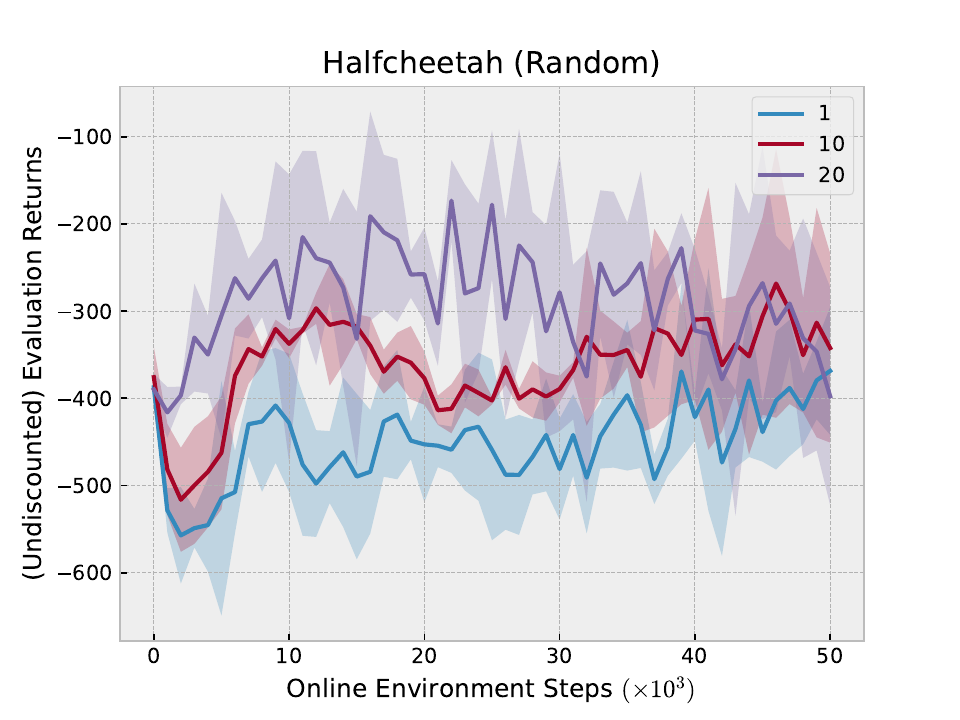}
    \end{subfigure}
    \begin{subfigure}{0.32\textwidth}
        \includegraphics[width=\textwidth]{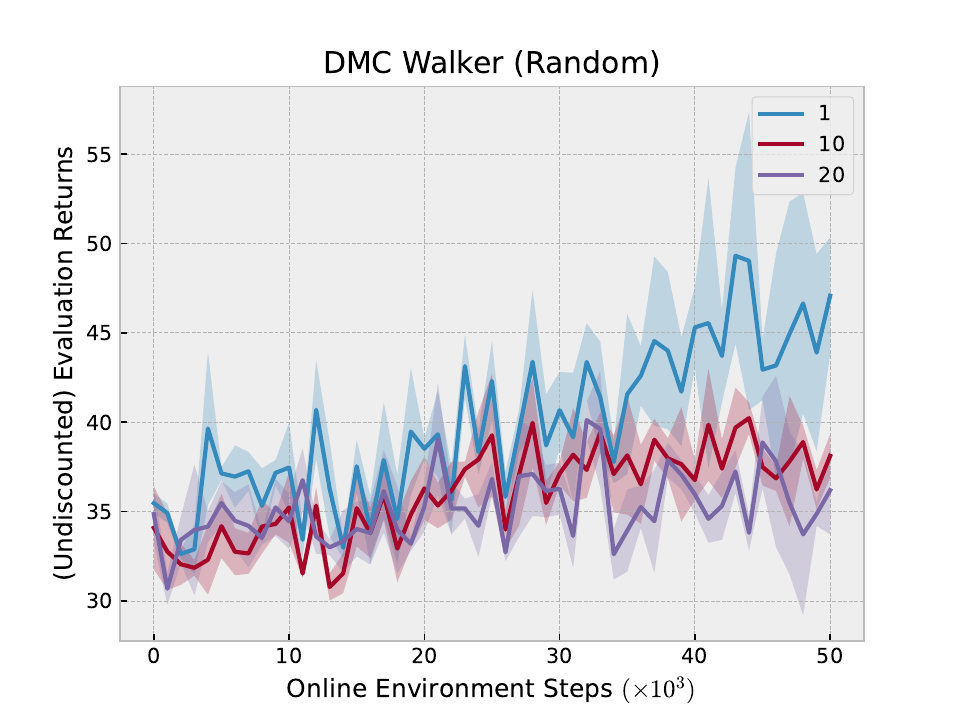}
    \end{subfigure}
    \begin{subfigure}{0.32\textwidth}
        \includegraphics[width=\textwidth]{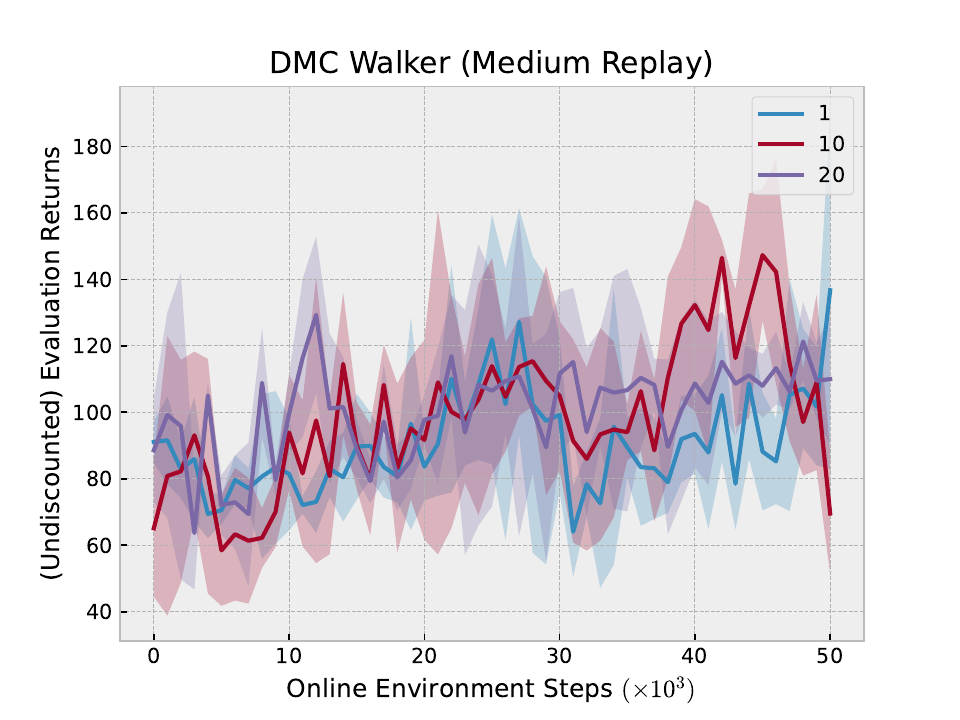}
    \end{subfigure} \hfill
    \begin{subfigure}{0.32\textwidth}
        \includegraphics[width=\textwidth]{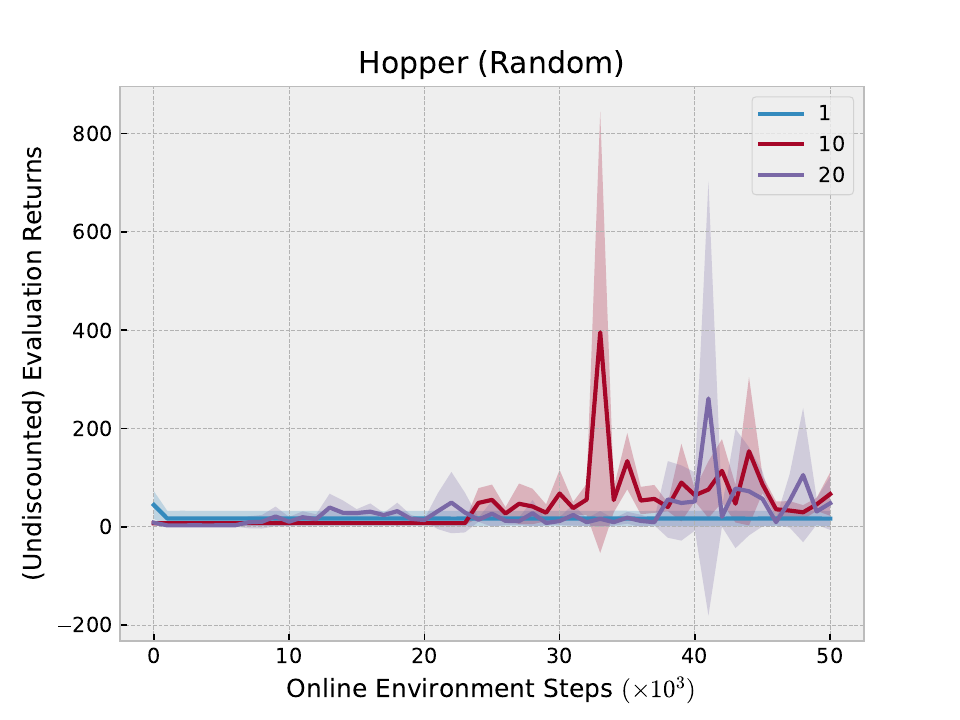}
    \end{subfigure}
    \begin{subfigure}{0.32\textwidth}
        \includegraphics[width=\textwidth]{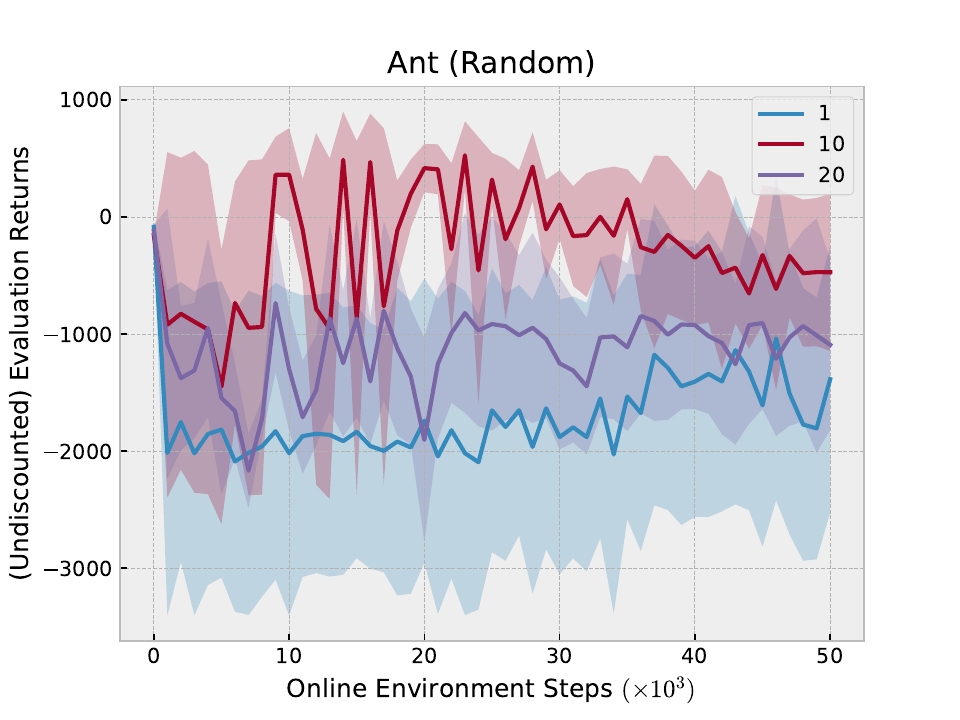}
    \end{subfigure}
    \begin{subfigure}{0.32\textwidth}
        \includegraphics[width=\textwidth]{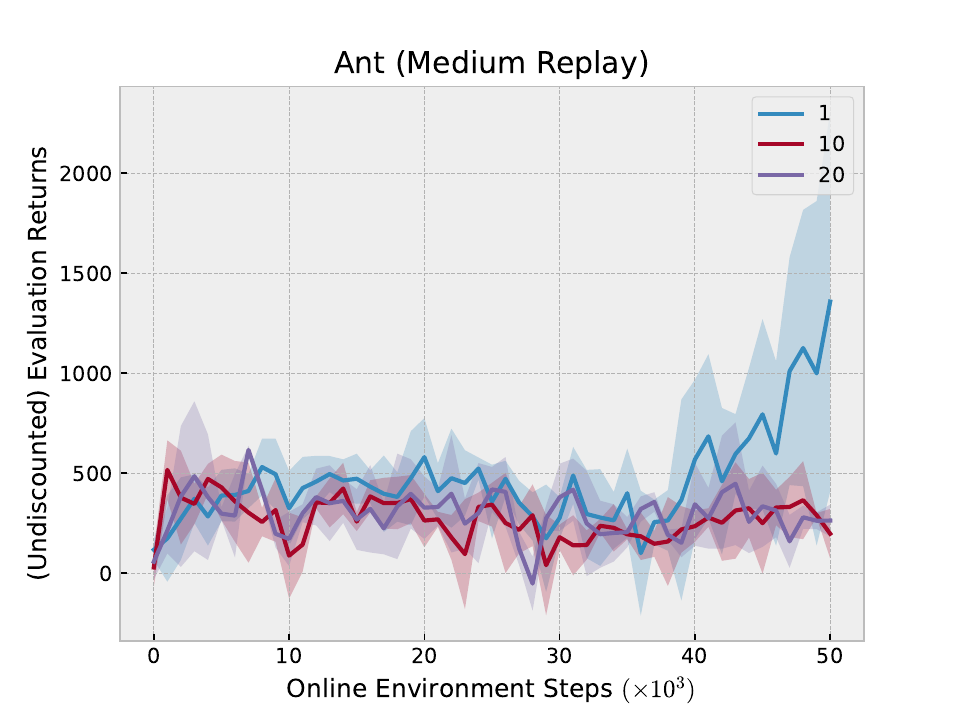}
    \end{subfigure}
    \begin{subfigure}{0.32\textwidth}
        \includegraphics[width=\textwidth]{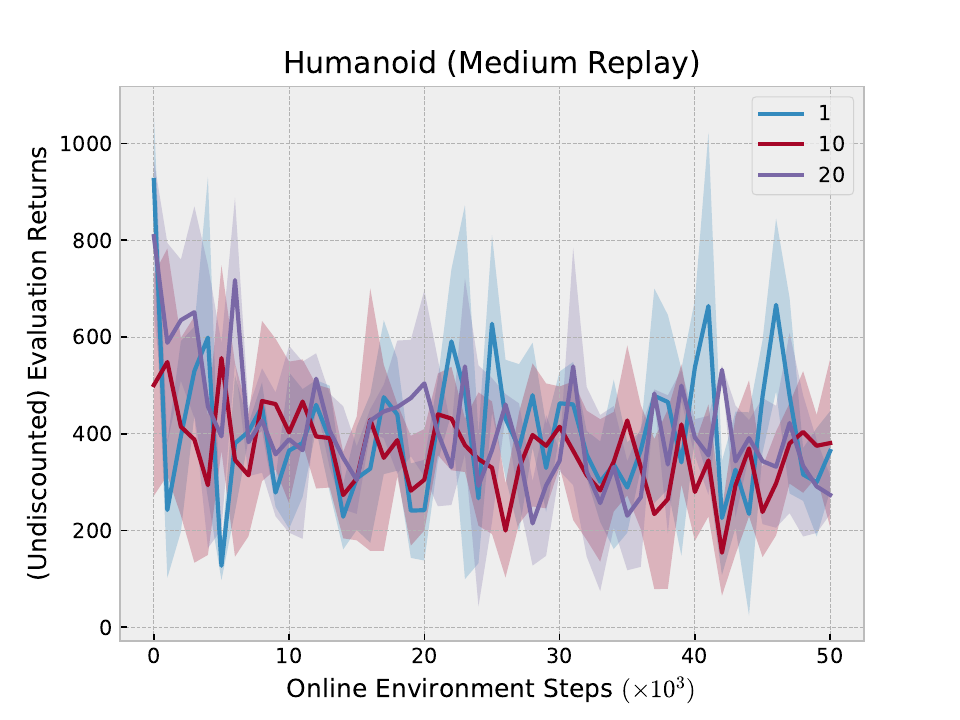}
    \end{subfigure}
    \caption{Undiscounted evaluation returns for Cal-QL (UTD) hyperparameter tuning.}
    \label{fig:calql-tuning-utd}
\end{figure*}

\begin{figure*}[!ht]
    \centering
    \begin{subfigure}{0.32\textwidth}
        \includegraphics[width=\textwidth]{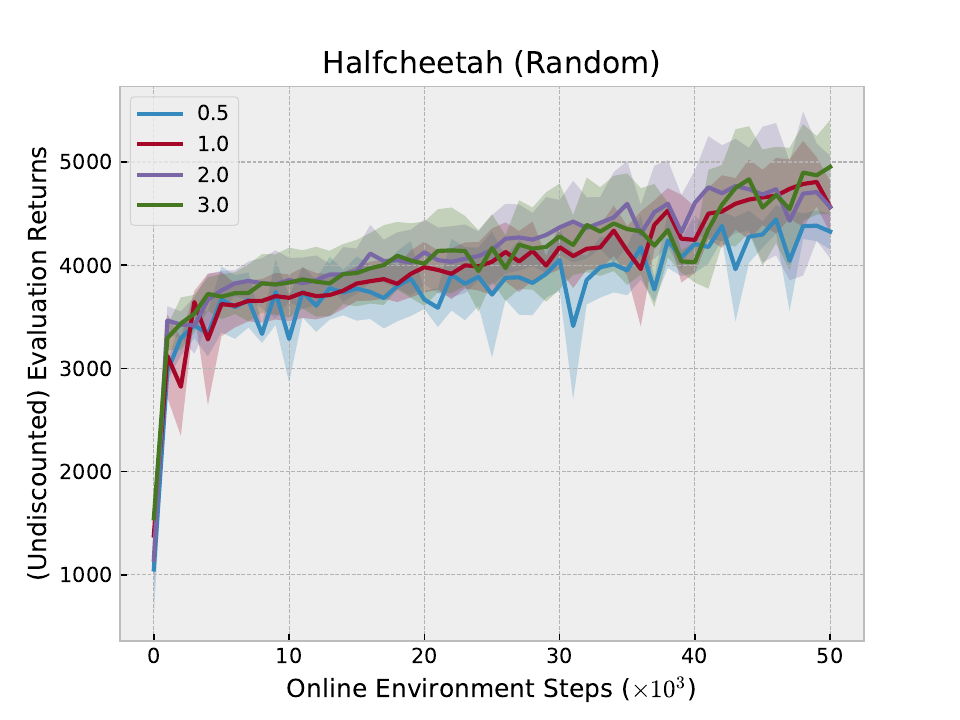}
    \end{subfigure}
    \begin{subfigure}{0.32\textwidth}
        \includegraphics[width=\textwidth]{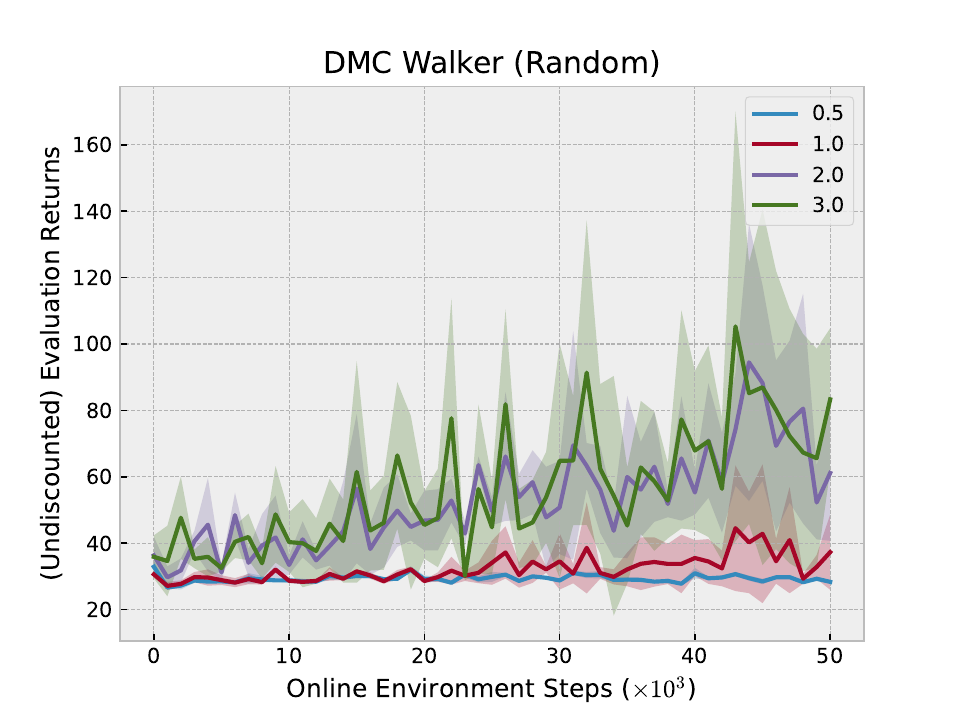}
    \end{subfigure}
    \begin{subfigure}{0.32\textwidth}
        \includegraphics[width=\textwidth]{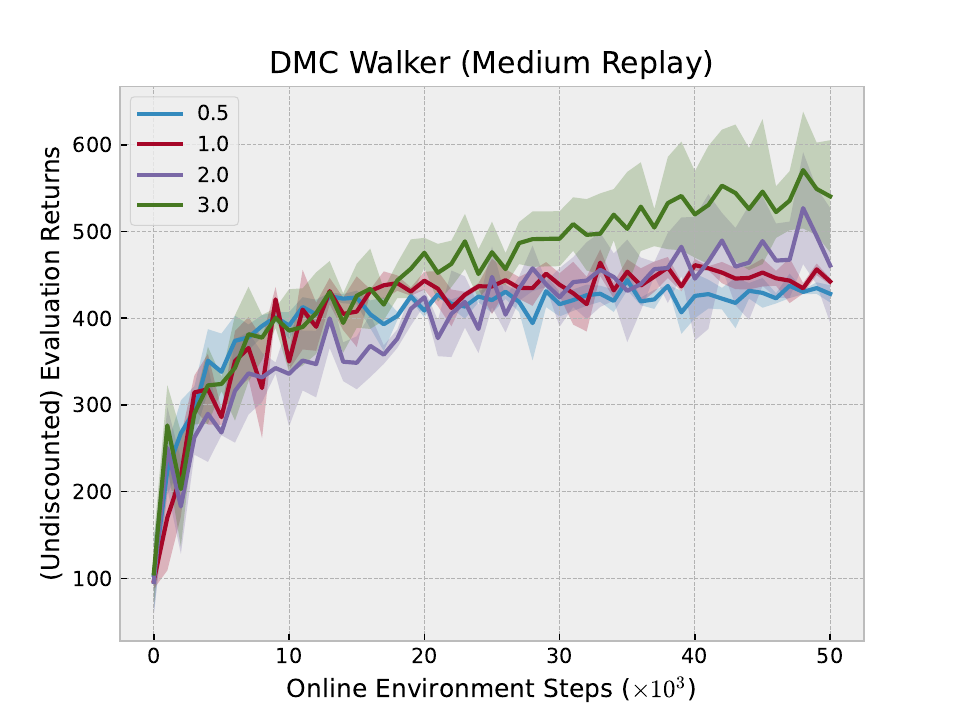}
    \end{subfigure} \hfill
    \begin{subfigure}{0.32\textwidth}
        \includegraphics[width=\textwidth]{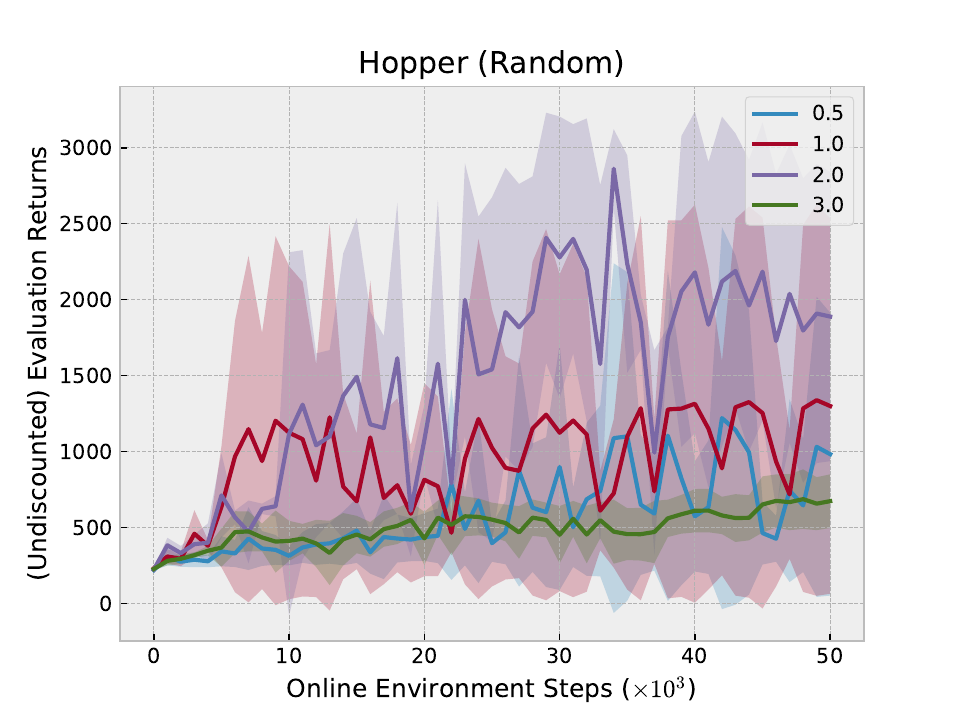}
    \end{subfigure}
    \begin{subfigure}{0.32\textwidth}
        \includegraphics[width=\textwidth]{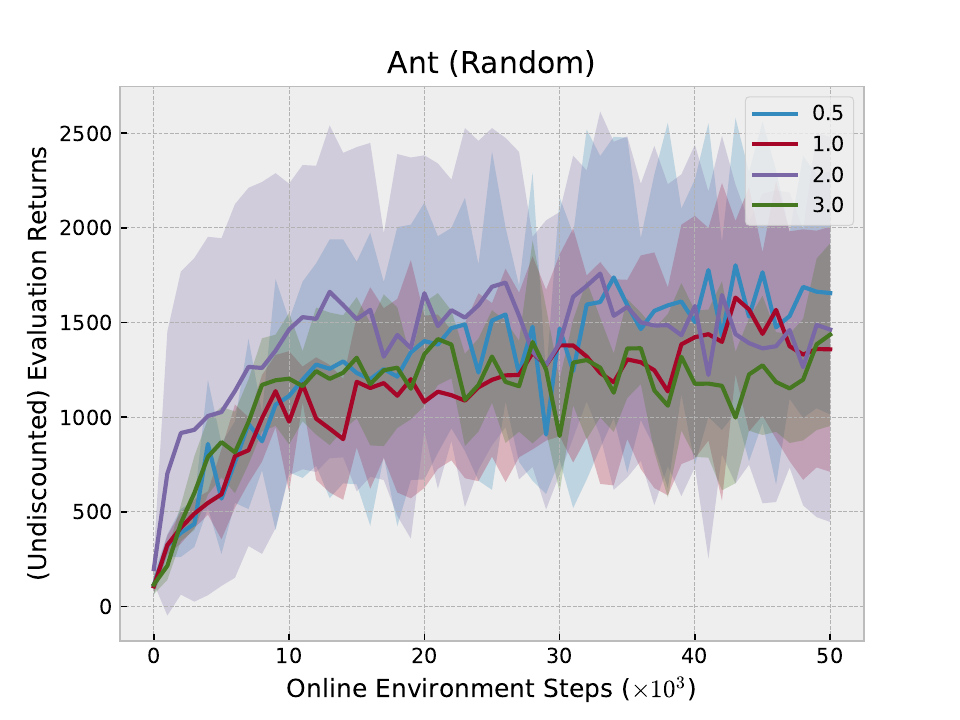}
    \end{subfigure}
    \begin{subfigure}{0.32\textwidth}
        \includegraphics[width=\textwidth]{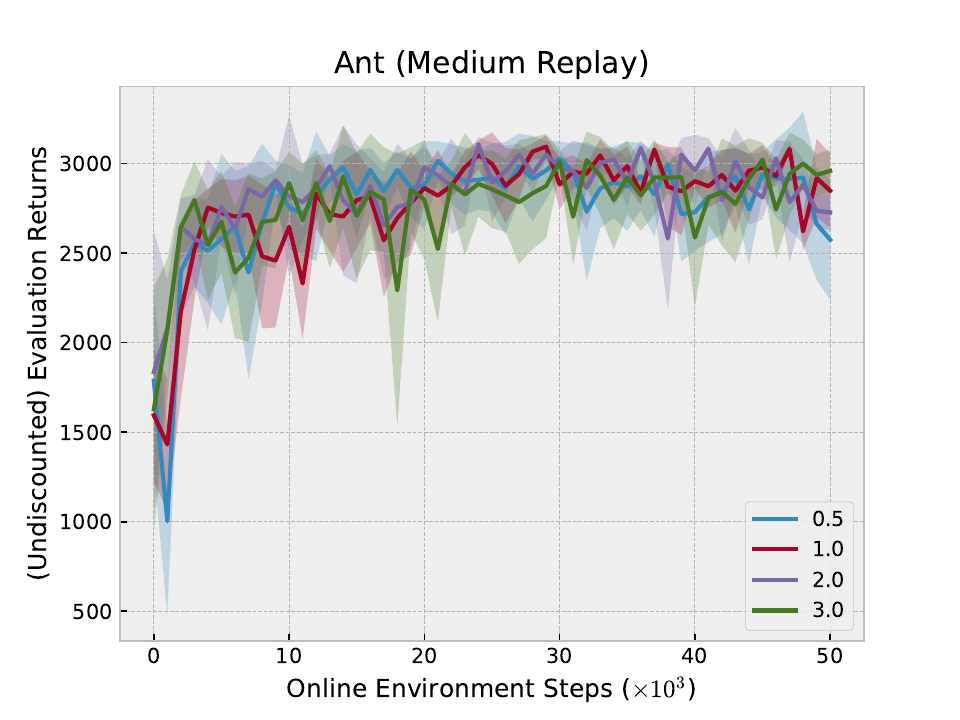}
    \end{subfigure}
    \begin{subfigure}{0.32\textwidth}
        \includegraphics[width=\textwidth]{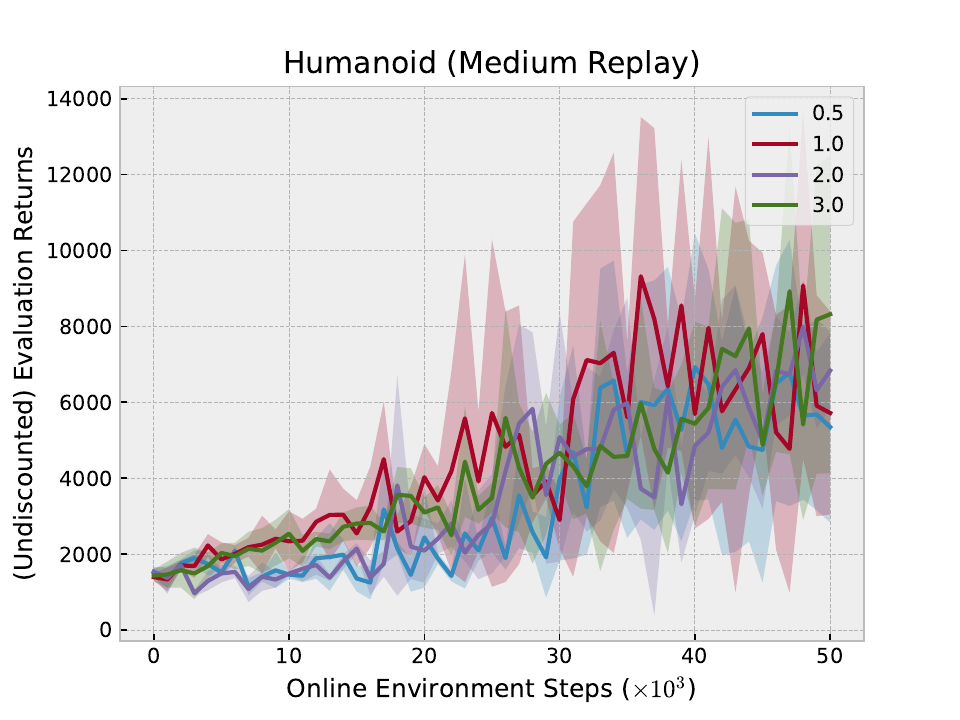}
    \end{subfigure}
    \caption{Undiscounted evaluation returns for PEX hyperparameter tuning.}
    \label{fig:pex-tuning}
\end{figure*}

We use $\lambda$ as a generic weighting hyperparameter. For RND/DeRL (Figure~\ref{fig:rnd/derl-tuning}), it weights intrinsic rewards at timestep $t$: $r_t = r^e_t + \lambda r^i_t$, and we scan $\lambda \in \{0.1, 5, 10, 25\}$. For UCB(Q) (Figure~\ref{fig:ucb(q)-tuning}), it weights the impact of uncertainty on action selection: $Q_{mean}(\cdot) + \lambda Q_{std}(\cdot)$, and we scan $\lambda \in \{1, 10, 50\}$. For UCB(T) (Figure~\ref{fig:ucb(t)-tuning}), it weights the impact of uncertainty on action selection: $Q(\cdot) + \lambda T_{std}(\cdot)$, and we scan $\lambda \in \{1, 10, 50\}$. For Cal-QL (Figure~\ref{fig:calql-tuning-minq}), it weights the Min Q-weight, which we found to be particularly impactful based on the hyperparameter sweeps found here: \url{https://wandb.ai/ygx/JaxCQL--jax_cql_gym_sweep_3}. In addition, we performed a sweep over the number of RL updates per environment step (Figure~\ref{fig:calql-tuning-utd}), called ``UTD" in the Cal-QL paper. For Min Q-Weight, we scan $\lambda \in \{0.1, 1, 5, 25\}$, and for UTD we  scan $\lambda \in \{1, 10, 20\}$. We also fine-tuned PEX (Figure~\ref{fig:pex-tuning}). Figure 7 in the PEX paper shows that PEX is sensitive to the ``inverse temperature" hyperparameter. For this hyperparameter, we follow the original authors and scan $\alpha^{-1} \in \{0.5, 1, 2, 3\}$. Interestingly, the PROTO paper shows that PROTO is not sensitive to the value of hyperparameters that impact important PROTO-specific mechanisms.  Specifically, Figure 14 in the PROTO paper shows that adjusting the conservative annealing speed $\eta$ does not affect PROTO agent performance in the slightest. As such, we choose not to waste GPU compute and instead use the hyperparameters suggested by the original authors.

For each hyperparameter setting, we run three seeds. Each plot shows the average (bold line) $\pm$ one standard deviation (shaded area). For the final results we present in the paper, we select the best performing hyperparameter setting for each algorithm on a per-environment basis and run two additional seeds.

\clearpage
\section{PTGOOD Pseudocode}\label{app:ptgood-pseudocode}
\begin{algorithm}
\caption{PTGOOD Planning Procedure}\label{alg:ptgood-planning}
\begin{algorithmic}[1]
\Require Dynamics model $\mathcal{\hat{T}}$, encoder $e$, marginal $m$, depth $d$, width $w$, state $s$, policy $\pi$, noise hyperparameter $\epsilon$
    \State Initialize empty ordered list \texttt{action\_list}
    \State Initialize empty ordered list \texttt{rate\_sums}
    \State Initialize empty list \texttt{inner\_action\_list}
    \State Initialize empty list \texttt{state\_list}
    \State Initialize empty list \texttt{inner\_state\_list}
    \State Initialize tree \texttt{rate\_tree} with single node for $s$
    \For{$i$ in range($w$)}
        \State Sample action, add sampled noise $a \sim \pi(\cdot \vert s), \; u \sim N(0, \epsilon), \; a \leftarrow a + u$
        \State Append $a$ to \texttt{action\_list}
        \State Create branch associated to $a$ and linked to $s$ in \texttt{rate\_tree}
    \EndFor
    \For{$a$ in \texttt{action\_list}}
        \State Predict next-state $s^{\prime} \sim \mathcal{\hat{T}}(s,a)$
        \State Append $s^{\prime}$ to \texttt{state\_list}
        \State Create node for $s^{\prime}$ linked to branch $a$ in  \texttt{rate\_tree}
    \EndFor
    \For{$i$ in range($d$)}
        \For{$s$ in \texttt{state\_list}}
            \For{$j$ in range($w$)}
                \State Sample action, add sampled noise $a^{\prime} \sim \pi(\cdot \vert s), \; u \sim N(0, \epsilon), \; a^{\prime} \leftarrow a^{\prime} + u$
                \State Append $a^{\prime}$ to \texttt{inner\_action\_list}
                \State Measure rate $p \leftarrow \mathcal{R}(s,a)$
                \State Store rate $p$ in \texttt{rate\_tree} node $s$ 
                \State Create branch associated to $a^{\prime}$ and linked to $s$ in \texttt{rate\_tree}
            \EndFor
            
            \For{$a$ in \texttt{inner\_action\_list}}
                \State Predict next-state $s^{\prime} \sim \mathcal{\hat{T}}(s,a)$
                \State Append $s^{\prime}$ to \texttt{inner\_state\_list}
                \State Create node for $s^{\prime}$  linked to branch $a$ in \texttt{rate\_tree}
            \EndFor
        \EndFor
        \State \texttt{state\_list} $\leftarrow$ \texttt{inner\_state\_list}
        \State Clear \texttt{inner\_action\_list} and \texttt{inner\_state\_list}
    \EndFor
    \For{$a$ in \texttt{action\_list}}
        \State \texttt{rate\_sum} $\leftarrow 0$
        \State Traverse tree until terminal node all the while summing all rates $p$ within each node: \texttt{rate\_sum} $\leftarrow$ \texttt{rate\_sum} $+ p$
        \State Append \texttt{rate\_sum} to \texttt{rate\_sums}
    \EndFor
    \State Find index of maximum summed rate \texttt{max\_idx} $\leftarrow \argmax$ \texttt{rate\_sums}
    \State \texttt{max\_rate\_action} $\leftarrow$ \texttt{action\_list}[\texttt{max\_idx}]
\Ensure \texttt{max\_rate\_action}
\end{algorithmic}
\end{algorithm}

\clearpage
\section{Compute Cost Comparison}\label{app:compute-cost-compare}
\begin{figure}[!t]
    \centering
    \includegraphics[width=0.3\textwidth]{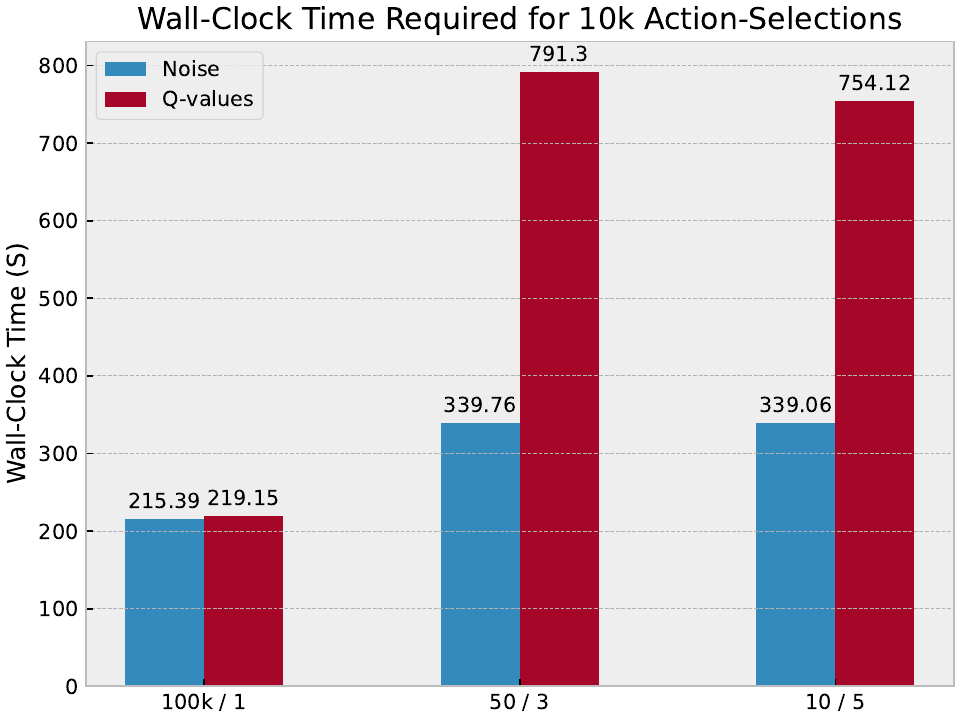}
    \caption{Wall-clock time (y-axis) comparison between noise-only planning (blue) and planning with Q-values (red) for three different width / depth combinations (x-axis).}
    \label{fig:wall-clock}
\end{figure}

We compare the wall-clock time of a PTGOOD planning process that uses only additive random noise (Noise) and one that uses additive random noise \textbf{and} computes Q-values (Q-values). We evaluate these two variations over three depths and widths (reported as width / depth): 100000 / 1, 50 / 3, 10 / 5. Specifically, we run each planning procedure for 10k environment steps five times and reports the average wall-clock time in seconds in Figure~\ref{fig:wall-clock}. We highlight that as soon as planning becomes non-myopic, using only noise provides significant gains in compute time.

\section{Environments and Datasets}\label{app:envs}

From the D4RL~\citep{d4rl} dataset we use Halfcheetah (Random) and Hopper (Medium Replay). We collect our own datasets in the Walk task in Walker from DMC, the Walk task in Humanoid from the original MBPO~\citep{mbpo} study, and the Walk task in the Ant environment from the original MBPO study. All (Random) datasets were collected with a policy that selects actions uniformly at random. All (Medium Replay) datasets were collected by saving the replay buffer of an MBPO+SAC agent trained purely online until ``medium" performance. The medium performance is defined as generating evaluation returns of 400, 3000, and 6000 for DMC Walker, Ant, and Humanoid, respectively. Table~\ref{tab:custom-dataset-info} lists the number of transitions included in each custom dataset used in this study.

\begin{table}[]
    \centering
    \begin{tabular}{c|c}
         Environment-dataset & Number of Transitions \\ \hline
         DMC Walker (R) & 50,000 \\
         Ant (R) & 1,000,000 \\
         DMC Walker (MR) & 22,000 \\
         Ant (MR) & 102,000 \\
         Humanoid (MR) & 206,000 \\
    \end{tabular}
    \caption{The number of transitions included in the custom datasets used for this study.}
    \label{tab:custom-dataset-info}
\end{table}

\section{Architecture, Hyperparameters,  and More Details}\label{app:arch-and-hypers}

The MBPO+SAC agents use an ensemble of seven MLP dynamics models that parameterize Gaussians. In Humanoid environments, the MLPs are four layers with 800 hidden units each. In the Ant environments, the MLPs are four layers with 400 hidden units each. In all other environments, the MLPs are four layers with 300 hidden units each. All MLPs use \textit{elu} activations. We train and perform inference in the same way as the original MBPO paper (see Table 1 in~\citep{mbpo}). For any differences in hyperparameters, see Table~\ref{tab:ptgood-hypers}. For environments with early-termination conditions, we zero out the rate value in states within the planning process that would terminate the episode to avoid incentivizing the agent to explore these paths.

Also, the MBPO+SAC agents use MLP actor and critic networks. In Humanoid and Ant environments, the MLPs are three layers with 512 hidden units each. In all other environments, the MLPs are three layers with 256 hidden units each. All MLPs use \textit{elu} activations and the critic networks use layer norm operations. At each training step, data are sampled from the offline dataset, dataset of online interactions, and the model-generated synthetic transitions in equal parts. 

The CEB encoder and decoder networks are both three-layer MLPs with 256, 128, and 64 hidden units and \textit{elu} activations. The learned marginal is a Gaussian mixture model with 32 components.

All networks were trained with the Adam optimizer. The dynamics models used a learning rate of 1e-3 and a weight decay of 1e-5. The critic networks and learnable alpha were trained with a learning rate of 3e-4, while the actor networks used a learning rate of 1e-4. The target critic networks used a tau of 5e-3 with an update frequency of every other step.

For Cal-QL, we used the code and default architecture settings provided by the authors here: \url{https://github.com/nakamotoo/Cal-QL}.

UCB(Q) and UCB(T) both used seven ensemble members for their respective uncertainty computations.

RND/DeRL fine-tunes its RND predictor at the same frequency as its base agent updates its ensemble of world models (shown in Table~\ref{tab:ptgood-hypers}).

\begin{table}[!th]
    \centering
    \resizebox{\textwidth}{!}{
    \begin{tabular}{c|c|c|c|c|c|c}
        Environment-dataset & $\epsilon$ & $w$ & $d$ & imagination horizon & world model train freq & imagination freq  \\ \hline
         Halfcheetah (R) & 0.15 & 5 & 10 & 5 & 1000 & 1000 \\
         DMC Walker (R) & 0.3 & 5 & 10 & 5 & 1000 & 1000 \\
         Hopper (R) & 0.1 & 50 & 3 & 3 & 1000 & 1000 \\
         Ant (R) & 0.025 & 50 & 3 & 3 & 250 & 250 \\
         DMC Walker (MR) & 0.3 & 5 & 10 & 5 & 1000 & 1000 \\
         Ant (MR) & 0.025 & 10 & 5 & 5 & 250 & 250 \\
         Humanoid (MR) & 0.005 & 50 & 3 & 3 & 250 & 250 \\
    \end{tabular}
    }
    \caption{Hyperparameters used for PTGOOD and base MBPO+SAC agent.}
    \label{tab:ptgood-hypers}
\end{table}

For the uncertainty-comparison experiments in \S\ref{sec:ucb-methods}, we measure the ``uncertainty'' of a given input as the average standard deviation across outputs from all members in the ensemble. For example, members of a ``Transition'' ensemble may each output a prediction for the next-state where $\hat{s} \in \mathbb{R}^6$ for a given $(s,a)$. Here, if the ensemble has $7$ members, uncertainty for $(s,a)$ is computed with $\frac{1}{6} \sum_{i=1}^6 \sqrt{\frac{\sum_{j=1}^7(\hat{S}_{j,i} - \mu_i)}{7}} $ where $\hat{S} \in \mathbb{R}^{7\times6}$ is a matrix whose entry $\hat{s}_{j,i}$ is $i$th value in the $j$th ensemble member's output, and $\mu_i$ is the mean value of the $i$th column of $\hat{S}$.

\subsection{The Conditional Entropy Bottleneck} \label{app:ceb}
The Conditional Entropy Bottleneck (CEB)~\citep{ceb} is an information-theoretic method for learning a representation $Z$ of input data $X$ useful for predicting target data $Y$. CEB's simplest formulation is to learn a $Z$ that minimizes $\beta I(X;Z \vert Y) - I(Z;Y)$, where $\beta$ is a weighting hyperparameter and $I(\cdot)$ denotes mutual information. Intuitively, CEB learns a representation that minimizes the extra information $Z$ captures about $X$ when $Y$ is known and maximizes the information $Z$ captures about $Y$. This form treats $X$ and $Y$ asymmetrically. Instead, the bidirectional CEB objective uses two separate representations $Z_X$ and $Z_Y$ for $X$ and $Y$, respectively:

\begin{equation}\label{eqn:bidir-ceb}
    \begin{aligned}
        \mathrm{CEB_{bidir}} \triangleq \mathrm{min} &-H(Z_X \vert  X) + H(Z_X \vert Y) + H(Y \vert Z_X) \\ 
        &-H(Z_Y \vert Y) + H(Z_Y \vert X) + H(X \vert Z_Y),
    \end{aligned}
\end{equation}
where $H(\cdot)$ and $H(\cdot \vert \cdot)$ are entropy and conditional entropy, respectively. We can form Equation~\ref{eqn:bidir-ceb} as a self-supervised objective via a noise function $X^{\prime} = f(X, U)$ with noise variable $U$, and treating the noised data $X^{\prime}$ as the target $Y$. Additionally,~\citet{ceb} show that we can place variational bounds on Equation~\ref{eqn:bidir-ceb} using a sampling distribution encoder $e(z_X \vert x)$, and variational approximations of the backwards encoder $b(z_{X^{\prime}} \vert x^{\prime})$, classifier $c(x^{\prime} \vert z_X)$, and decoder $d(x \vert z_{X^{\prime}})$ distributions. At convergence, we learn a unified representation that is consistent with both $z_{X}$ and $z_{X^{\prime}}$ by applying the CEB objective in both directions with the original and noised data:

\begin{equation}\label{eqn:denoise-ae-ceb}
    \begin{aligned}
        \mathrm{min} \; &\langle \mathrm{log} \: e(z_X \vert x) \rangle - \langle \mathrm{log} \: b(z_X \vert x^{\prime}) \rangle - \langle \mathrm{log} \: c(x^{\prime}  \vert z_X) \rangle \\
        &+ \langle \mathrm{log} \: b(z_{X^{\prime}} \vert x^{\prime}) \rangle - \langle \mathrm{log} \: e(z_{X^{\prime}} \vert x) \rangle - \langle \mathrm{log} \: d(x \vert z_{X^{\prime}}) \rangle,
    \end{aligned}
\end{equation}
where each  $\langle \cdot \rangle$ denotes the expectation over the joint distribution $p(x, x^{\prime}, u, z_X, z_{X^{\prime}})=p(x)p(u)p(x^{\prime} \vert f(x,u))e(z_X \vert x)b(z_{X^{\prime}} \vert x^{\prime})$. We refer the reader to the original CEB paper for more details.~\citet{ceb} show that we do not need to learn parameters for $c(\cdot)$ in Equation~\ref{eqn:denoise-ae-ceb} because $c(x^{\prime} \vert z_X) \propto b(z_X \vert x^{\prime})p(z_{X^{\prime}})$, which can be simplified further by marginalizing $p(z_{X^{\prime}})$ over a minibatch of size $K$. The same can be done for $d(\cdot)$ using $e(\cdot)$. Altogether, this forms the contrastive ``CatGen" formulation with the following upper bound:

\begin{equation}\label{eqn:catgen-ceb}
    \begin{aligned}
        \mathrm{CEB_{denoise}} \leq \mathrm{min}_{e(\cdot), b(\cdot)} \: &\mathbb{E} \left[  \mathbb{E}_{z_X \sim e(z_X \vert x)} [ \beta \: \mathrm{log} \: \frac{e(z_X \vert x)}{b(z_X \vert x^{\prime})} - \mathrm{log} \: \frac{b(z_X \vert x^{\prime})}{\frac{1}{K} \sum_{i=1}^K b(z_X \vert x^{\prime}_i)} ] \right. \\
        &\left. + \mathbb{E}_{z_{X^{\prime}} \sim b(z_{X^{\prime}} \vert x^{\prime})} [ \beta \: \mathrm{log} \: \frac{b(z_{X^{\prime}} \vert x^{\prime})}{e(z_{X^{\prime}} \vert x)} - \mathrm{log} \: \frac{e(z_{X^{\prime}} \vert x)}{\frac{1}{K} \sum_{i=1}^K e(z_{X^{\prime}} \vert x_i)} ] \right]
    \end{aligned}
\end{equation}
where the outer expectation is over the joint distribution $x,x^{\prime} \sim p(x, x^{\prime}, u, z_X, z_{X^{\prime}})$.

\section{Suboptimal Convergence}\label{app:subopt-convergence}

\begin{figure}[t]
    \centering
    \begin{subfigure}{0.49\textwidth}
        \includegraphics[width=\textwidth]{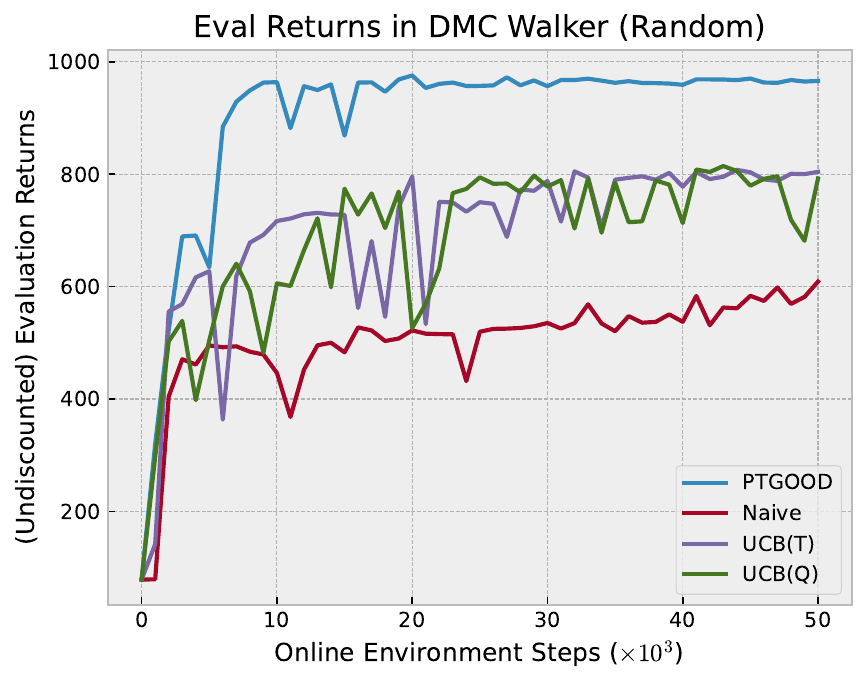}
    \end{subfigure}
    \begin{subfigure}{0.49\textwidth}
        \includegraphics[width=\textwidth]{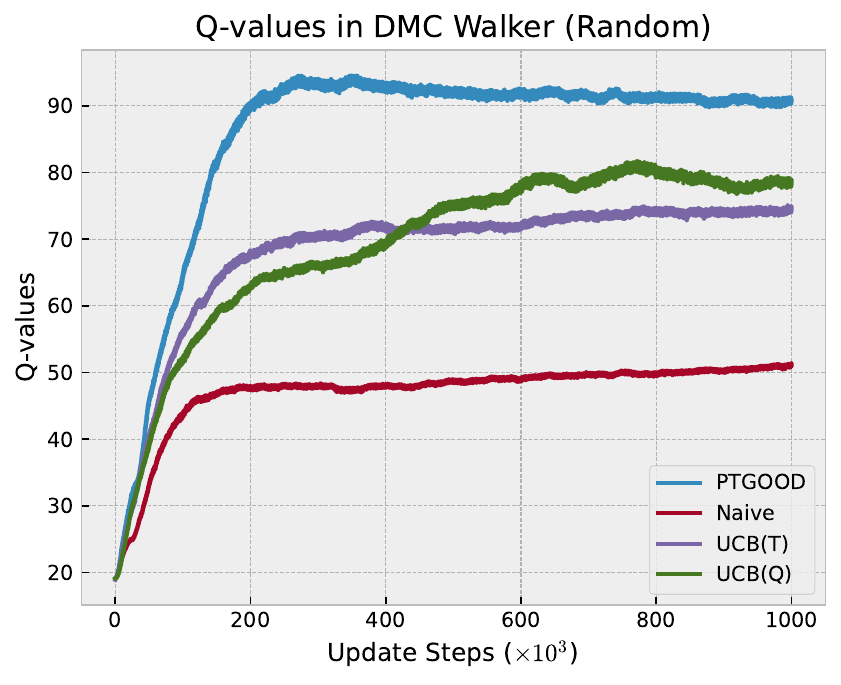}
    \end{subfigure}
    \begin{subfigure}{0.49\textwidth}
        \includegraphics[width=\textwidth]{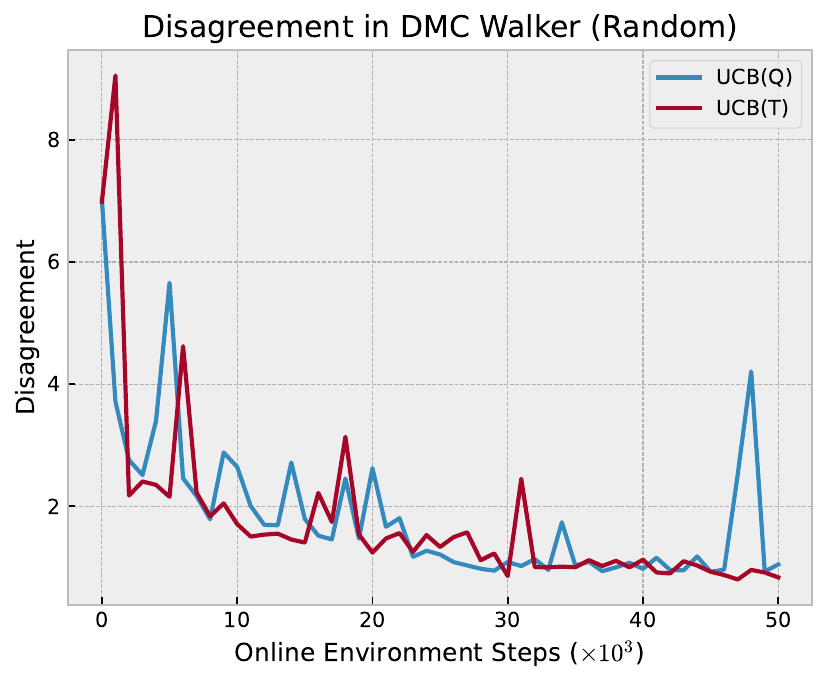}
    \end{subfigure}
    \begin{subfigure}{0.49\textwidth}
        \includegraphics[width=\textwidth]{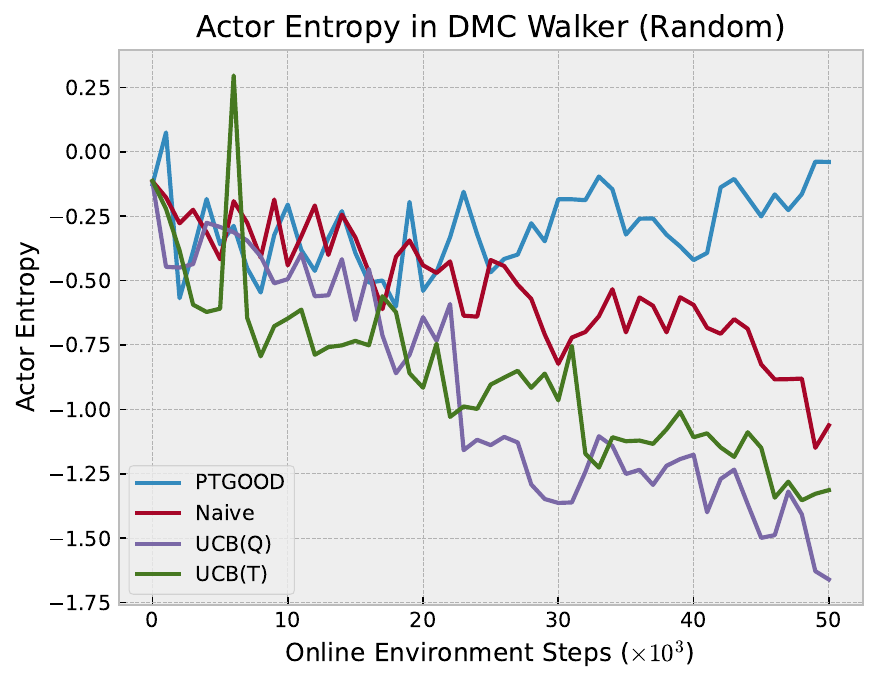}
    \end{subfigure}
    \caption{Metrics collected over 50k steps on online fine-tuning for the premature convergence experiment in DMC Walker (Random).}
    \label{fig:prem-converge-random}
\end{figure}

\begin{figure}[t]
    \centering
    \begin{subfigure}{0.49\textwidth}
        \includegraphics[width=\textwidth]{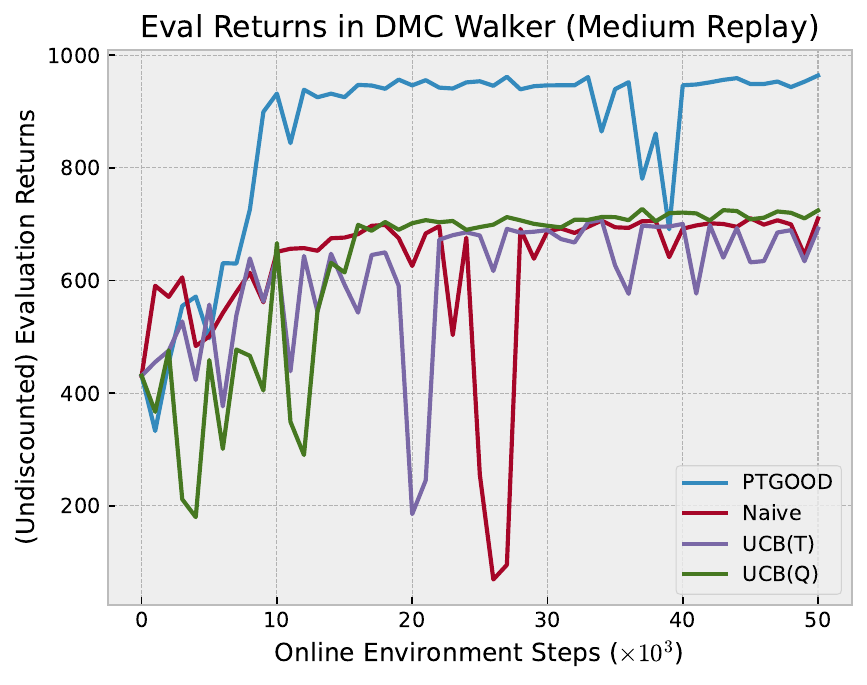}
    \end{subfigure}
    \begin{subfigure}{0.49\textwidth}
        \includegraphics[width=\textwidth]{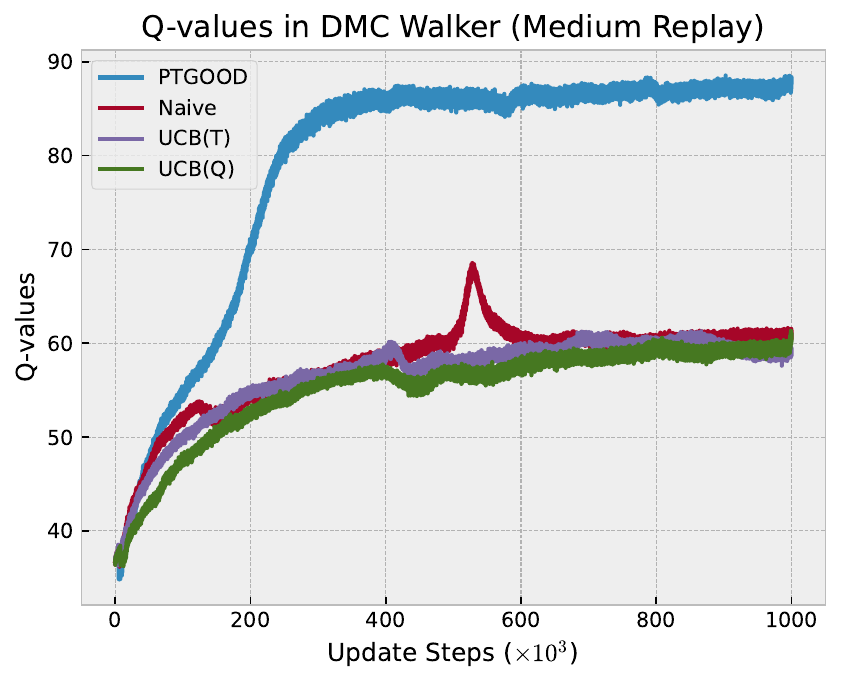}
    \end{subfigure}
    \begin{subfigure}{0.49\textwidth}
        \includegraphics[width=\textwidth]{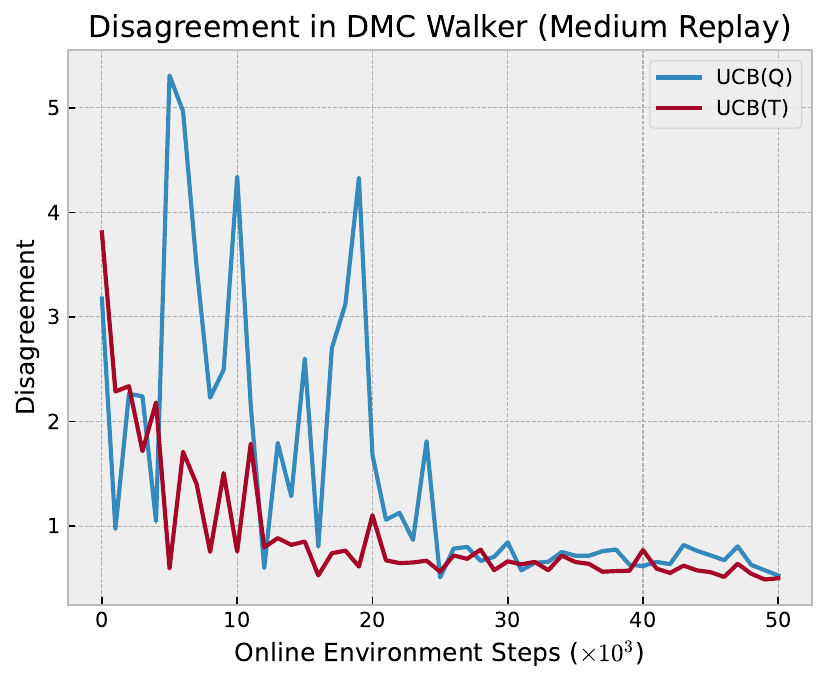}
    \end{subfigure}
    \begin{subfigure}{0.49\textwidth}
        \includegraphics[width=\textwidth]{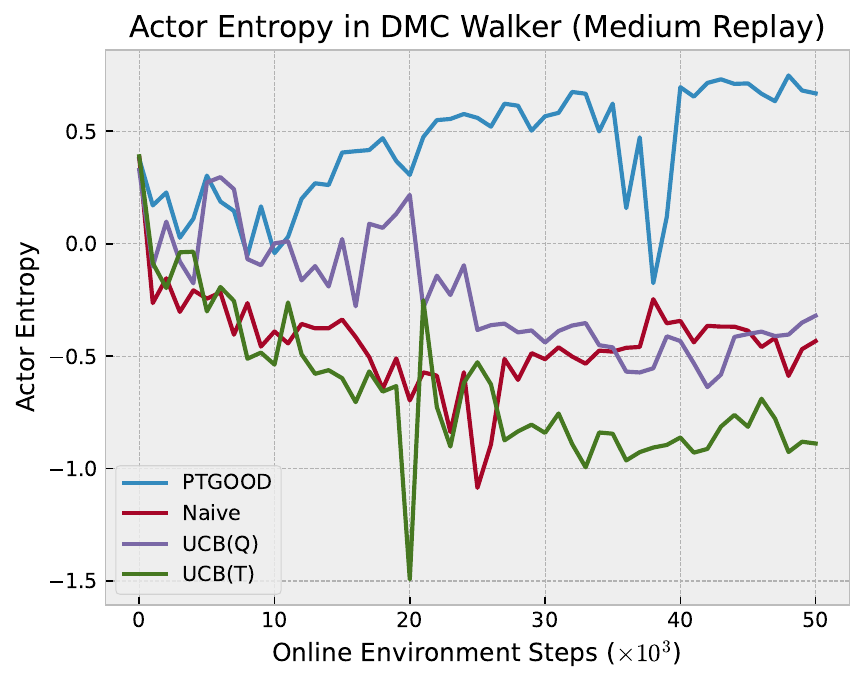}
    \end{subfigure}
    \caption{Metrics collected over 50k steps on online fine-tuning for the premature convergence experiment in DMC Walker (Medium Replay).}
    \label{fig:prem-converge-mr}
\end{figure}

We highlight that many of our baselines' policies converge prematurely to suboptimal returns in both DMC Walker datasets. To help explain the phenomenon and describe how PTGOOD avoids this issue, we examine several metrics throughout online fine-tuning. Specifically, for UCB-style baselines, we examine ensemble disagreement and policy entropy. UCB-style methods sample the policy to create the set of actions over which disagreement is evaluated. Therefore, both of these metrics drive exploration. For the other methods, such as No Pretrain and Naive, we examine only policy entropy. For these methods, the policies are sampled for action selection during online fine-tuning, and, therefore, its entropy is important for exploration. Both policy entropy and disagreement are captured during the evaluation episodes rolled out every 1k steps during online fine-tuning. We also capture average Q-values of each mini-batch used during agent training and evaluation returns. Finally, we collect all metrics except for disagreement for a PTGOOD agent. Figure~\ref{fig:prem-converge-random} and Figure~\ref{fig:prem-converge-mr} show these metrics for DMC Walker (Random) and DMC Walker (Medium Replay), respectively. 

We highlight that the disagreement metric for both UCB methods in both environment-dataset combinations starts relatively high but quickly collapses to a low number roughly around the time evaluation returns converge. Also, we note that the policy entropy of both UCB agents and the naive agent shows a consistent downward trend in both environment-dataset combinations. Such a reducing entropy will reduce the diversity in the action sets used for exploration in all three of these methods. In contrast, the PTGOOD agents' policy entropy remains relatively high throughout online fine-tuning. 

Next, we show that the reduced exploration mentioned above causes the three baselines to miss exploring the same regions of the state-action space that PTGOOD explores. We demonstrate this by showing that the baselines' critics undervalue the state-action pairs collected by a higher-return PTGOOD agent and overvalue the state-action pairs that they themselves collect. If the baselines were to explore as well as PTGOOD, such erroneous Q-values would not exist. At the end of online fine-tuning, we collect 10 episodic trajectories of state-action pairs from each of the four agents. For their returns, see Figure~\ref{fig:prem-converge-random} and Figure~\ref{fig:prem-converge-mr}. Table~\ref{tab:q-value-prem-collapse} displays the average Q-values over the trajectories for each baseline in each environment-dataset combination.

\begin{table}[]
    \centering
    \resizebox{\textwidth}{!}{
    \begin{tabular}{c|c|c|c}
    Dataset & Baseline & Q-value on PTGOOD trajectory & Q-value on own trajectory \\ \hline
     MR   & Naive & 51.1 $\pm$ 4.3 & 62.6 $\pm$ 2.8 \\
      MR  & UCB(T) & 48.6 $\pm$ 3.1 & 63.4 $\pm$ 3.7 \\
       MR & UCB(Q) & 53.3 $\pm$ 2.9 & 64.1 $\pm$ 3.8 \\
       R & Naive & 46.8 $\pm$ 2.8 & 54.1 $\pm$ 4.3 \\
       R & UCB(T) & 71.6 $\pm$ 3.9 & 79.7 $\pm$ 3.6 \\
       R & UCB(Q) & 68.7 $\pm$ 2.4 & 79.2 $\pm$ 4.1 
    \end{tabular}
    }
    \caption{Q-value over trajectory comparison for the premature convergence experiment.}
    \label{tab:q-value-prem-collapse}
\end{table}

\clearpage
\section{Investigating Cal-QL and other Cal-QL Environments}\label{app:cal-ql-study}
Here we benchmark PTGOOD, Cal-QL, PEX, PROTO, and Scratch (same as No Pretrain) on two of the datasets provided by the Cal-QL authors in the Adroit environments. The Cal-QL authors altered the base Adroit environments to be ``sparse-like''. That is, their reward function is $R: \mathcal{S} \times \mathcal{A} \rightarrow \{-5, 5\}$. We specifically chose these environments because the dataset are ``narrow'' in the sense that the information about the MDP contained within the datasets is a very small subset of all possible information contained in the MDP. Due to this characteristic, the offline pre-training phase is unlikely to be useful. In such a case, our dataset selection criterion (b) is violated, which we hypothesize would cause our Scratch (same as No Pretrain) baseline to be tough to beat. 

\begin{figure*}[t]
    \centering
    \begin{subfigure}{0.49\textwidth}
        \includegraphics[width=\textwidth]{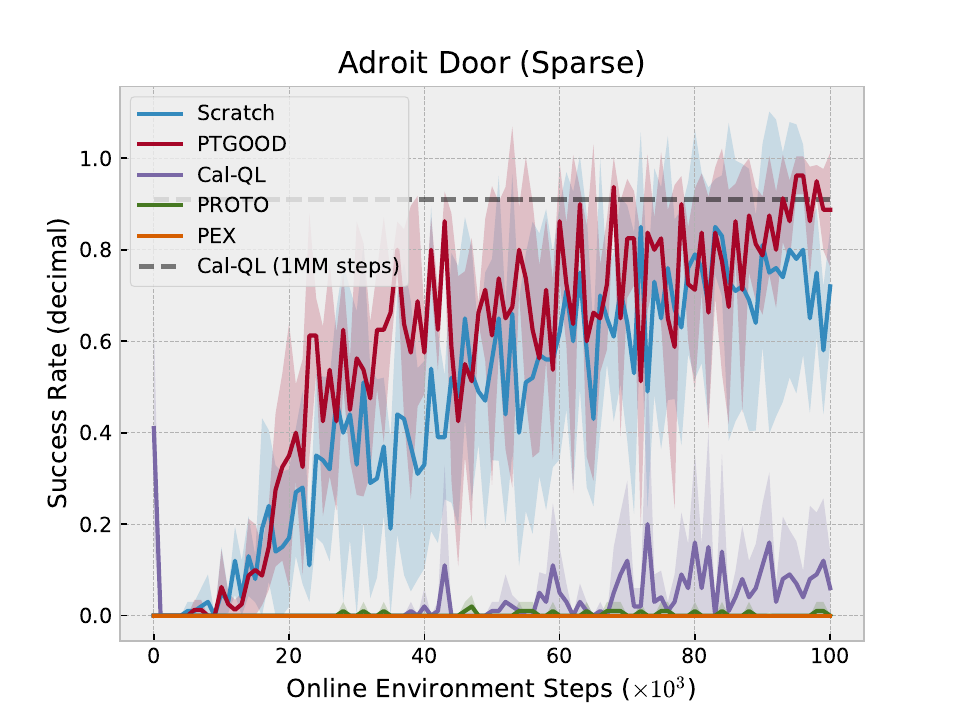}
    \end{subfigure}
    \begin{subfigure}{0.49\textwidth}
        \includegraphics[width=\textwidth]{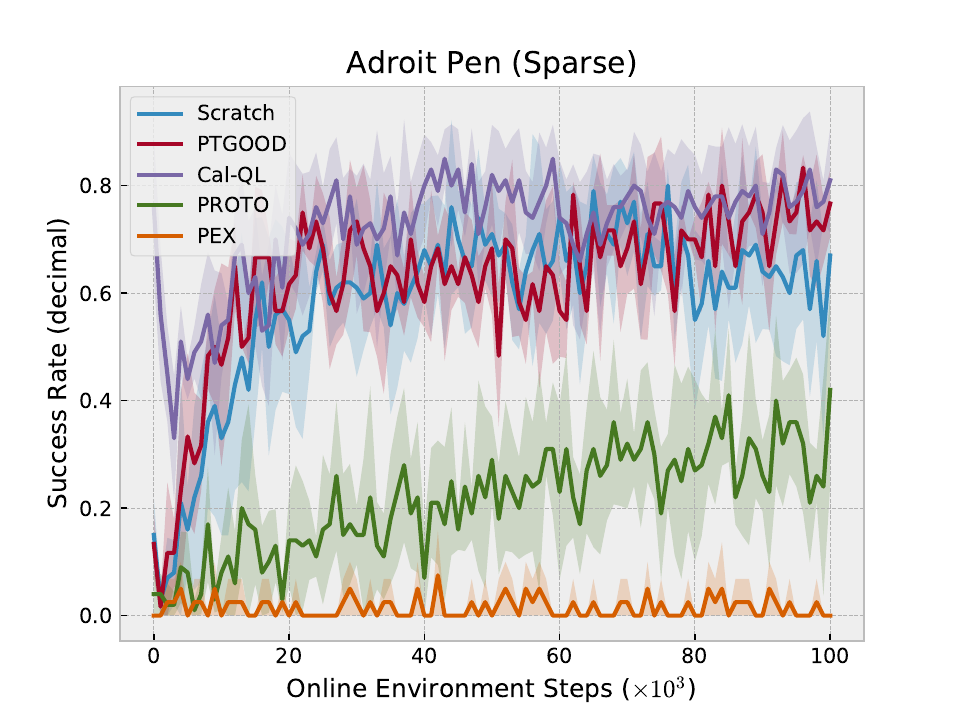}
    \end{subfigure}
    \caption{Undiscounted evaluation returns over 100 thousand environment steps in the sparse-like Adroit environments from the Cal-QL paper.}
    \label{fig:adroit-envs}
\end{figure*}

We highlight that our hypothesis is confirmed when comparing PTGOOD and Scratch (same as No Pretrain) results in Figure~\ref{fig:adroit-envs}.

\begin{figure*}[t]
    \centering
    \begin{subfigure}{0.32\textwidth}
        \includegraphics[width=\textwidth]{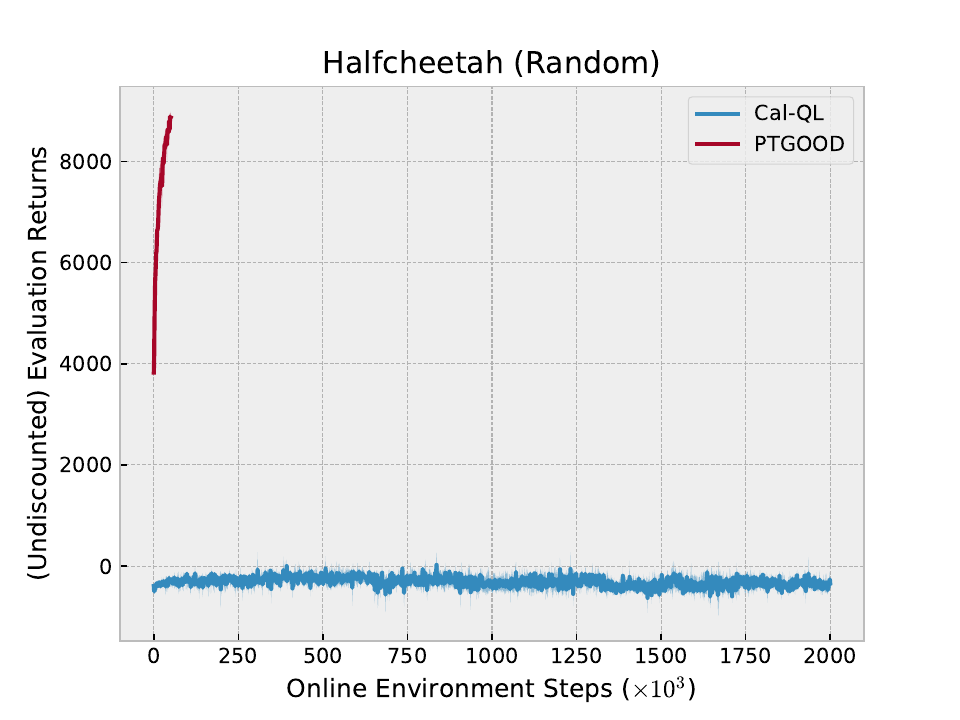}
    \end{subfigure}
    \begin{subfigure}{0.32\textwidth}
        \includegraphics[width=\textwidth]{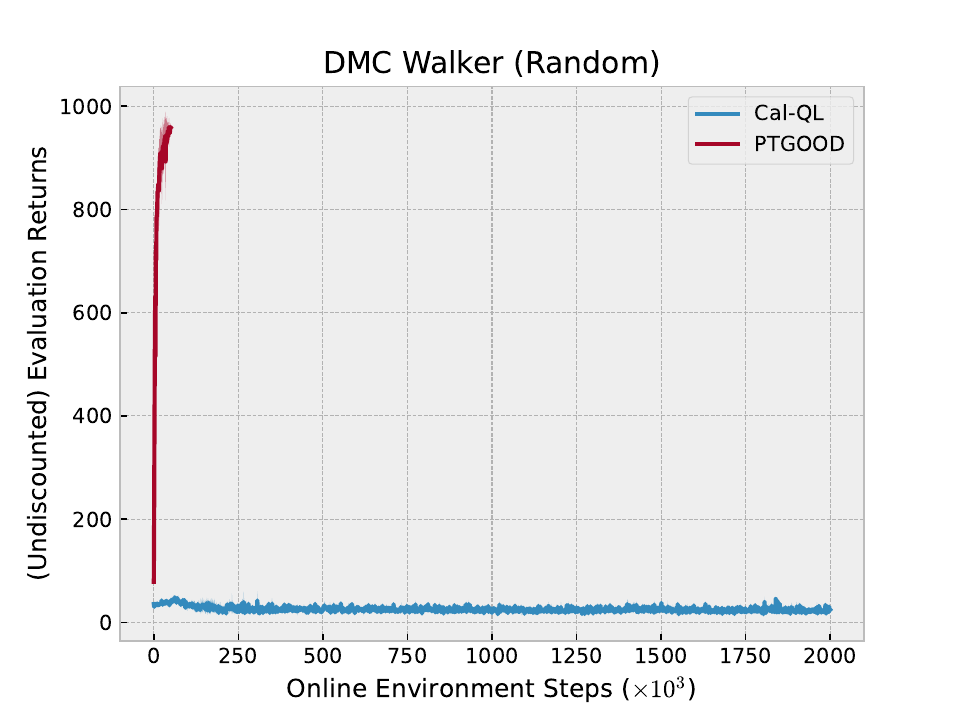}
    \end{subfigure}
    \begin{subfigure}{0.32\textwidth}
        \includegraphics[width=\textwidth]{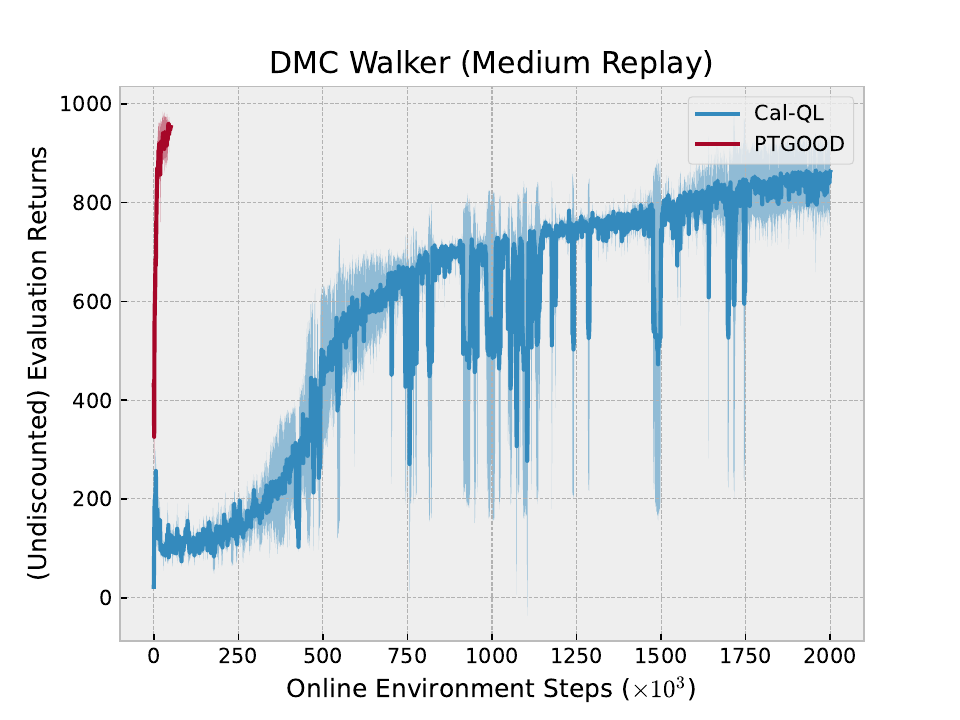}
    \end{subfigure} \hfill
    \begin{subfigure}{0.32\textwidth}
        \includegraphics[width=\textwidth]{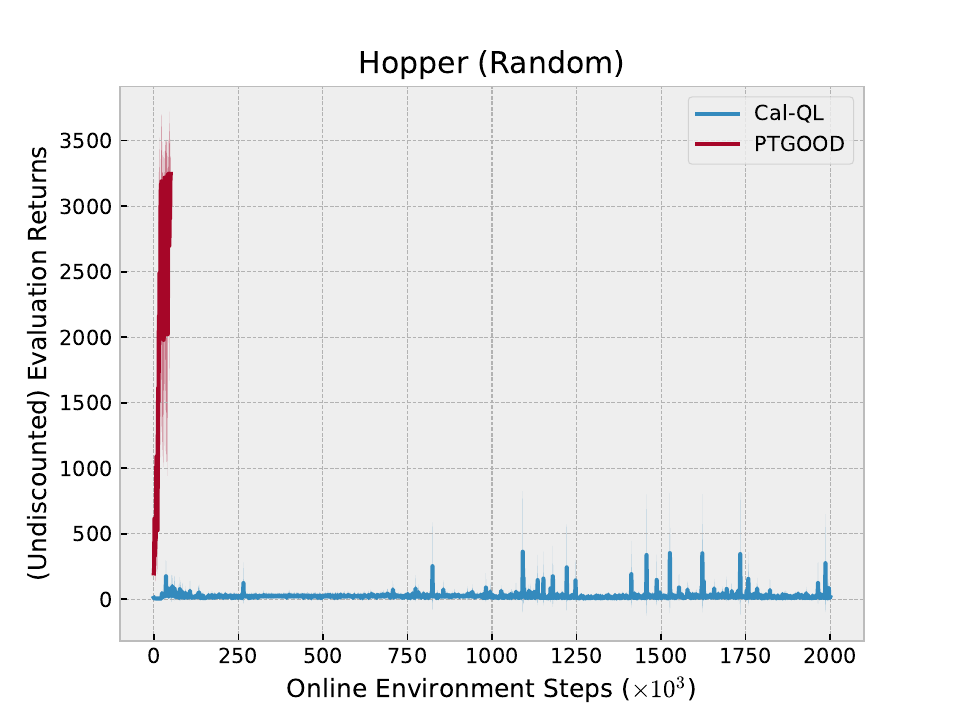}
    \end{subfigure}
    \begin{subfigure}{0.32\textwidth}
        \includegraphics[width=\textwidth]{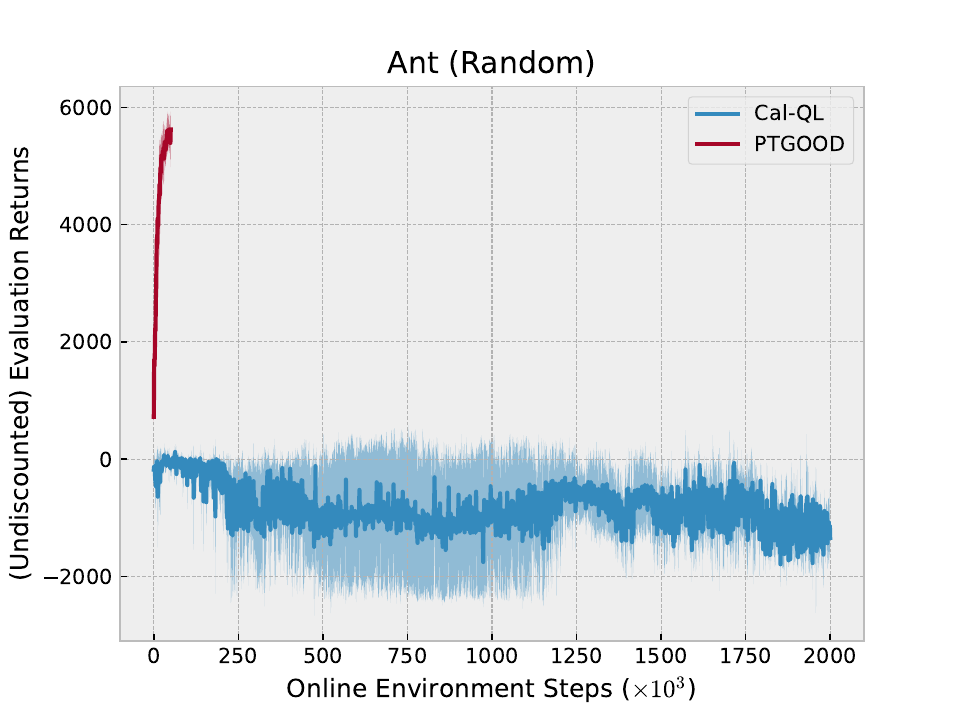}
    \end{subfigure}
    \begin{subfigure}{0.32\textwidth}
        \includegraphics[width=\textwidth]{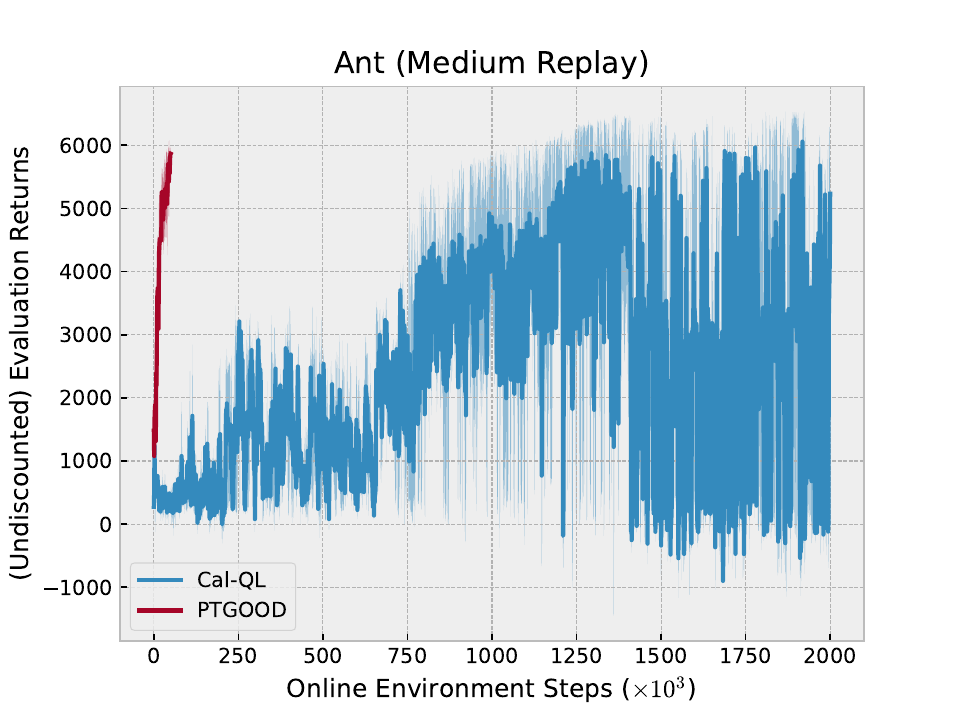}
    \end{subfigure}
    \begin{subfigure}{0.32\textwidth}
        \includegraphics[width=\textwidth]{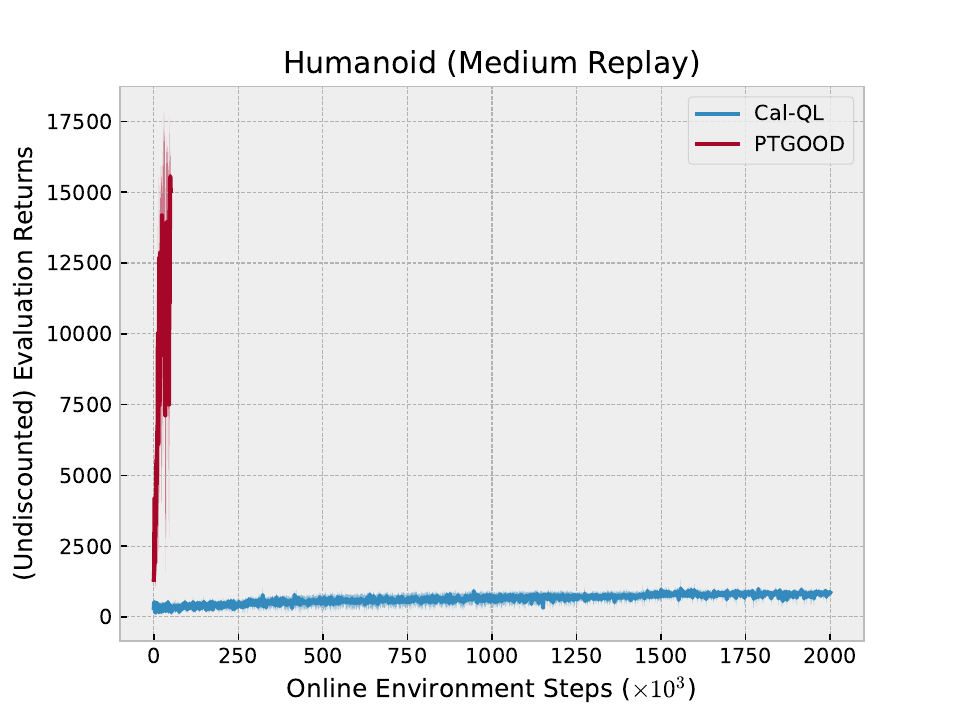}
    \end{subfigure}
    \caption{Undiscounted evaluation returns for Cal-QL over two million online steps versus 50 thousand online steps for PTGOOD.}
    \label{fig:calql-more-steps}
\end{figure*}

Next, we examine Cal-QL's performance in the datasets used in the main study but with many more (2 million) online finetuning steps allowed. Figure~\ref{fig:calql-more-steps} shows that Cal-QL struggles to learn much in any of the (Random) datasets and in Humanoid (Medium Replay). However, in the remaining (Medium Replay) datasets, Cal-QL does eventually find the optimal policy. 

\clearpage
\section{Additional Results}\label{app:additional-results}
\begin{figure*}[t]
    \centering
    \begin{subfigure}{0.32\textwidth}
        \includegraphics[width=\textwidth]{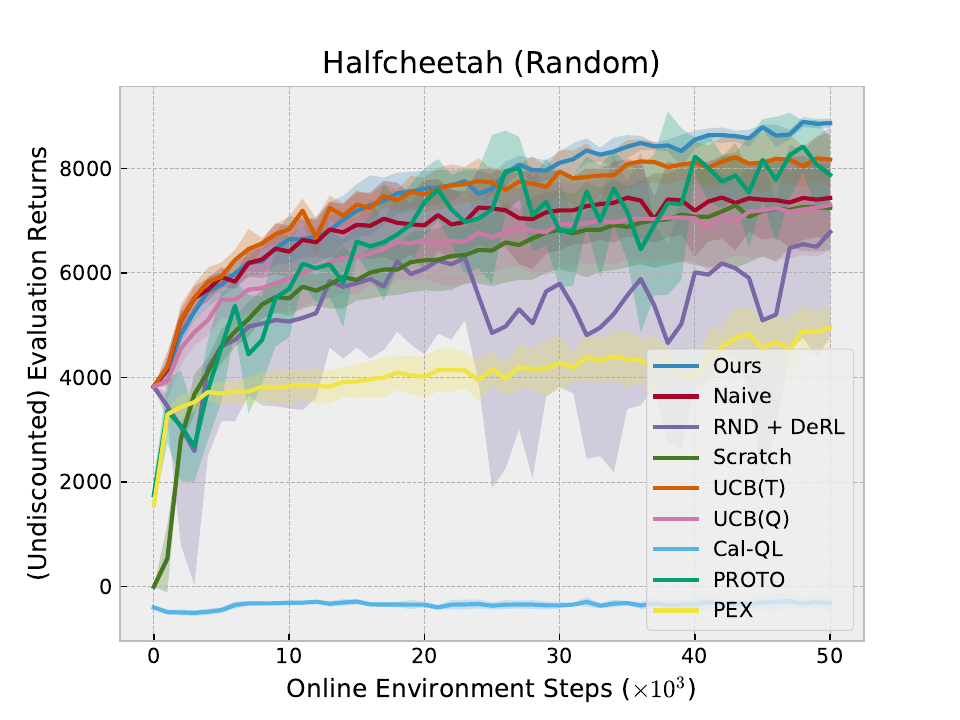}
    \end{subfigure}
    \begin{subfigure}{0.32\textwidth}
        \includegraphics[width=\textwidth]{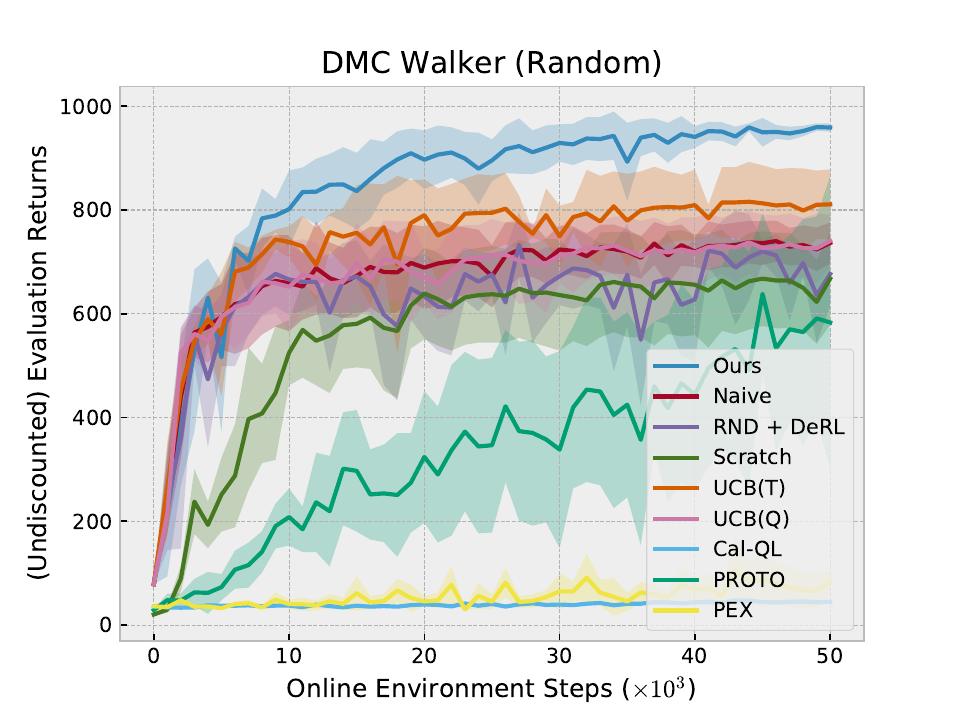}
    \end{subfigure}
    \begin{subfigure}{0.32\textwidth}
        \includegraphics[width=\textwidth]{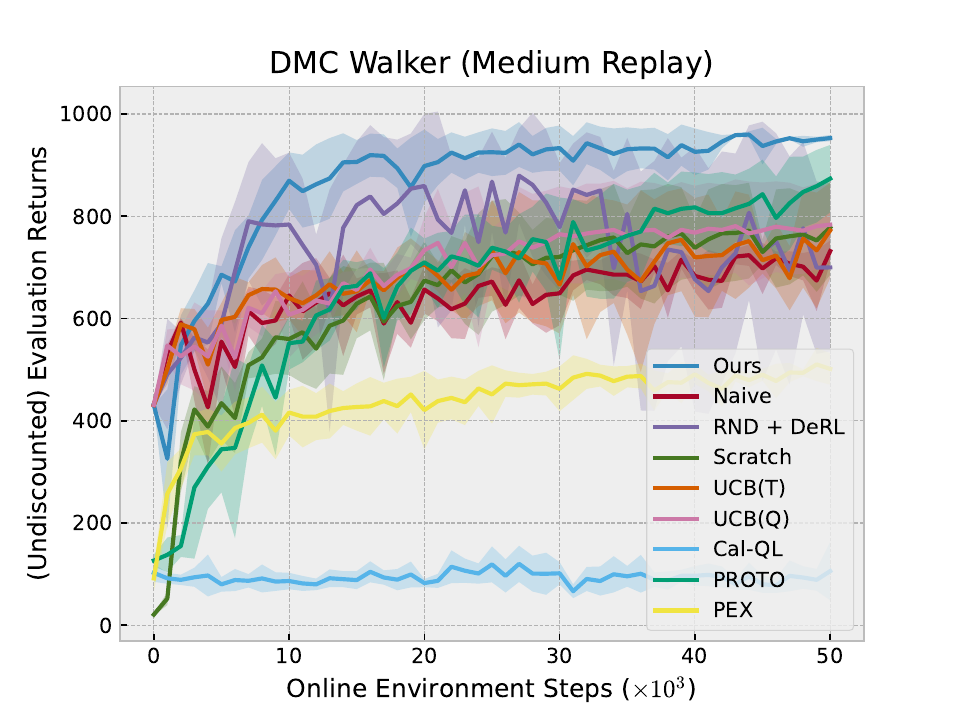}
    \end{subfigure} \hfill
    \begin{subfigure}{0.32\textwidth}
        \includegraphics[width=\textwidth]{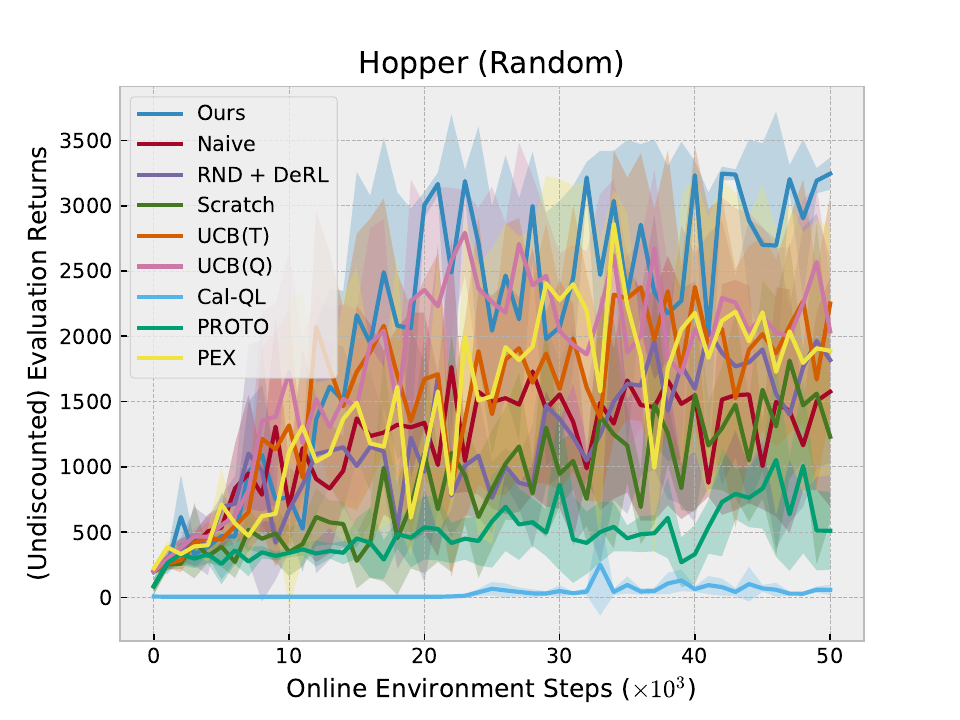}
    \end{subfigure}
    \begin{subfigure}{0.32\textwidth}
        \includegraphics[width=\textwidth]{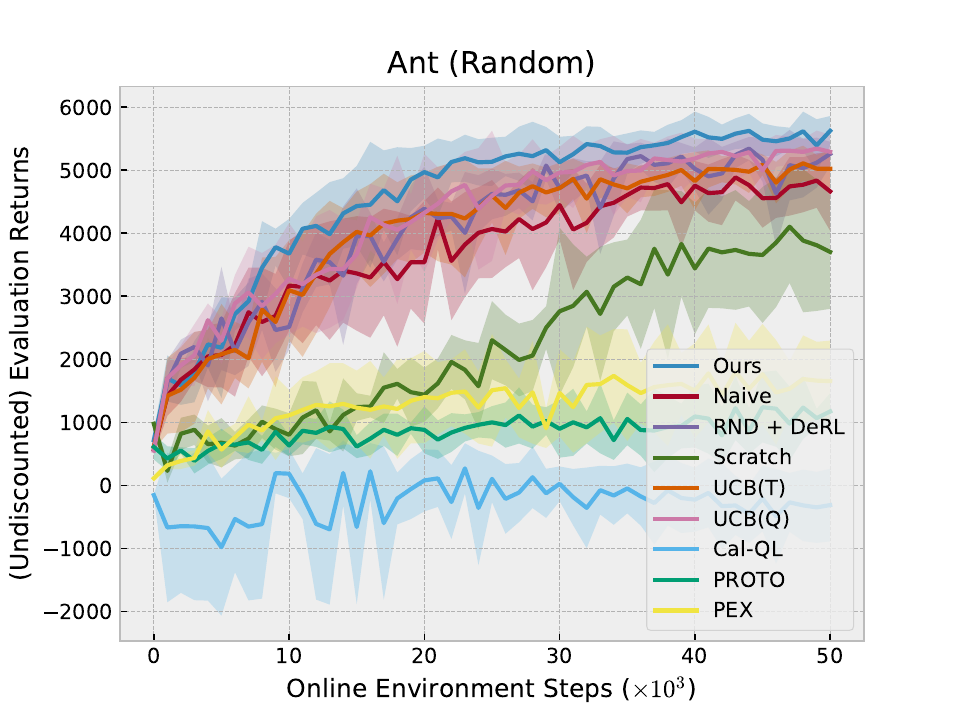}
    \end{subfigure}
    \begin{subfigure}{0.32\textwidth}
        \includegraphics[width=\textwidth]{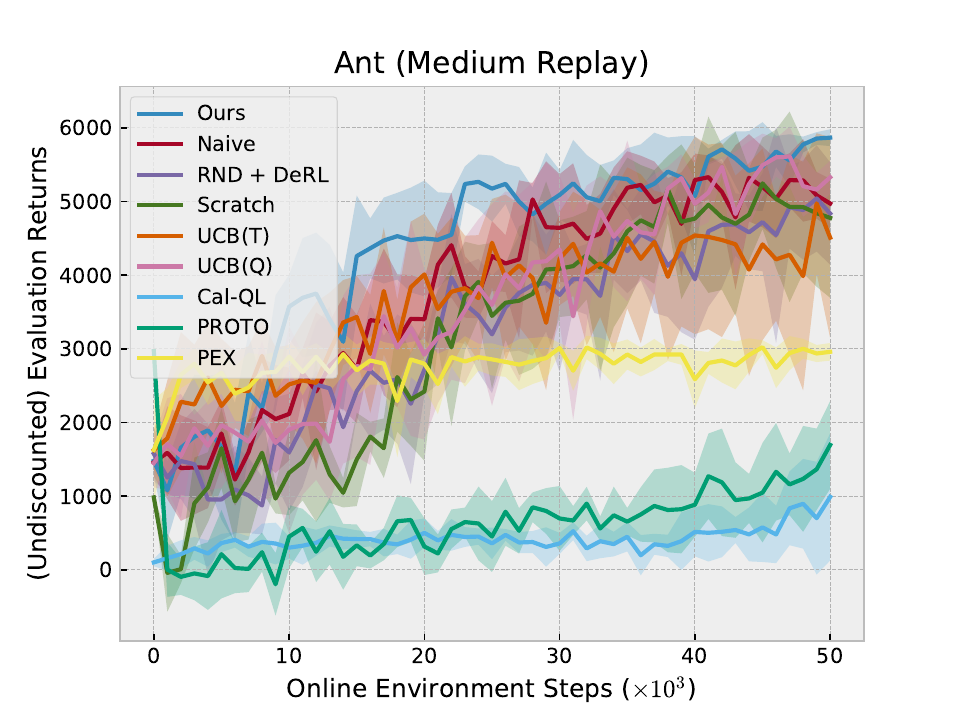}
    \end{subfigure}
    \begin{subfigure}{0.32\textwidth}
        \includegraphics[width=\textwidth]{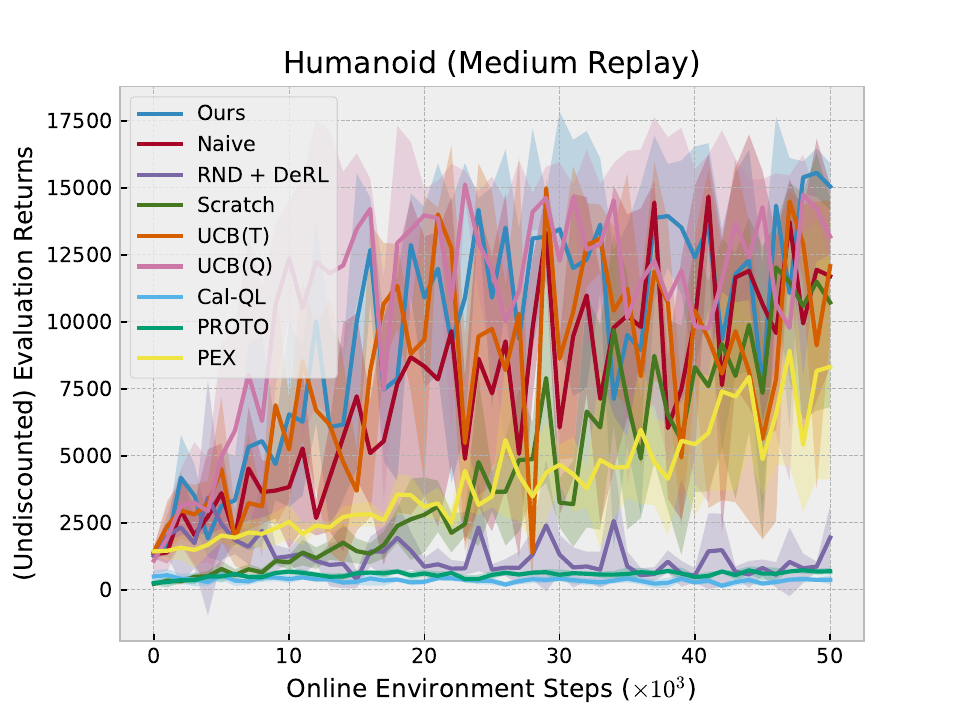}
    \end{subfigure}
    \caption{Undiscounted evaluation returns for all algorithms over the 50k online fine-tuning stage. Average (bold) $\pm$ one standard deviation (shaded area) displayed. Scratch is the same as No Pretrain.}
    \label{fig:all-algos-full-results}
\end{figure*}

\begin{figure*}[t]
    \centering
    \begin{subfigure}{0.32\textwidth}
        \includegraphics[width=\textwidth]{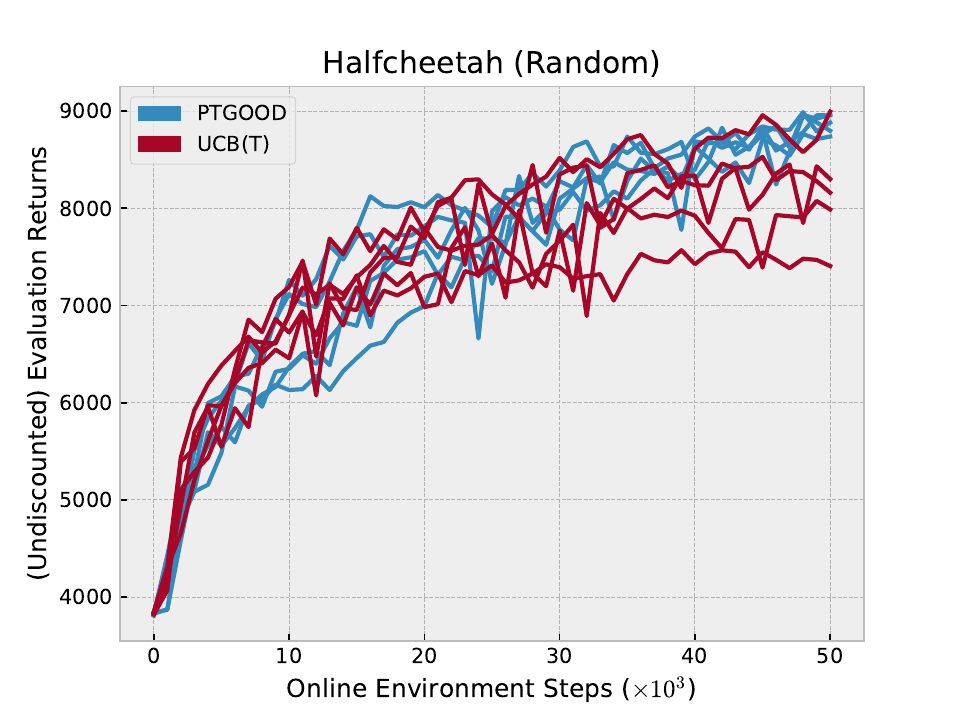}
    \end{subfigure}
    \begin{subfigure}{0.32\textwidth}
        \includegraphics[width=\textwidth]{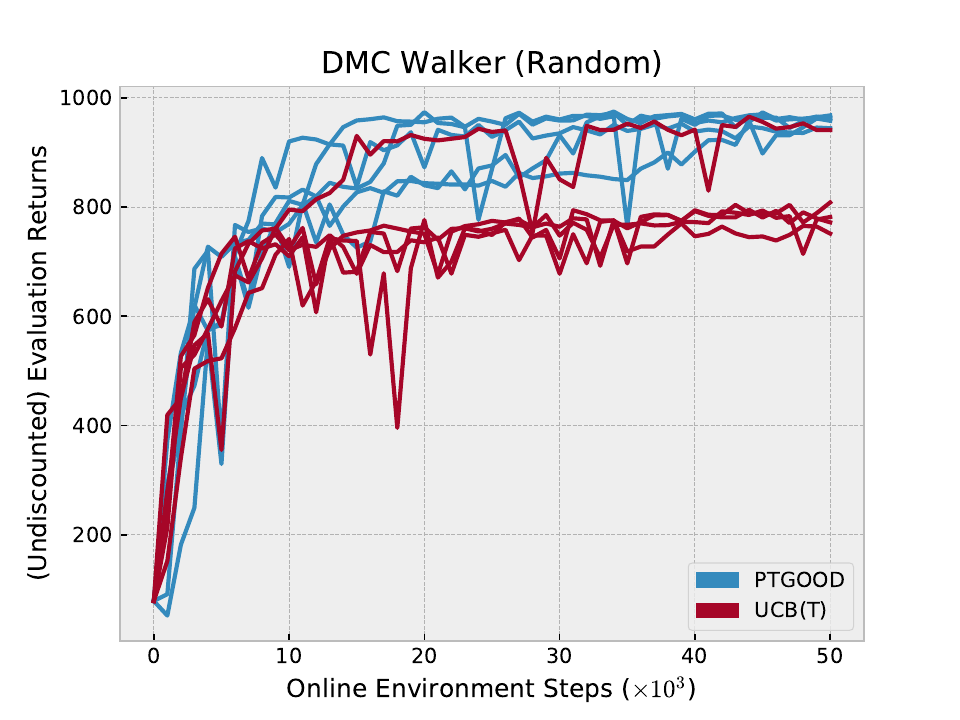}
    \end{subfigure}
    \begin{subfigure}{0.32\textwidth}
        \includegraphics[width=\textwidth]{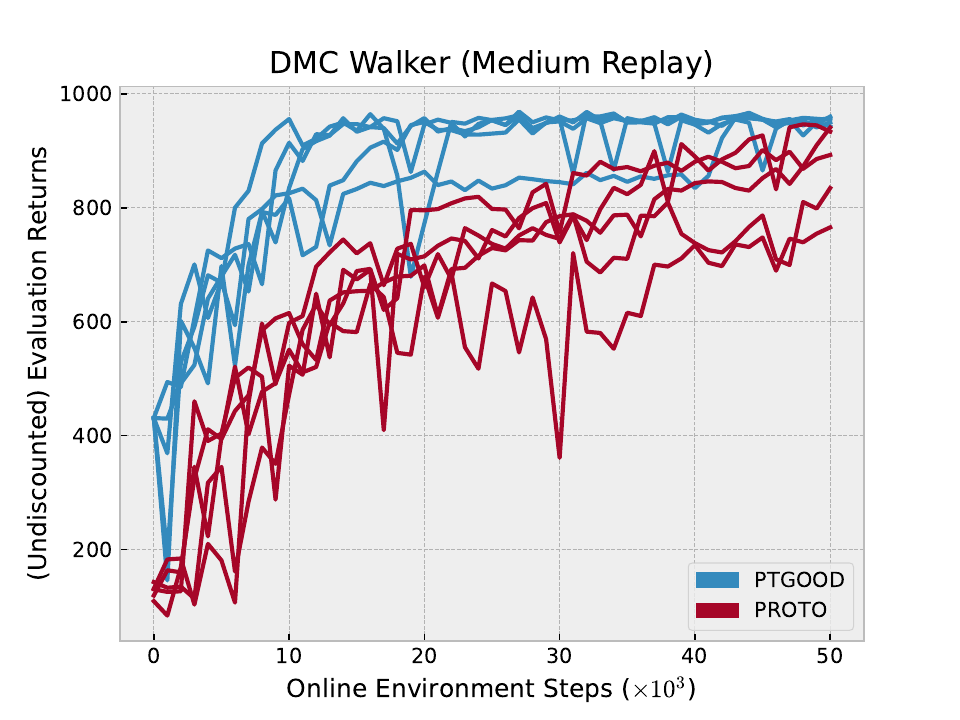}
    \end{subfigure} \hfill
    \begin{subfigure}{0.32\textwidth}
        \includegraphics[width=\textwidth]{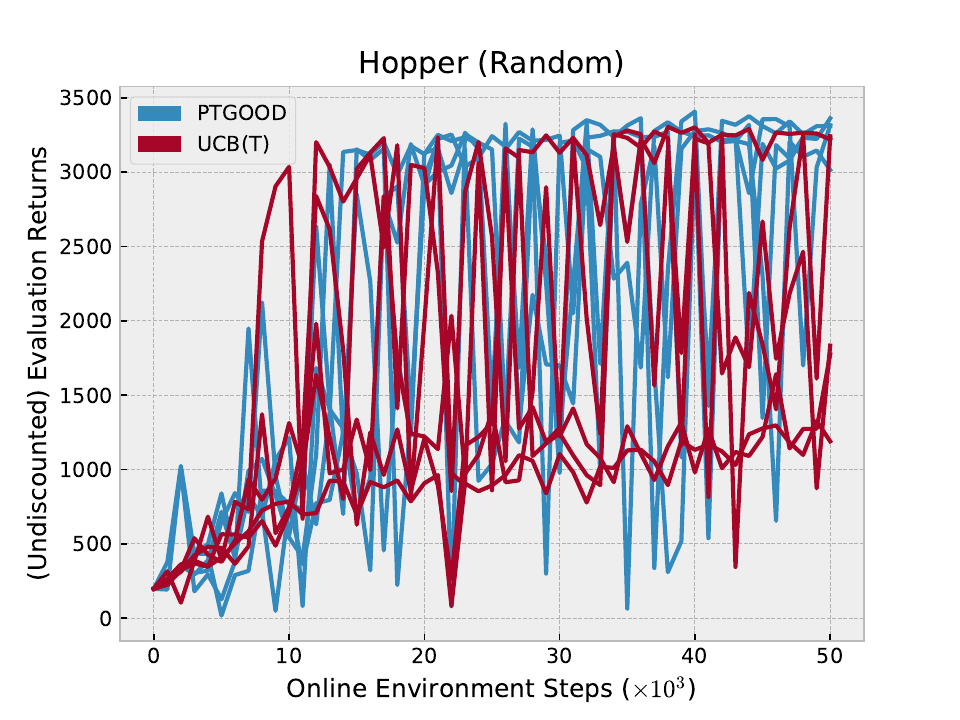}
    \end{subfigure}
    \begin{subfigure}{0.32\textwidth}
        \includegraphics[width=\textwidth]{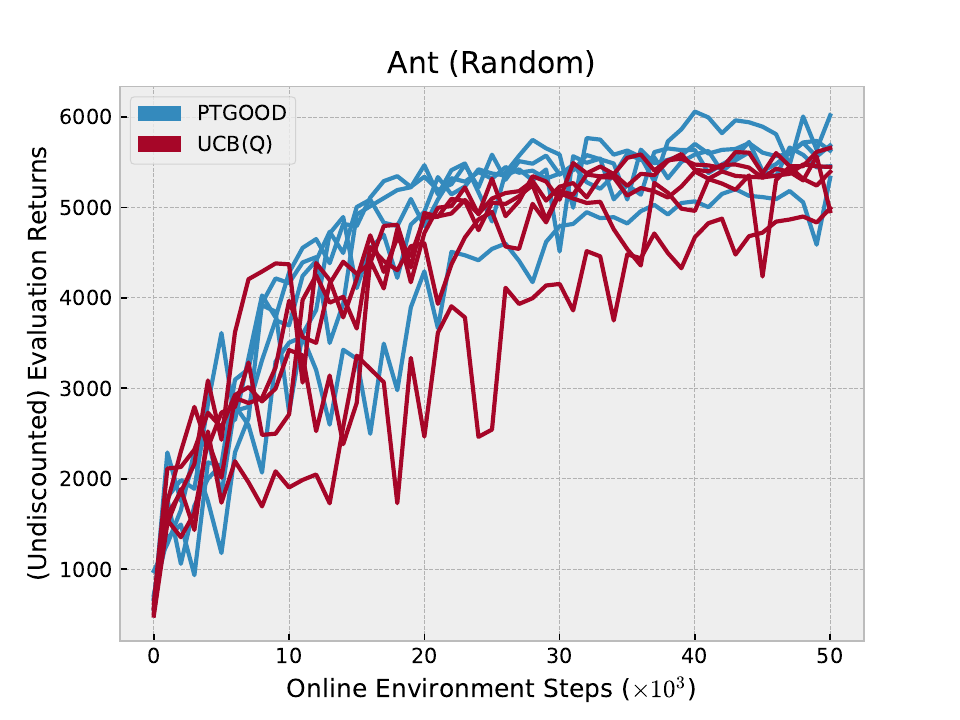}
    \end{subfigure}
    \begin{subfigure}{0.32\textwidth}
        \includegraphics[width=\textwidth]{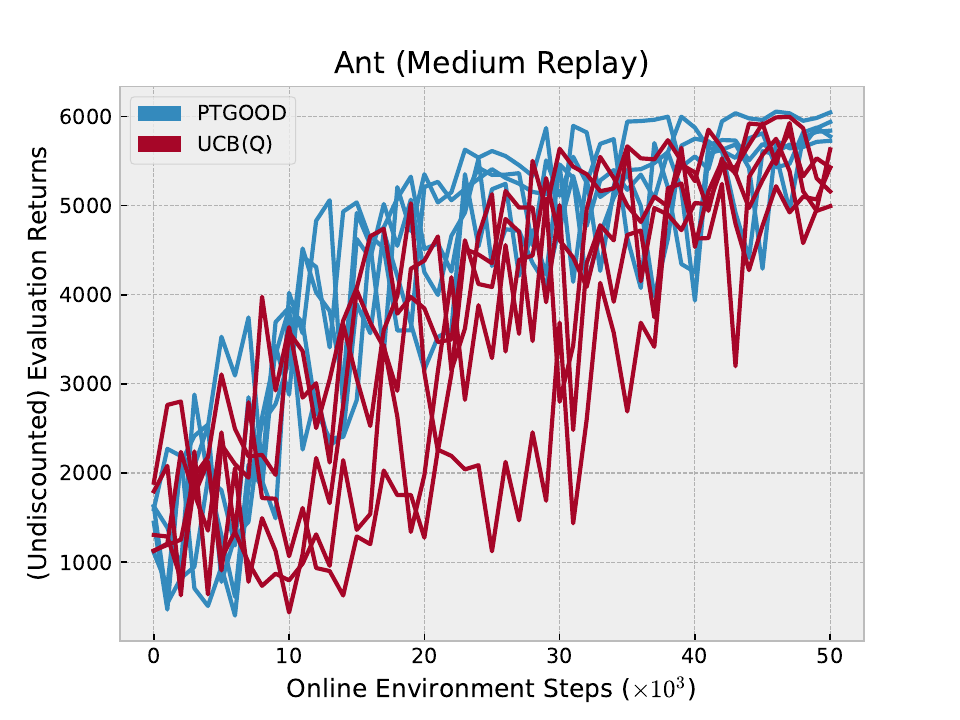}
    \end{subfigure}
    \begin{subfigure}{0.32\textwidth}
        \includegraphics[width=\textwidth]{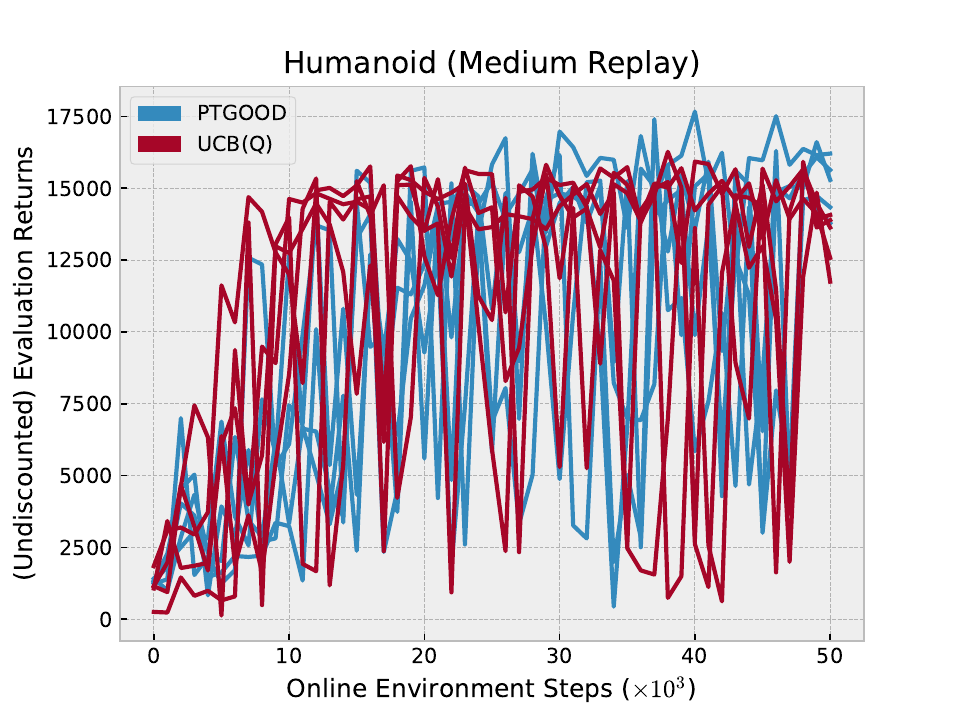}
    \end{subfigure}
    \caption{Undiscounted evaluation returns in all five training runs for the best and second-best performing algorithms over the 50k online fine-tuning stage.}
    \label{fig:first-second-compare}
\end{figure*}

Here, we present the full evaluation curves for all algorithms in all environment-dataset combinations in Figure~\ref{fig:all-algos-full-results}. Also, we provide the full evaluation curves for all seeds for the best and second-best performing algorithms in all environment-dataset combinations in Figure~\ref{fig:first-second-compare}.

\clearpage
\section{More Planning Noise Ablations}\label{app:planning-noise-extended}
\begin{figure*}[t]
    \centering
    \begin{subfigure}{0.32\textwidth}
        \includegraphics[width=\textwidth]{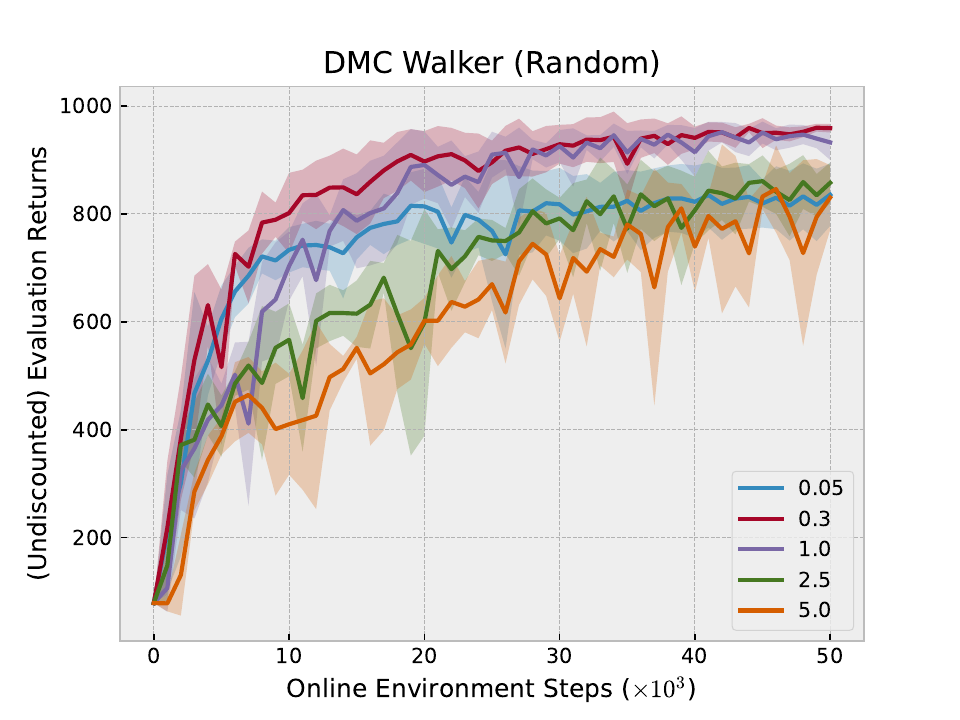}
    \end{subfigure}
    \begin{subfigure}{0.32\textwidth}
        \includegraphics[width=\textwidth]{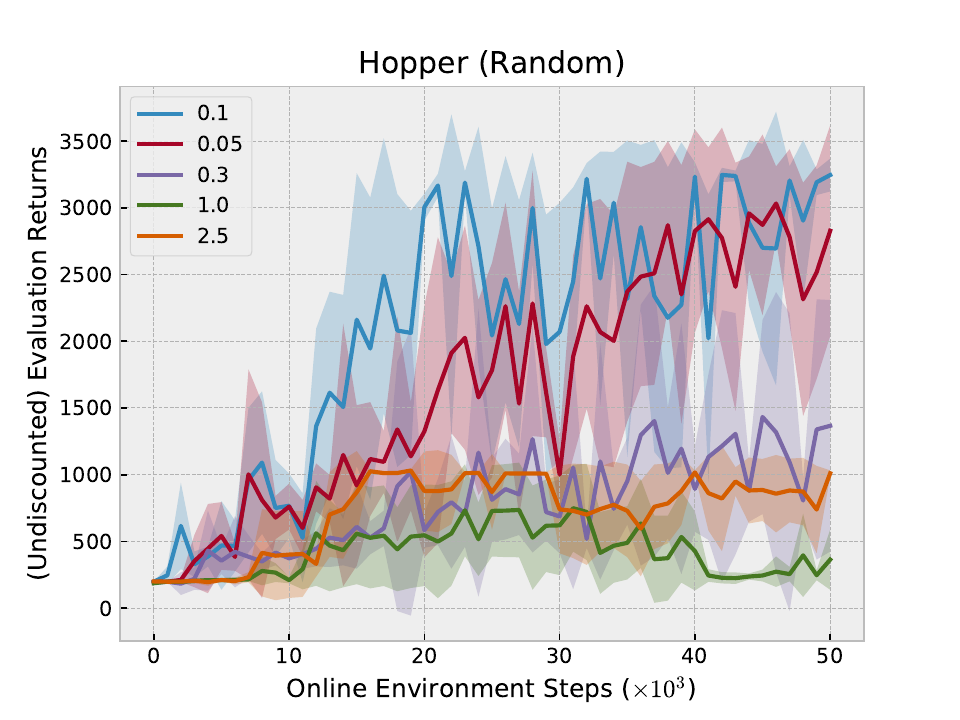}
    \end{subfigure}
    \begin{subfigure}{0.32\textwidth}
        \includegraphics[width=\textwidth]{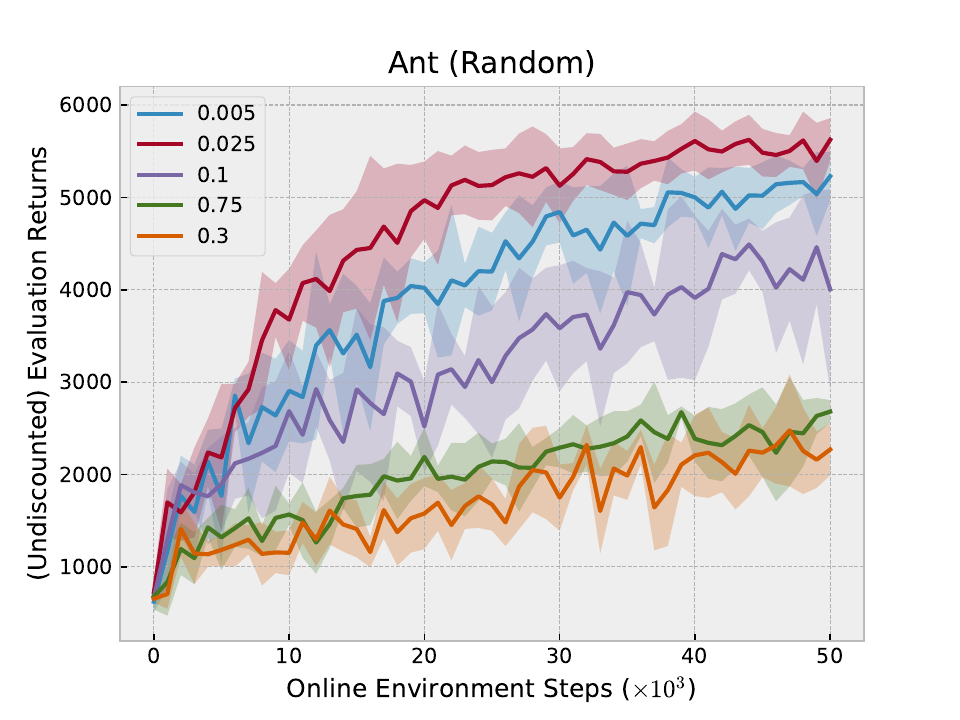}
    \end{subfigure} \hfill
    \begin{subfigure}{0.32\textwidth}
        \includegraphics[width=\textwidth]{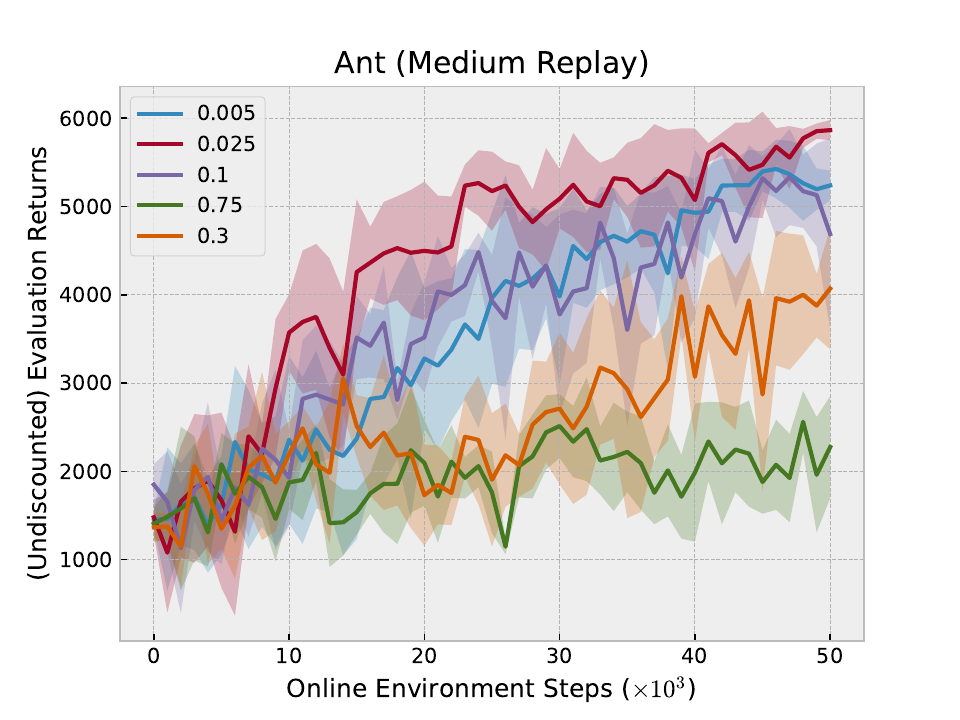}
    \end{subfigure}
    \begin{subfigure}{0.32\textwidth}
        \includegraphics[width=\textwidth]{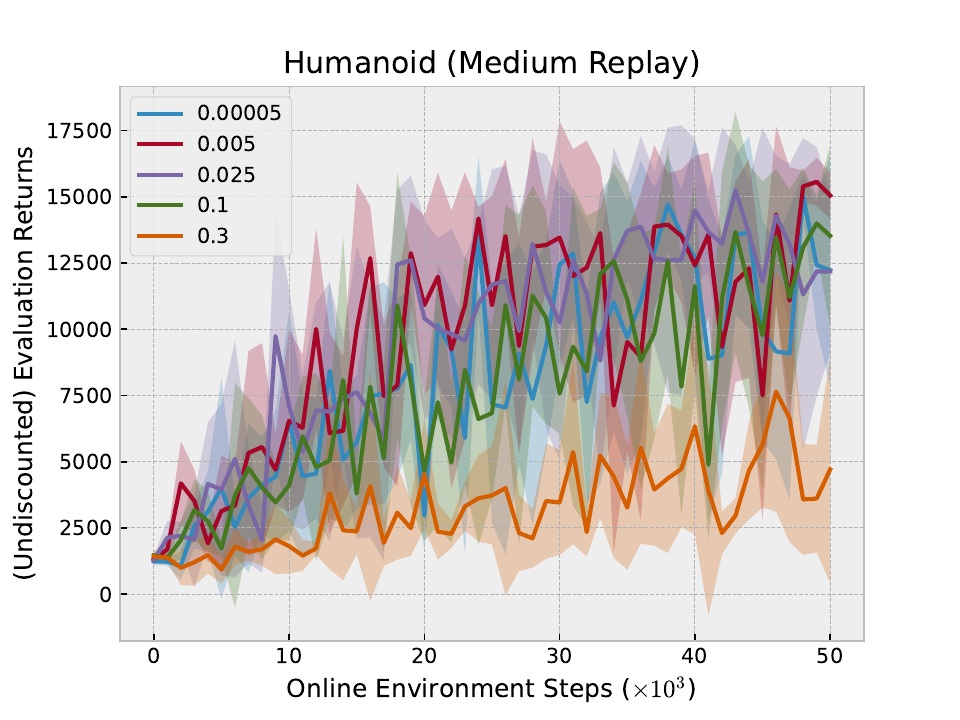}
    \end{subfigure}
    \caption{Undiscounted evaluation returns for the planning noise experiment.}
    \label{fig:more-planning-noise}
\end{figure*}

In Figure~\ref{fig:more-planning-noise}, we repeat the experiment in \S\ref{sec:planning-noise} for all environment-dataset combinations. We highlight that we find the same pattern as shown in the main body of the paper.

\end{document}